\documentclass[twoside,11pt]{article}

\usepackage[abbrvbib]{jmlr2e}

\usepackage{mathtools}
\usepackage{xcolor}
\usepackage{ntheorem} 
\usepackage{cleveref}
\usepackage{lastpage}
\usepackage{enumitem}
\usepackage{bbm}
\usepackage{bm}
\usepackage{float}
\usepackage[noend]{algpseudocode}
\usepackage{algorithm}
\usepackage{caption}
\usepackage{setspace}
\usepackage{placeins}
\usepackage{afterpage}
\usepackage[nottoc]{tocbibind}
\usepackage{multirow}
\usepackage{array}
\usepackage{tabularx}
\usepackage{hhline}
\usepackage{colortbl} 
\usepackage{hyperref}
\usepackage[mathscr]{euscript}
\usepackage{booktabs}
\hypersetup{hidelinks}

\newcommand{\R}{\mathbb R}
\newcommand{\N}{\mathbb N} 
\newcommand{\Nz}{{\mathbb N}\setminus\{0\}}

\newcommand{\HH}{\mathcal{H}}

\newcommand{\Xm}{\mathbb{X}_m}
\newcommand{\Xn}{\mathbb{X}_n}
\newcommand{\Yn}{\mathbb{Y}_n}

\newcommand{\mmdv}{\mathrm{MMD}^2_\lambda(p,q)}
\newcommand{\mmdh}{\widehat{\mathrm{MMD}}^2_\lambda}
\newcommand{\mmdhv}{\widehat{\mathrm{MMD}}^2_\lambda(\Xm,\Yn)}
\newcommand{\mmdhveq}{\widehat{\mathrm{MMD}}^2_\lambda(\Xn,\Yn)}
\newcommand{\mmdhmn}{\widehat{\mathrm{MMD}}^2_{\lambda,\mathtt{a}}}
\newcommand{\mmdhmnv}{\widehat{\mathrm{MMD}}^2_{\lambda,\mathtt{a}}(\Xm,\Yn)}
\newcommand{\mmdhmm}{\widehat{\mathrm{MMD}}^2_{\lambda,\mathtt{b}}}
\newcommand{\mmdhmmv}{\widehat{\mathrm{MMD}}^2_{\lambda,\mathtt{b}}(\Xn,\Yn)}

\newcommand{\pp}[1]{{{\mathbb P}_{p\times p}\!\left(#1\right)}}
\newcommand{\pq}[1]{{{\mathbb P}_{p \times q}\!\left(#1\right)}}
\newcommand{\pqr}[1]{{{\mathbb P}_{p \times q \times r}\!\left(#1\right)}}
\newcommand{\pqro}[1]{{{\mathbb P}_{p \times q \times r}\!\left(#1\right)}}
\newcommand{\ppro}[1]{{{\mathbb P}_{p \times p \times r}\!\left(#1\right)}}
\newcommand{\pqrr}[1]{{{\mathbb P}_{p \times q \times r\times r}\!\left(#1\right)}}
\newcommand{\ppr}[1]{{{\mathbb P}_{p \times p \times r}\!\left(#1\right)}}
\newcommand{\pprr}[1]{{{\mathbb P}_{p \times p \times r\times r}\!\left(#1\right)}}
\newcommand{\PPP}{{\mathbb P}}
\newcommand{\pii}{\pqr{\dbv=0}}

\newcommand{\e}{{\mathbb E}}

\newcommand{\epq}[1]{{{\mathbb E}_{p \times q}\!\left[#1\right]}}

\newcommand{\EE}[2]{{{\mathbb E}_{#1}\!\left[#2\right]}}

\newcommand{\vpq}[1]{{\mathrm{var}_{p \times q}\!\left(#1\right)}}
\newcommand{\VV}[2]{{\mathrm{var}_{#1}\!\left(#2\right)}}

\newcommand{\Zb}{\mathbb{Z}_B}
\newcommand{\Zbo}{\mathbb{Z}_{B_1}}
\newcommand{\Zbt}{\mathbb{Z}_{B_2}}
\newcommand{\Zbl}{\mathbb{Z}_{B_\ell}}

\newcommand{\qb}{\widehat{q}^{\,\lambda, B}_{1-\alpha}}
\newcommand{\qbv}{\widehat{q}^{\,\lambda, B}_{1-\alpha}\!\left(\Zb \big|\Xm,\Yn\right)}

\newcommand{\db}{{\Delta}^{\lambda, B}_\alpha}
\newcommand{\dbv}{{\Delta}^{\lambda, B}_\alpha\!\left(\Xm,\Yn,\Zb \right)}
\newcommand{\dbb}{{\Delta}^{\Lambda^{\!w}\!, B_{1:3}}_\alpha}
\newcommand{\dbbv}{{\Delta}^{\Lambda^{\!w}\!, B_{1:3}}_\alpha\!\left(\Xm,\Yn,\Zbo,\Zbt\right)}

\newcommand{\uuv}{u_\alpha^{B_2}\!\big(\Zbt\big|\Xm,\Yn,\Zbo\big)}
\newcommand{\uu}{u_\alpha^{B_2}}
\newcommand{\wuu}{{\widehat u}_{\alpha}^{B_{2:3}}}
\newcommand{\wuuv}{{\widehat u}_{\alpha}^{B_{2:3}}\!\big(\Zbt\big|\Xm,\Yn,\Zbo\big)}

\newcommand{\bb}{\beta}
\newcommand{\LL}{\lambda_1\cdots\lambda_d }
\newcommand{\h}{h_\lambda}
\newcommand{\kk}{k_\lambda}

\newcommand{\dd}{\mathrm{d}}
\newcommand{\vl}{\varphi_\lambda}
\newcommand{\Sb}{\mathcal{S}_d^s(R)}
\newcommand{\Sbb}{\big\{\Sb: s>0, R>0\big\}}
\newcommand{\lna}{\ln\!\left(\frac{1}{\alpha}\right)}
\newcommand{\one}[1]{\mathbbm{1}\!\left(#1\right)}
\newcommand{\mn}{\p{\frac{1}{m}+\frac{1}{n}}}
\newcommand{\mpn}{\p{m+n}}
\newcommand{\ww}{w_\lambda}

\newcommand{\p}[1]{\!\left( #1 \right)}
\newcommand{\acc}[1]{\left\{ #1 \right\}}
\newcommand{\cc}[1]{\!\left[\, #1 \,\right]}

\newcommand{\aamax}[1]{\underset{#1}{\mathrm{max}\, }}
\newcommand{\aasup}[1]{\underset{#1}{\mathrm{sup}\, }}

\newcommand{\aamin}[1]{\underset{#1}{\mathrm{min}\, }}
\newcommand{\aainf}[1]{\underset{#1}{\mathrm{inf}\, }}

\newcommand{\mat}[1]{\begin{pmatrix}#1\end{pmatrix}}

\newcommand{\ost}{\texttt{ost}}
\newcommand{\ora}{\texttt{oracle}}
\newcommand{\med}{\texttt{median}}
\newcommand{\spl}{\texttt{split}}
\newcommand{\automl}{\texttt{AutoML}}
\newcommand{\mmdi}{\texttt{MMDAgg increasing}}
\newcommand{\mmdd}{\texttt{MMDAgg decreasing}}
\newcommand{\mmdu}{\texttt{MMDAgg uniform}}
\newcommand{\mmdc}{\texttt{MMDAgg centred}}
\newcommand{\mmdagg}{\texttt{MMDAgg}$^\star$}
\newcommand{\mmdaggl}{\texttt{MMDAgg}$^\star$ \texttt{Laplace}}
\newcommand{\mmdaggg}{\texttt{MMDAgg}$^\star$ \texttt{Gaussian}}
\newcommand{\mmdagglg}{\texttt{MMDAgg}$^\star$ \texttt{Laplace Gaussian}}
\newcommand{\mmdagga}{\texttt{MMDAgg}$^\star$ \texttt{All}}

\newcommand\numberthis{\addtocounter{equation}{1}\tag{\theequation}}

\DeclareMathAlphabet{\mymathbb}{U}{BOONDOX-ds}{m}{n}

\DeclarePairedDelimiter\abs{\lvert}{\rvert}
\DeclarePairedDelimiter\norm{\lVert}{\rVert}
\makeatletter
\let\oldabs\abs
\def\abs{\@ifstar{\oldabs}{\oldabs*}}
\let\oldnorm\norm
\def\norm{\@ifstar{\oldnorm}{\oldnorm*}}
\makeatother

\DeclarePairedDelimiter\floor{\lfloor}{\rfloor}
\DeclarePairedDelimiter{\ceil}{\lceil}{\rceil}

\makeatletter
\newcommand*\bigcdot{\mathpalette\bigcdot@{.5}}
\newcommand*\bigcdot@[2]{\mathbin{\vcenter{\hbox{\scalebox{#2}{$\m@th#1\bullet$}}}}}
\makeatother

\newtheorem{theorem1}{Theorem}
\Crefname{theorem1}{Theorem}{Theorems}

\newtheorem{lemma1}[theorem1]{Lemma} 
\Crefname{lemma1}{Lemma}{Lemmas}

\newtheorem{proposition1}[theorem1]{Proposition} 
\Crefname{proposition1}{Proposition}{Propositions}

\newtheorem{corollary1}[theorem1]{Corollary}
\Crefname{corollary1}{Corollary}{Corollaries}

\newcommand{\theop}[2]{\begin{theorem1}[proof in #1] #2\end{theorem1}}
\newcommand{\propp}[2]{\begin{proposition1}[proof in #1] #2\end{proposition1}}
\newcommand{\corp}[2]{\begin{corollary1}[proof in #1] #2\end{corollary1}}
\newcommand{\lemp}[2]{\begin{lemma1}[proof in #1] #2\end{lemma1}}

\newcommand{\prop}[1]{\begin{proposition1}#1\end{proposition1}}

\makeatletter
\newcommand\footnoteref[1]{\protected@xdef\@thefnmark{\ref{#1}}\@footnotemark}
\makeatother

\algblockdefx{MRepeat}{EndRepeat}{\textbf{repeat}}{}
\algnotext{EndRepeat}

\makeatletter
\newcommand\fs@nobottomruled{\def\@fs@cfont{\bfseries}\let\@fs@capt\floatc@ruled
  \def\@fs@pre{\hrule height.8pt depth0pt \kern2pt}%
  \def\@fs@post{}
  \def\@fs@mid{\kern2pt\hrule\kern2pt}%
  \let\@fs@iftopcapt\iftrue}
\makeatother

\floatstyle{nobottomruled}
\restylefloat{algorithm}

\makeatletter
\AtBeginDocument{%
\def\HH@loop{%
  \ifx\@tempb`\def\next##1{\the\toks@\cr}\else\let\next\HH@let
  \ifx\@tempb|\if@tempswa
          \ifx\CT@drsc@\relax
           \HH@add{\hskip\doublerulesep}%
          \else
           \HH@add{{\CT@drsc@\vrule\@width\doublerulesep}}%
           \fi
          \fi\@tempswatrue
          \HH@add{{\CT@arc@\vline}}\else
  \ifx\@tempb:\if@tempswa
          \ifx\CT@drsc@\relax
           \HH@add{\hskip\doublerulesep}%
          \else
           \HH@add{{\CT@drsc@\vrule\@width\doublerulesep}}%
           \fi
              \fi\@tempswatrue
      \HH@add{\@tempc\HH@box\arrayrulewidth\arrayrulewidth\@tempc}\else
  \ifx\@tempb;\if@tempswa
          \ifx\CT@drsc@\relax
           \HH@add{\hskip\doublerulesep}%
          \else
           \HH@add{{\CT@drsc@\vrule\@width\doublerulesep}}%
           \fi
              \fi\@tempswatrue
      \HH@add{\@tempc\HH@box\z@\arrayrulewidth\@tempc}\else
  \ifx\@tempb##\if@tempswa\HH@add{\hskip\doublerulesep}\fi\@tempswatrue
         \HH@add{{\CT@arc@\vline\copy\@ne\@tempc\vline}}\else
  \ifx\@tempb~\@tempswafalse
           \if@firstamp\@firstampfalse\else\HH@add{&\omit}\fi
              \ifx\CT@drsc@\relax
                \HH@add{\hfil}\else
                 \HH@add{{%
                   \CT@drsc@\leaders\hrule\@height\HH@height\hfil}}%
               \fi
                 \else
  \ifx\@tempb-\@tempswafalse
           \gdef\HH@height{\arrayrulewidth}%
           \if@firstamp\@firstampfalse\else\HH@add{&\omit}\fi
              \HH@add{{%
                \CT@arc@\leaders\hrule\@height\arrayrulewidth\hfil}}%
                           \else
  \ifx\@tempb=\@tempswafalse
       \gdef\HH@height{\dimen\thr@@}%
       \if@firstamp\@firstampfalse\else\HH@add{&\omit}\fi
       \HH@add
          {\rlap{\copy\@ne}\leaders\copy\@ne\hfil\llap{\copy\@ne}}\else
  \ifx\@tempb t\HH@add{%
    \def\HH@height{\dimen\thr@@}%
    \HH@box\doublerulesep\z@}\@tempswafalse\else
  \ifx\@tempb b\HH@add{%
    \def\HH@height{\dimen\thr@@}%
    \HH@box\z@\doublerulesep}\@tempswafalse\else
  \ifx\@tempb>\def\next##1##2{%
     \HH@add{%
      {\baselineskip\p@\relax
       ##2%
      \global\setbox\@ne\HH@box\doublerulesep\doublerulesep}}%
       \HH@let!}\else
  \PackageWarning{hhline}%
      {\meaning\@tempb\space ignored in \noexpand\hhline argument%
       \MessageBreak}%
  \fi\fi\fi\fi\fi\fi\fi\fi\fi\fi\fi
  \next}
}

\jmlrheading{24}{2023}{1-\pageref{LastPage}}{10/21; Revised
3/23}{6/23}{21-1289}{Antonin Schrab, Ilmun Kim, M\'elisande Albert, B\'eatrice Laurent, Benjamin Guedj and Arthur Gretton}

\ShortHeadings{MMD Aggregated Two-Sample Test}{Schrab, Kim, Albert, Laurent, Guedj and Gretton}
\firstpageno{1}

\begin{document} 

\title{MMD Aggregated Two-Sample Test}

\author{\name Antonin Schrab \email a.schrab@ucl.ac.uk\\ 
	\addr Centre for Artificial Intelligence, University College London \& Inria London \\
	Gatsby Computational Neuroscience Unit, University College London \\
	London, WC1V 6LJ, UK
\AND
	\name Ilmun Kim \email ilmun@yonsei.ac.kr \\
	\addr Department of Statistics \& Data Science, Department of Applied Statistics, Yonsei University\\
	 Seoul, 03722, South Korea 
\AND
    \name M\'elisande Albert \email melisande.albert@insa-toulouse.fr \\
    \addr Institut de Math\'ematiques de Toulouse; UMR 5219, Universit\'e de Toulouse; CNRS, INSA; France
\AND
    \name B\'eatrice Laurent \email beatrice.laurent@insa-toulouse.fr \\ 
    \addr Institut de Math\'ematiques de Toulouse; UMR 5219, Universit\'e de Toulouse; CNRS, INSA; France
\AND
	\name Benjamin Guedj \email b.guedj@ucl.ac.uk\\ 
	\addr Centre for Artificial Intelligence, University College London \& Inria London \\
	London, WC1V 6LJ, UK
\AND
	\name Arthur Gretton \email arthur.gretton@gmail.com\\ 
	\addr Gatsby Computational Neuroscience Unit, University College London \\
	London, W1T 4JG, UK
}

\editor{Ingo Steinwart}

\maketitle

\vspace{-0.7cm}

\begin{abstract}


We propose two novel nonparametric two-sample kernel tests based on the Maximum Mean Discrepancy (MMD). 
First, for a fixed kernel, we construct an MMD test using either permutations or a wild bootstrap, two popular numerical procedures to determine the test threshold.
We prove that this test controls the probability of type I error non-asymptotically.
Hence, it can be used reliably even in settings with small sample sizes as it remains well-calibrated, which differs from previous MMD tests which only guarantee correct test level asymptotically.
When the difference in densities lies in a Sobolev ball, we prove minimax optimality of our MMD test with a specific kernel depending on the smoothness parameter of the Sobolev ball.
In practice, this parameter is unknown and, hence, the optimal MMD test with this particular kernel cannot be used.
To overcome this issue, we construct an aggregated test, called MMDAgg,
which is adaptive to the smoothness parameter.
The test power is maximised over the collection of kernels used, without requiring held-out data for kernel selection (which results in a loss of test power), or arbitrary kernel choices such as the median heuristic.
We prove that MMDAgg still controls the level non-asymptotically, and achieves the minimax rate over Sobolev balls, up to an iterated logarithmic term.
Our guarantees are not restricted to a specific type of kernel, but hold for any product of one-dimensional translation invariant characteristic kernels.
We provide a user-friendly parameter-free implementation of MMDAgg using an adaptive collection of bandwidths.
We demonstrate that MMDAgg significantly outperforms alternative state-of-the-art MMD-based two-sample tests on synthetic data satisfying the Sobolev smoothness assumption, and that, on real-world image data, MMDAgg closely matches the power of tests leveraging the use of models such as neural networks.

\end{abstract}
\smallskip


\clearpage

\tableofcontents


\section{Introduction}
\label{introduction}

We consider the problem of nonparametric two-sample testing, where we are given two independent sets of i.i.d.\ samples, and we want to determine whether these two samples come from the same distribution. 
This fundamental problem has a long history in statistics and machine learning, with numerous real-world applications in various fields,
including 
clinical laboratory science \citep{rodney2004comparison},
genomics \citep{chen2010two},
biology \citep{fisher2006alteration},
geology \citep{vermeesch2013multi}
and
finance \citep{horvath2013estimation}.

To compare samples from two probability distributions, we use a statistical test of the null hypothesis that the two distributions are equal, against the alternative hypothesis that they are different. Many such tests exist, and rely on different assumptions.
If we assume that the two probability distributions are Gaussian with the same variance, then we can perform a Student's t-test~\citep{student1908probable} to decide whether or not to reject the null hypothesis. 
However, the t-test is parametric in nature, and designed only for comparing two Gaussian distributions.
By contrast, our interest is in nonparametric tests, which are sensitive to general alternatives, without relying on specific distributional assumptions.
An example of such a nonparametric test is the Kolmogorov--Smirnov test \citep{massey1951kolmogorov} which uses as its test statistic the largest distance between empirical distribution functions of the two  samples. The limitation of the Kolmogorov--Smirnov test, however, is that it applies only to univariate data, and its multivariate extension is challenging \citep{bickel69kolmogorov}.

The test statistic we consider is an estimate of the Maximum Mean Discrepancy (MMD---\citealp{GreBorRasSchetal07c,gretton2012kernel}) which is a kernel-based metric on the space of probability distributions.
The MMD is an integral probability metric \citep{muller1997integral} and hence is defined as the supremum, taken over a class of smooth functions, of the difference of their expectations  under the two probability distributions.
This function class is taken to be the unit ball of a characteristic Reproducing Kernel Hilbert Space \citep{azonszajn1950theory, fukumizu2008kernel, sriperumbudur2011universality}, so the Maximum Mean Discrepancy depends on the choice of kernel. 
We work with a wide range of kernels, each parametrised by their bandwidths. 

There exist several heuristics to choose the kernel bandwidths. In the Gaussian kernel case, for example, bandwidths are often simply set to the median distance between pairs of points from both samples~\citep{gretton2012kernel}.
This strategy for  bandwidth choice does not provide any  guarantee of optimality, however. In fact, existing empirical results demonstrate that the median heuristic performs poorly (i.e.\ it leads to low test power) when differences between the two distributions occur at a lengthscale that differs sufficiently from the median inter-sample distance \cite[Figure 1]{gretton2012optimal}.
\citet{ramdas2015decreasing} and \citet{reddi2015high} show that the median heuristic scales as the square root of the dimension, and that using a bandwidth of the higher order with respect to the dimension generally leads to higher power.
Another approach is to split the data and learn a good kernel choice on data held out for this purpose \citep[e.g.][]{gretton2012optimal,liu2020learning},
however, the resultant reduction in data for testing can reduce overall test power at smaller sample sizes.

\paragraph{Our contributions.} Having motivated the problem, we summarize our contributions.
We first address the case where the  ``smoothness parameter''  $s$ of the task is known: that is, the distributions being tested
have densities in $\R^d$ whose difference lies in a Sobolev ball $\Sb$ with smoothness parameter $s$ and radius $R$.
For this setting, we construct a single MMD test that is optimal in the minimax sense over $\Sb$, for a specific choice of bandwidths which depend on $s$.

In practice, $s$ is unknown,  and our test must be adaptive to it. We therefore construct a test which is adaptive to $s$ in the minimax sense, by aggregating across tests  with different bandwidths,  and rejecting the null hypothesis if any individual test (with appropriately corrected level) rejects it. 
We refer to our proposed MMD aggregated test as MMDAgg.
By upper bounding the uniform separation rate of testing of MMDAgg, we prove that it is optimal (up to an iterated logarithmic term) over the Sobolev ball $\Sb$ for any $s>0$ and $R>0$. 

For the practical deployment of our test, we require numerical procedures for computing the test thresholds. 
We may obtain the threshold for a test of level $\alpha$ using either permutations or a wild bootstrap to estimate the $(1\!-\!\alpha)$-quantile of the test statistic distribution under the null hypothesis.
We prove that our theoretical guarantees still hold under both test threshold estimation procedures.
In the process of establishing these results, we demonstrate the equivalence between using a wild bootstrap and using a restricted set of permutations, which is of independent interest.
Through the use of either permutations or a wild bootstrap, we can theoretically guarantee that our proposed single MMD test and MMDAgg test both have non-asymptotic level, which differs from the original MMD test of \citet{gretton2012kernel}.
We believe that this non-asymptotic property of our tests contributes to their real-world applications. 
Indeed, in practice, settings in which the sample sizes are fixed and may not be assumed to be asymptotically large are very common (\emph{e.g.} medical data, seismological data, data for materials science, \emph{etc.}). 
While existing tests relying on asymptotic may fail to be well-calibrated in those settings, ours are guaranteed to correctly control the test level non-asymptotically.

We stress that the implementation of MMDAgg corresponds exactly to the test for which we prove theoretical guarantees: we do not make any further approximations in our implementation, nor do we require any prior knowledge on the underlying distribution smoothness.
All our theoretical results hold for any product of one-dimensional translation invariant characteristic kernels which are absolutely and square integrable. 
Our test is, to the best of our knowledge, the first to be {\em minimax adaptive} (up to an iterated logarithmic term) for various kernels, and not only for the Gaussian kernel.

Since our approach combines multiple MMD single tests across a large collection of bandwidths, it requires no tuning and our implementation is parameter-free.
Furthermore, since we consider various bandwidths simultaneously, our test is adaptive: 
it performs well both in cases requiring the kernel to have a small bandwidth and in those necessitating a large bandwidth.
This means that the same MMDAgg test can detect both local and global differences in densities, which is not the case for a single MMD test with fixed bandwidth.

The key contributions of our paper can be summarised as follows.
\begin{itemize}
	\item Based on either permutations or a wild bootstrap, we propose a single MMD test and theoretically prove that it has non-asymptotic level.
	By setting the kernel bandwidth adequately, we show that the test is minimax optimal over a Sobolev ball with known smoothness parameter.
	\item In order to be adaptive to this smoothness parameter (which is unknown in practice), we construct a two-sample aggregated MMD test, called MMDAgg, which does not require data splitting.
	We prove that MMDAgg controls the type I error non-asymptotically, and that it is optimal in the minimax sense (up to an iterated logarithmic term) for a wide range of kernels when using either permutations or a wild bootstrap to estimate the test threshold.
	\item In our experiments on synthetic data, on which the Sobolev smoothness assumption holds, we observe that MMDAgg obtains significantly higher power than all other state-of-the-art MMD adaptive tests considered.
	On real-world image data, MMDAgg almost matches the power obtained by tests leveraging the capacity of models such as neural networks to detect differences in image distributions.
	The power of MMDAgg is retained even when a large collection of kernels is considered (up to 12000 kernels in the experiments). 
	Experimentally, no cost in power is incurred for aggregating more kernels, the overall power appears to match the highest power achieved with a single kernel while correctly controlling the test level.
	As such, in practice, the user can consider as many kernels as computationally feasible.
	\item We provide a user-friendly parameter-free implementation of MMDAgg, both in Jax and in Numpy, available at \url{https://github.com/antoninschrab/mmdagg-paper}. This repository also contains code for the reproducibility of our experiments.
\end{itemize}

\paragraph{Related Works.}
We present an overview of works related to ours, more details are provided in \Cref{discussion}.
Our non-asymptotic aggregated test, which is minimax adaptive over the Sobolev balls $\Sbb$, originates from the works of \citet{fromont2012kernels,fromont2013two} and \citet{albert2019adaptive}.

\citet{fromont2012kernels,fromont2013two} consider the two-sample problem with sample sizes following independent Poisson processes.
In this framework, they construct an aggregated test using a different kernel-based estimator with a wild bootstrap. 
In the multivariate setting, with some condition on the kernel in the Fourier domain, they show minimax adaptivity of their test over Sobolev and anisotropic Nikol'skii-Besov balls.
\citet{albert2019adaptive} construct an aggregated independence test using Gaussian kernels.
The theoretical guarantees for the single MMD tests using a permutation-based threshold are related to the result of \citet[Proposition~8.4]{kim2020minimax}, also for Gaussian kernels.
Besides treating the adaptive two-sample case, rather than the independence case considered by \citet{albert2019adaptive}, the present work builds on these earlier results in two important ways. 
First, the optimality of our aggregated test is not restricted to the use of a specific kernel; it holds more generally for many popular choices of kernels.
Second, our theoretical guarantees for this adaptive test are proved to hold even under practical choices for the test threshold: namely the permutation and wild bootstrap approaches.

In this paper, we propose quadratic-time aggregated tests for two-sample testing. 
This present work, along with those of \citet{albert2019adaptive} and \citet{schrab2022ksd} on independence and goodness-of-fit testing, respectively, have been the basis of the later work of \cite{schrab2022efficient} who construct efficient (linear-time) variants of these three aggregated tests using incomplete $U$-statistics, and 
quantify the trade-off between computational efficiency and test power (in terms of uniform separation rate over Sobolev balls).

\paragraph{Outline.} 
The  paper is organised as follows. In \Cref{background}, we formalize the two-sample problem, review the theory of statistical hypothesis testing, and recall the definition of the Maximum Mean Discrepancy.
In \Cref{theory}, we construct the MMD single and aggregated (MMDAgg) two-sample tests, provide pseudocode for the latter, and derive theoretical guarantees for both.
Having introduced the required terminology, we then discuss how our results relate to other works in \Cref{discussion}.
We run various experiments in \Cref{experiments} to evaluate how well MMDAgg performs compared to alternative state-of-the-art MMD adaptive tests.
The paper closes with discussions and perspectives in \Cref{conclusion}.
Proofs, additional discussions, and further experimental results are provided in the Appendices.


\section{Background}
\label{background}

First, we formalise the two-sample problem in mathematical terms.
\medskip

\noindent\textbf{Two-sample problem.} Given independent samples $\Xm\! \coloneqq (X_i)_{1\leq i\leq m}$ and $\Yn\!\coloneqq (Y_j)_{1\leq j\leq n}$, consisting of i.i.d.\ random variables with respective probability density functions $p$ and $q$ on $\R^d$ with respect to the Lebesgue measure, can we decide whether $p\neq q$ holds?

\medskip
To tackle this problem, we work in the non-asymptotic framework and construct two nonparametric hypothesis tests in \Cref{theory}: a single MMD test for fixed kernel/bandwidth, and MMDAgg which aggregates multiple kernels/bandwidths.
In \Cref{hypothesis testing}, we first introduce the required notions about hypothesis testing.
We then recall the definition of the Maximum Mean Discrepancy and present two estimators for it in \Cref{mmd section}.

\subsection{Hypothesis testing}
	\label{hypothesis testing}

	We use the convention that $\PPP_{p\times q}$ denotes the probability with respect to $X_1,\dots,X_m\overset{\textrm{iid}}{\sim} p$ and $Y_1,\dots,Y_n\overset{\textrm{iid}}{\sim} q$ all independent of each other. 
	If given more random variables, say $Z_1,\dots,Z_t\overset{\textrm{iid}}{\sim} r$ for some probability density or mass function $r$, we use the notation $\PPP_{p\times q \times r}$.
	We follow similar conventions for expectations and variances.

	We address this two-sample problem by testing the null hypothesis $\HH_0\colon p=q$ against the alternative hypothesis $\HH_a\colon  p \neq q$. 
	Given a {\it test} $\Delta$ which is a function of $\Xm$ and $\Yn$, the null hypothesis is rejected if and only if $\Delta(\Xm,\Yn)=1$. 
	The test is usually designed to control the probability of {\it type I error} 
	$$
		\sup_{p} \pp{\Delta(\Xm,\Yn)=1} \leq \alpha
	$$ 
	for a given $\alpha\in(0,1)$, where the supremum is taken over all probability densities on $\R^d$.
	We then say that the test has {\it level} $\alpha$.
	For all the definitions, if the test $\Delta$ depends on other random variables, we take the probability with respect to those too. 
	For a given fixed level $\alpha$, the aim is then to construct a test with the smallest possible probability of {\it type II error} 
	$$
		\pq{\Delta(\Xm,\Yn) = 0}
	$$ 
	for specific choices of alternatives for which $p \neq q$. If this probability is bounded by some $\beta\in(0,1)$, we say that the test has {\it power} $1-\beta$ against that particular alternative.
	In the asymptotic framework, for a consistent test and a fixed alternative with $\norm{p-q}_2>0$, we can find large enough sample sizes $m$ and $n$ so that the test has power close to 1 against this alternative.
	In the non-asymptotic framework of this paper, the sample sizes $m$ and $n$ are fixed.
	We can then find an alternative with $\norm{p-q}_2$ small enough so that the test has power close to 0 against this alternative.
	Given a test $\Delta$, a class of functions $\mathcal{C}$ and some $\beta\in(0,1)$, one can ask what the smallest value $\tilde \rho>0$ is such that the test $\Delta$ has power at least $1-\beta$ against all alternative hypotheses satisfying $p-q\in \mathcal{C}$ and $\norm{p-q}_2>\tilde\rho$. 
	Clearly, this depends on the sample sizes: as $m$ and $n$ increase, the value of $\tilde \rho$ decreases.
	This motivates the definition of \textit{uniform separation rate} \citep{baraud2002non}
	$$
		\rho\!\left(\Delta, \mathcal{C}, \beta, M\right)
		\coloneqq \inf\biggl\{\tilde\rho>0: \sup_{(p,q)\in\mathcal{F}^M_{\tilde\rho}(\mathcal{C})}
		\pq{\Delta(\Xm,\Yn)=0} \leq \bb\biggr\} 
	$$
	where
	$
		\mathcal{F}^M_{\tilde\rho}(\mathcal{C}) \coloneqq \acc{(p,q): \max\p{\norm{p}_\infty,\norm{q}_\infty} \leq M, p-q \in \mathcal{C},\, \norm{p-q}_2 > \tilde\rho}
	$.
	For uniform separation rates, we are mainly interested in the dependence on $m+n$: for example we will show upper bounds of the form $a(m+n)^{-b}$ for positive constants $a$ and $b$ independent of $m$ and $n$. 
	The greatest lower bound on the uniform separation rates of all tests with non-asymptotic level $\alpha\in(0,1)$ is called the \textit{minimax rate of testing} \citep{baraud2002non}
	$$
		\underline\rho\!\left(\mathcal{C}, \alpha,\beta,M\right) \coloneqq \aainf{\Delta_\alpha}{\rho\!\left(\Delta_\alpha, \mathcal{C}, \beta,M\right)},
	$$
	where the infimum is taken over all tests $\Delta_\alpha$ of non-asymptotic level $\alpha$ for testing $\HH_0\colon  p=q$ against $\HH_a\colon  p \neq q$, and where we compare uniform separation rates in terms of growth rates as functions of $m+n$.
	This is a generalisation of the concept of critical radius introduced by \citet{ingster1993asymptotically, ingster1993minimax} to the non-asymptotic framework.
	A test is \textit{optimal in the minimax sense} \citep{baraud2002non} if its uniform separation rate is upper-bounded up to a constant by the minimax rate of testing. 
	As the class of functions $\mathcal{C}$, we consider the Sobolev ball
	\begin{equation}
		\label{sobolev}
		\Sb \coloneqq 
		\left\{ 
		f\in L^1\!\big(\R^d\big)\cap L^2\!\big(\R^d\big): \int_{\R^d} \norm{\xi}^{2s}_2 |\widehat f(\xi)|^2 \,\dd \xi \leq (2\pi)^d R^2
		\right\}
	\end{equation}
	with smoothness parameter $s>0$, radius $R>0$, and where $\widehat f$ denotes the Fourier transform of $f$, that is, $\widehat f(\xi) \coloneqq \int_{\R^d} f(x) e^{-ix^\top\xi} \,\dd x$ for all $\xi\in\R^d$. 
	Our aim is to construct a test which achieves the minimax rate of testing over $\Sb$ (up to an iterated logarithmic term) and which does not depend on the smoothness parameter $s$ of the Sobolev ball; such a test is called \textit{minimax adaptive}.

	As shown by \citet[Theorems 3 and 5]{li2019optimality}, the  minimax rate of testing over the Sobolev ball $\Sb$ is lower bounded as 
	\begin{equation}
		\label{lower bound equation}
		\underline\rho\!\left(\Sb, \alpha,\beta,M\right) \geq C_0(M,d,s,R,\alpha,\beta)\, \mpn^{-2s/(4s+d)}
	\end{equation}
	for some constant $C_0>0$ depending on $\alpha,\beta\in(0,1)$, $d\in\Nz$ and $M,s,R\in(0,\infty)$.
	Their proof is an extension of the results of 
	\citet{ingster1987minimax, ingster1993minimax} and we provide more details in \Cref{appendix lower}.
	We later construct a test with non-asymptotic level $\alpha$ and show in \Cref{sobolevusroptimal} that its uniform separation rate over $\Sb$ with respect to $m+n$ is at most $\mpn^{-2s/(4s+d)}$, up to some multiplicative constant. 
	This implies that the minimax rate of testing over the Sobolev ball $\Sb$ with respect to $m+n$ is exactly of order $\mpn^{-2s/(4s+d)}$.

\subsection{Maximum Mean Discrepancy}
	\label{mmd section}

	As a measure between two probability distributions, we consider the kernel-based {\it Maximum Mean Discrepancy} (MMD---\citealp{GreBorRasSchetal07c,gretton2012kernel}). In detail, for a given Reproducing Kernel Hilbert Space~$\mathcal{H}_k$ \citep{azonszajn1950theory} with kernel $k$, the MMD can be formalized as the integral probability metric~\citep{muller1997integral}
	$$
		\mathrm{MMD}(p, q; \mathcal{H}_k) 
		\coloneqq \sup_{f\in\mathcal{H}_k \,:\, \norm{f}_{\mathcal{H}_k} \leq 1} |\mathbb{E}_{X\sim p}[f(X)]
		- \mathbb{E}_{Y\sim q}  [f(Y)]|.
	$$
	Our particular interest is in a characteristic kernel $k$, which guarantees that we have $\mathrm{MMD}(p, q; \mathcal{H}_k) =0$
	if and only if $p=q$. We refer to the works of \citet{fukumizu2008kernel} and \citet{sriperumbudur2011universality} for details on characteristic kernels.
	It can easily be shown \citep[Lemma~4]{gretton2012kernel} that the MMD is the $\mathcal{H}_k$-norm of the difference between the mean embeddings 
	$\mu_p(u) \coloneqq \mathbb{E}_{X\sim p}\left[k(X,u)\right]$
	and $\mu_q(u) \coloneqq \mathbb{E}_{Y\sim q}\left[k(Y,u)\right]$ for $u\in\R^d$.
	Using this fact, a natural unbiased quadratic-time estimator for $\mathrm{MMD}^2(p,q; \mathcal{H}_k)$ \citep[Lemma~6]{gretton2012kernel} is 
	\begin{equation}
		\begin{aligned}
			\label{mmdmn}
			\widehat{\mathrm{MMD}}^2_{\mathtt{a}}(\Xm,\Yn;\mathcal{H}_k) \coloneqq 
			\frac{1}{m(m-1)} \sum_{1\leq i\neq i' \leq m} k(X_i,X_{i'})
			&+ \frac{1}{n(n-1)}  \sum_{1\leq j\neq j' \leq n} k(Y_j,Y_{j'})\\
			&- \frac{2}{mn} \sum_{i=1}^m \sum_{j=1}^n k(X_i,Y_j).
		\end{aligned}
	\end{equation}
	This is the minimum variance MMD estimator \citep[Section 5.1.4]{serfling1980approximation}.
	As pointed out by \citet{kim2020minimax}, this quadratic-time estimator can be written as a two-sample $U$-statistic (both of second order) \citep{hoeffding1992class} 
	\begin{equation}
		\label{mmdmnU}
		\widehat{\mathrm{MMD}}^2_{\mathtt{a}}(\Xm,\Yn;\mathcal{H}_k) = 
		\frac{1}{m(m-1)n(n-1)} 
		\sum_{1\leq i\neq i' \leq m}
		\sum_{1\leq j\neq j' \leq n}
		h_k(X_i, X_{i'}, Y_j, Y_{j'})
	\end{equation}
	where
	\begin{equation}
		\label{h}
		h_k(x, x', y, y') \coloneqq k(x,x') + k(y,y') - k(x,y') - k(x',y)
	\end{equation}
	for $x,y,x',y'\in\R^d$.
	Writing the estimator $\widehat{\mathrm{MMD}}^2_{\mathtt{a}}(\Xm,\Yn;\mathcal{H}_k)$ as a two-sample $U$-statistic can be theoretically appealing but we stress the fact that it can be computed in quadratic time using \Cref{mmdmn}.
	The unnormalised version of the test statistic $\widehat{\mathrm{MMD}}^2_{\mathtt{a}}(\Xm,\Yn;\mathcal{H}_k)$ was also considered in the work of \citet{fromont2012kernels}.

	For the special case when $m=n$, \citet[Lemma~6]{gretton2012kernel} also propose to consider a different estimator for the Maximum Mean Discrepancy which is the one-sample second-order $U$-statistic
	\begin{equation}
		\label{mmdmm}
		\widehat{\mathrm{MMD}}^2_{\mathtt{b}}(\Xn,\Yn;\mathcal{H}_k) \coloneqq \frac{1}{n(n-1)} 
		\sum_{1\leq i\neq j \leq n}
		h_k(X_i, X_j, Y_i, Y_j).
	\end{equation}
	Note that, unlike the estimator $\widehat{\mathrm{MMD}}^2_{\mathtt{a}}(\Xn,\Yn;\mathcal{H}_k)$, the estimator $\widehat{\mathrm{MMD}}^2_{\mathtt{b}}(\Xn,\Yn;\mathcal{H}_k)$ does not incorporate the terms $\{k(X_i,Y_i):i=1,\dots,n\}$.
	This means that the ordering of $\Xn=(X_i)_{1\leq i\leq n}$ and $\Yn=(Y_j)_{1\leq j\leq n}$ changes the estimator $\widehat{\mathrm{MMD}}^2_{\mathtt{b}}(\Xn,\Yn;\mathcal{H}_k)$.
	So, when using this estimator, we have to assume we are given a specific ordering of the samples. 
	While $\widehat{\mathrm{MMD}}^2_{\mathtt{b}}$ has slightly higher variance than $\widehat{\mathrm{MMD}}^2_{\mathtt{a}}$, computing $\widehat{\mathrm{MMD}}^2_{\mathtt{b}}$ is computationally much faster than evaluating $\widehat{\mathrm{MMD}}^2_{\mathtt{a}}$, as discussed in \Cref{efficientMMDAgg}.  

	The MMD depends on the choice of kernel, which we explore for our hypothesis tests.


\section{Construction of tests and bounds}
\label{theory}

This section contains our main contributions.
In \Cref{kernel}, we introduce some notation along with technical assumptions for our analysis. 
We then present in \Cref{single} two data-dependent procedures to construct a single MMD test that makes use of some specific kernel bandwidth. 
\Cref{powerMMD} provides sufficient conditions under which this single MMD test is powerful when the difference in densities is measured in terms of the MMD and of the $L^2$-norm. 
Based on these preliminary results, we prove an upper bound on its uniform separation rate over the Sobolev ball $\Sb$ in \Cref{usr}, which shows that for a specific choice of bandwidth, the corresponding single MMD test is optimal in the minimax sense. 
However, the optimal single MMD test relies on the unknown smoothness parameter $s$ of the Sobolev ball $\Sb$, which motivates the introduction of our aggregated MMDAgg test in \Cref{agg}.
Finally, we prove in \Cref{usragg} that MMDAgg is minimax adaptive over the Sobolev balls $\Sbb$.

\subsection{Assumptions and notation}
	\label{kernel}

	We assume that the sample sizes $m$ and $n$ are balanced up to a constant factor, meaning that there exists a positive constant $C >0$ such that
	\begin{equation}
		\label{mn}
		m \leq n 
		\quad \quad \text{and}\quad \quad
		n \leq C m.
	\end{equation}
	As can be seen in \Cref{mn equivalence}, this assumption allows us to upper bound terms such as $m^{-1}$ and $n^{-1}$ by $(m+n)^{-1}$ up to a constant depending on $C$. The smaller this constant $C$ is, the tighter our bounds on uniform separation rates will be.
	For fixed sample sizes, this condition of being balanced is always satisfied.
	For increasing sample sizes, the condition requires that both sample sizes increase at the same rate. 
	In particular, under this condition, it is not possible to fix one sample size and let the other tend to infinity.

	In general, we write $C_i(p_1,\dots,p_\ell)$ to express the dependence of a positive constant $C_i$ on some parameters $p_1,\dots,p_\ell$.

	We assume that we have $d$ characteristic kernels $(x,y)\mapsto K_i(x-y)$ on $\R\times\R$ for some  functions $K_i\colon \R\to\R$ lying in $L^1(\R)\cap L^2(\R)$ and satisfying
	$\int_{\R} K_i(u) \dd u = 1$ for $i=1,\dots,d$. 
	Then, for some bandwidth $\lambda = (\lambda_1,\dots,\lambda_d)\in(0,\infty)^d$, the function\footnote{\label{kernelfootnote}Multiplying the kernel $k_\lambda$ by a positive constant $C_\lambda$ does not affect the outputs of the single and aggregated tests as it simply scales both the test statistic and the quantile. 
	The requirements that $\int_{\R} K_i(u) \dd u = 1$ for $i=1,\dots,d$ and the scaling term $(\LL)^{-1}$ in the definition of the kernel $k_\lambda$ are not required for our theoretical results to hold. We introduce those simply for ease of notation in our statements and proofs.
	} 
	\begin{equation*}
		\label{k_eq}
		\kk(x,y) \coloneqq \prod_{i=1}^d \frac{1}{\lambda_i}K_i\p{\frac{x_i-y_i}{\lambda_i}}
	\end{equation*}
	is a characteristic kernel on $\R^d\times\R^d$ satisfying\footnote{Detailed calculations are presented at the beginning of \Cref{proofs}.} 
	\begin{equation}\label{hypintkcarre}
		\int_{\R^d} \kk(x,y) \dd x = 1
		\quad \quad \text{and}\quad \quad 
		\int_{\R^d} \kk(x,y)^2 \dd x  = \frac{\kappa_2(d)}{\LL}
	\end{equation}
	for the constant $\kappa_2(d)$ defined later in \Cref{kappa}.
	Using $K_i(u) =  \frac{1}{\sqrt{\pi}} \exp\p{-u^2}$ for $u\in\R$ and $i=1,\dots,d$, for example, yields the Gaussian kernel $\kk$. 
	Using $K_i(u) =  \frac{1}{2} \exp\p{-\abs{u}}$ for $u\in\R$ and
	$i=1,\dots,d$ yields the Laplace kernel $\kk$. 
	For notation purposes, given $\lambda = (\lambda_1,\dots,\lambda_d)\in(0,\infty)^d$, we also write 
	\begin{equation}
		\label{phi definition}
		\vl(u) \coloneqq \prod_{i=1}^d \frac{1}{\lambda_i}K_i\p{\frac{u_i}{\lambda_i}}
	\end{equation}
	for $u\in\R^d$, so that $\kk(x,y) = \vl(x-y)$ for all $x,y\in\R^d$.
	In this paper, we investigate the choice of kernel bandwidth for MMD tests.
	While our theoretical results on test power only hold for translation-invariant kernels on $\R^d\times\R^d$ (as introduced above), we stress that our single and aggregated MMD tests are well-defined and have well-calibrated non-asymptotic levels (\Cref{level,levelagg}) on any domain and for any positive definite characteristic kernel.

	For clarity, we denote $\mathrm{MMD}(p, q; \mathcal{H}_{\kk})$, $\widehat{\mathrm{MMD}}^2_{\mathtt{a}}(\Xm,\Yn;\mathcal{H}_{\kk})$, $\widehat{\mathrm{MMD}}^2_{\mathtt{b}}(\Xn,\Yn;\mathcal{H}_{\kk})$ and $h_{\kk}$ (all defined in \Cref{background}) simply by $\mathrm{MMD}_\lambda(p,q)$, $\mmdhmnv$, $\mmdhmmv$ and $\h$, respectively. 

	When $m\neq n$, we let $\mmdhv$ denote the estimator $\mmdhmnv$. 
	When $m=n$, we let $\mmdhveq$ denote either $\mmdhmn(\Xn,\Yn)$ or $\mmdhmmv$. 
	This means that, when $m=n$, all our results hold for both estimators.

\subsection{Non-asymptotic single MMD test with a fixed bandwidth}
	\label{single}

	Here, we consider the bandwidth 
	$\lambda\in(0,\infty)^d$ to be fixed {\em a priori}.
	The null and alternative hypotheses for the two-sample problem are 
	$\HH_0\colon  p=q$ against $\HH_a\colon  p \neq q$,
	or equivalently
	$\HH_0\colon  \mmdv =0$ against $\HH_a\colon  \mmdv >0$,
	provided that the kernels $K_1,\dots,K_d$ are characteristic.
	Using the samples $\Xm=(X_i)_{1\leq i\leq m}$ and $\Yn=(Y_j)_{1\leq j\leq n}$, we calculate the test statistic $\mmdh(\Xm,\Yn)$.
	Since we want our test to be valid in the non-asymptotic framework, we cannot rely on the asymptotic distribution of $\mmdh$ under the null hypothesis to compute the required threshold which guarantees the desired level $\alpha\in(0,1)$. 
	Instead, we use a Monte Carlo approximation to estimate the conditional $(1\!-\!\alpha)$-quantile of the permutation-based and wild bootstrap procedures given the samples $\Xm$ and $\Yn$ under the null hypothesis.
	For the estimator $\mmdhmnv$ defined in \Cref{mmdmn} we use permutations, while for the estimator $\mmdhmmv$ defined in \Cref{mmdmm} we use a wild bootstrap.

	In \Cref{relation section}, we provide some more in-depth discussion about the relation between those two procedures. 
	In particular, for the estimate $\mmdhmmv$, we show in \Cref{relation proposition} that using
	a wild bootstrap corresponds exactly to using permutations which either fix or swap $X_i$ and $Y_i$ for $i=1,\dots,n$.

\subsubsection{Permutation approach}
	\label{permutations}

	In this case, we consider the MMD estimator defined in \Cref{mmdmn} which can be written as 
	\begin{equation*}
		\allowdisplaybreaks
		\widehat{\mathrm{MMD}}^2_{\lambda,\mathtt{a}}(\Xm,\Yn) = 
		\frac{1}{m(m-1)n(n-1)} 
		\sum_{1\leq i\neq i' \leq m}
		\sum_{1\leq j\neq j' \leq n}
		\h(U_i, U_{i'}, U_{m+j}, U_{m+j'})
	\end{equation*}
	where
	$U_i \coloneqq X_i$ and $U_{m+j} \coloneqq Y_j$ 
	for $i=1,\dots,m$
	and $j=1,\dots,n$.
	Given a permutation function $\sigma\colon \{1,\dots,m+n\}\to \{1,\dots,m+n\}$, we can compute the MMD estimator on the permuted samples 
	$\Xm^\sigma \coloneqq \big(U_{\sigma(i)}\big)_{1\leq i \leq m}$
	and
	$\Yn^\sigma \coloneqq \big(U_{\sigma(m+j)}\big)_{1\leq j \leq n}$ to get
	\begin{align*}
		\widehat M_\lambda^{\,\sigma} &\coloneqq 
		\widehat{\mathrm{MMD}}^2_{\lambda,\mathtt{a}}(\Xm^\sigma,\Yn^\sigma) \numberthis \label{mmd permuted} \\
		&= \frac{1}{m(m-1)n(n-1)} 
		\sum_{1\leq i\neq i' \leq m}
		\sum_{1\leq j\neq j' \leq n}
		\h(U_{\sigma(i)}, U_{\sigma(i')}, U_{\sigma(m+j)}, U_{\sigma(m+j')}) \\
		&=\frac{1}{m(m-1)} \sum_{1\leq i\neq i' \leq m} \kk(U_{\sigma(i)},U_{\sigma(i')})
		+ \frac{1}{n(n-1)}  \sum_{1\leq j\neq j' \leq n} \kk(U_{\sigma(m+j)},U_{\sigma(m+j')})\\
		&\hspace{6.3cm}- \frac{2}{mn} \sum_{i=1}^m \sum_{j=1}^n \kk(U_{\sigma(i)},U_{\sigma(m+j)}).
	\end{align*}
	In order to estimate, with a Monte Carlo approximation, the conditional quantile of $\widehat M_\lambda^{\,\sigma}$ given $\Xm$ and $\Yn$, we uniformly sample $B$ i.i.d.\ permutations $\sigma^{(1)},\dots,\sigma^{(B)}$. 
	We denote their probability mass function by $r$, so that $\sigma^{(b)} \sim r$ for $b=1,\dots,B$.
	We introduce the notation $\Zb  \coloneqq \big(\sigma^{(b)}\big)_{1\leq b \leq B}$ and also simply write 
	$\widehat M_\lambda^{\, b} \coloneqq \widehat M_\lambda^{\,\sigma^{(b)}}$ for $b=1,\dots,B$.
	We can then use the values $\big(\widehat M_\lambda^{\,b}\big)_{1\leq b\leq B}$ to estimate the conditional quantile as explained in \Cref{single test subsubsection}.

\subsubsection{Wild bootstrap approach}
	\label{bootstrap subsection}

	In this case, we assume that $m=n$ and we work with the MMD estimator $\mmdhmm$ defined in \Cref{mmdmm}. 
	Recall that for this, we must assume an ordering of our samples which gives rise to a pairing $(X_i,Y_i)$ for $i=1,\dots,n$.
	Simply using permutations as presented in \Cref{permutations} would break this pairing and our estimators would consist of a signed sum of different terms because 
	\begin{equation*}
		\acc{\kk(U_{\sigma(i)},U_{\sigma(n+i)}):i=1,\dots,n} \neq \acc{\kk(U_{i},U_{n+i}):i=1,\dots,n}
	\end{equation*}
	for most permutations $\sigma\colon \{1,\dots,2n\}\to\{1,\dots,2n\}$. 
	The idea is then to restrict ourselves to the permutations $\sigma$ which, for $i=1,\dots,n$, either fix or swap $X_i$ and $Y_i$, in the sense that $\{U_{\sigma(i)},U_{\sigma(n+i)}\}=\{U_{i},U_{n+i}\}$, so that
	$\kk(U_{\sigma(i)},U_{\sigma(n+i)})=\kk(U_{i},U_{n+i})$.
	We show in \Cref{relation proposition} in \Cref{relation section} that this corresponds exactly to using a wild bootstrap, which we now define.

	Given $n$ i.i.d.\ Rademacher random variables
	$
		\epsilon\coloneqq(\epsilon_{1},\dots,\epsilon_{n})
	$ with values in $\{-1,1\}^n$,
	we let 
	\begin{equation}
		\label{mmd bootstrap}
		\widehat M_\lambda^{\,\epsilon} \coloneqq \frac{1}{n(n-1)} \sum_{1\leq i\neq j \leq n} \epsilon_{i}\epsilon_{j}\h(X_i,X_j,Y_i,Y_j).
	\end{equation}
	As for the permutation approach, in order to obtain a Monte Carlo estimate of the conditional quantile of $\widehat M_\lambda^{\,\epsilon}$ given $\Xn$ and $\Yn$, for $b=1,\dots,B$, we generate $n$ i.i.d.\ Rademacher random variables $\epsilon^{(b)}\coloneqq\big(\epsilon_{1}^{(b)},\dots,\epsilon_{n}^{(b)}\big)$ with values in $\{-1,1\}^n$ and compute $\widehat M_\lambda^{\, b} \coloneqq \widehat M_\lambda^{\,\epsilon^{(b)}}$.
	We write 
	$
		\Zb  \coloneqq \big(\epsilon^{(b)}\big)_{1\leq b \leq B}
	$
	and denote their probability mass function as $r$ to be consistent with the notation introduced in \Cref{permutations}, so that $\epsilon^{(b)}\sim r$ for $b=1,\dots,B$. 
	We next show in \Cref{single test subsubsection} how to estimate the conditional quantile using $\big(\widehat M_\lambda^{\,b}\big)_{1\leq b\leq B}$.

\subsubsection{Single MMD test: definition and level}
	\label{single test subsubsection}

	Depending on which MMD estimator we use, either $\mmdhmn$ from \Cref{mmdmn} or $\mmdhmm$ from \Cref{mmdmm}, we obtain $\big(\widehat M_\lambda^{\,b}\big)_{1\leq b\leq B}$ either as in \Cref{permutations} or as in \Cref{bootstrap subsection}, respectively. 
	Inspired by the work of \citet[Lemma~1]{romano2005exact} and \citet{albert2019adaptive}, in order to obtain the prescribed non-asymptotic test level, we also add the original MMD statistic
	$$
		\widehat{M}_\lambda^{\, B+1} \coloneqq
		\mmdh(\Xm,\Yn)
	$$
	which corresponds either to the case where the permutation is the identity or where the $n$ Rademacher random variables are equal to 1. 
	We can then estimate the conditional quantile of the distribution of either $\widehat M_\lambda^{\,\sigma}$ or $\widehat M_\lambda^{\,\epsilon}$ given $\Xm$ and $\Yn$ under the null hypothesis $\HH_0\colon  p=q$ by using a Monte Carlo approximation. In particular, our estimator of the conditional ($1-\alpha$)-quantile is given by  
	\begin{equation}
		\label{quantile}
		\qbv 
		\coloneqq \inf\!\bigg\{u \in \mathbb{R}: 1 - \alpha \leq \frac{1}{B+1}\sum_{b=1}^{B+1} \one{\widehat M_\lambda^{\, b}\leq u}\!\bigg\}
		= \widehat{M}_\lambda^{\,\bullet\ceil{(B+1)(1-\alpha)}} 
	\end{equation}
	where 
	$
		\widehat{M}_\lambda^{\,\bullet 1}  \leq \dots \leq \widehat{M}_\lambda^{\,\bullet B+1}
	$
	denote the ordered simulated test statistics $\big(\widehat M_\lambda^{\,b}\big)_{1\leq b\leq B+1}$.
	We then define the single MMD test $\db$ for some given bandwidth $\lambda\in(0,\infty)^d$ as
	$$
		\dbv \coloneqq \mathbbm{1}\p{\mmdh(\Xm, \Yn) > \qbv}.
	$$
	Intuitively, $\big(\widehat M_\lambda^{\,b}\big)_{1\leq b\leq B+1}$ simulate values of the MMD test statistic under the null hypothesis $\HH_0\colon p=q$, the quantile $\qb$ is defined such that only an $\alpha$-proportion of the simulated test statistics are greater than $\qb$. 
	As such, as shown in \Cref{level}, under $\HH_0$, the probability that the MMD test statistic is greater than the quantile (\emph{i.e.} rejecting the null) is non-asymptotically at most $\alpha$. 

	As shown in \Cref{prooflevel}, the $p$-value of the test can be computed as
	\begin{equation*}
		p_{\textrm{val}}^\lambda \coloneqq \frac{1}{B+1} \left(1 + \sum_{b=1}^{B} \one{\widehat{M}_\lambda^{\, b}\geq\widehat{M}_\lambda^{\, B+1}}\right)
	\end{equation*}
	and satisfies the property that
	\begin{equation*}
		p_{\textrm{val}}^\lambda\,\leq\, \alpha
		\quad
		\Longleftrightarrow
		\quad
		\mmdhv \,>\, \qbv.
	\end{equation*}
	Note that computing the quantile $\qbv$ requires sorting the simulated test statistics, while computing the $p$-value $p_{\textrm{val}}^\lambda$ does not.

	We now prove that this single MMD test has the desired non-asymptotic level $\alpha$, which we stress differs from the original asymptotic MMD test of \citet{gretton2012kernel}.
	We believe that this non-asymptotic property will contribute to the wide use of those MMD-based tests.

	\propp{\Cref{prooflevel}}{\label{level} 
		For fixed bandwidth $\lambda\in(0,\infty)^d$, $\alpha\in(0,1)$ and $B\in\Nz$, the test $\db$ has non-asymptotic level $\alpha$, that is 
		$$
		\ppr{\dbv=1}
		\leq \alpha
		$$
		for all probability density functions $p$ on $\R^d$.
	}

	This single MMD test $\db$ depends on the choice of bandwidth $\lambda$.
	In practice, one would like to choose $\lambda$ such that the test $\db$ has high power against most alternatives. 
	In general, a smaller bandwidth gives a narrower kernel which is well suited to detect local differences between probability densities such as small perturbations. 
	On the other hand, a larger bandwidth gives a wider kernel which is better at detecting global differences between probability densities. 
	We verify those intuitions in our experiments presented in \Cref{experiments}.
	While insightful, those do not tell us exactly how to choose the bandwidth.

	As mentioned in the introduction, in practice, there exist two common approaches to choosing the bandwidth of the single MMD test.
	The first one, proposed by \citet{gretton2012kernel}, is to set the bandwidth to be equal to the median inter-sample distance.
	The second approach involves splitting the data into two parts where the first half is used to choose the bandwidth that maximises the asymptotic power, and the second half is used to run the test.
	This was initially proposed by \citet{gretton2012optimal} for the linear-time MMD estimator, and later generalised by \citet{liu2020learning} to the case of the quadratic-time MMD estimator.
	The former approach has no theoretical guarantees, while the latter can suffer from a loss of power caused by the use of less data to run the test.
	Those two methods are further analysed in our experiments in \Cref{experiments}.

	In \Cref{powerMMD,usr}, we obtain theoretical guarantees for the power of the single MMD test $\db$ and specify the choice of the bandwidth that leads to minimax optimality.

\subsection{Controlling the power of the single MMD test}
	\label{powerMMD}

	We start by presenting conditions on the discrepancy measures $\textrm{MMD}_\lambda(p,q)$ and $\norm{p-q}_2$ under which the probability of type II error of the single MMD test 
	$$
		\pii = \pqr{\mmdhv\leq\qbv}
	$$ 
	is controlled by a small positive constant $\beta$. 
	We then express these conditions in terms of the bandwidth $\lambda$.
	We find a sufficient condition on the value of $\mmdv $ which guarantees that the single MMD test $\db$ has power at least $1\!-\!\beta$ against the alternative $\HH_a\colon p\neq q$.

	\lemp{\Cref{proofcontrolpower}}{\label{controlpower} 
		For $\alpha,\beta\in(0,1)$, and $B\in\Nz$, the condition
		$$
			\pqr{\!\mmdv \geq \sqrt{\frac{2}{\bb}\vpq{\mmdhv}} + \qbv\!} \geq 1-\frac{\beta}{2}
		$$
		is sufficient to control the probability of type II error such that 
		$$
			\pii\leq \bb.
		$$
	}

	If the densities $p$ and $q$ differ significantly in the sense that $\mmdv$ satisfies the condition of \Cref{controlpower}, then the probability of type II error of the single MMD test $\db$ against that alternative hypothesis is upper-bounded by $\beta$. 
	The condition includes two terms: the first term depends on $\beta$ as well as on the variance of $\mmdhv$, and the second is the conditional quantile estimated using the Monte Carlo method with either permutations or a wild bootstrap. 
	In the next two propositions, we make this condition more concrete by providing upper bounds for the variance and the estimated conditional quantile. 
	In particular, the upper bounds are expressed in terms of the bandwidth $\lambda$ and the sample sizes $m$ and $n$, which guides us towards the choice of the bandwidth with an optimal guarantee. We start with the variance term.

	\propp{\Cref{proofboundvar}}{\label{boundvar}
		Assume that
		$\max\left(\norm{p}_\infty,\norm{q}_\infty\right)\leq M$ for some $M>0$.
		Given $\vl$ as defined in \Cref{phi definition} and $\psi\coloneqq p-q$, there exists a positive constant $C_1(M,d)$ such that
		$$
			\vpq{\mmdhv} \leq C_1(M,d) \left(\frac{\norm{\psi*\vl}_2^2}{m+n}+\frac{1}{\mpn^2\LL}\right).
		$$
	}

	We now upper bound the estimated conditional quantile $\qbv$ in terms of $\lambda$ and $m+n$. Since this is a random variable, we provide a bound which holds with high probability.

	\propp{\Cref{proofboundquantile}}{\label{boundquantile}
		We assume $\max\left(\norm{p}_\infty,\norm{q}_\infty\right)\leq M$ for some $M>0$,
		$\alpha\in(0,0.5)$ and $\delta\in(0,1)$.
		For all $B\in\N$ satisfying $B\geq \frac{3}{\alpha^2}\p{\ln\p{\frac{4}{\delta}}+\alpha(1-\alpha)}$, we have
		$$
			\pqr{\qbv
			\leq C_2(M, d)\frac{1}{\sqrt{\delta}}\frac{\lna}{\mpn\sqrt{\LL}}}\geq 1-\delta
		$$
		for some positive constant $C_2(M, d)$.
	}

	Note that while this bound looks similar to the one proposed by \citet[Proposition 3]{albert2019adaptive} for independence testing, it differs in two major aspects.
	Firstly, while they consider the theoretical (unknown) quantile $q_{1-\alpha}^\lambda$,
	we stress that our bound holds for the random variable $\qbv$, which is the conditional quantile estimated using the Monte Carlo method with either permutations or a wild bootstrap. 
	Secondly, our bound holds for any bandwidth $\lambda\in(0,\infty)^d$ without any additional assumptions. In particular, we do not require the restrictive condition that  $\mpn\sqrt{\LL}>\lna$ which can in some cases imply that the sample sizes need to be very large.

	Having obtained upper bounds for $\textrm{var}_{p\times q}\big(\mmdhv\big)$ and $\qbv$, we now combine these with \Cref{controlpower} to obtain a more concrete condition for type II error control. 
	More specifically, the refined condition depends on $\lambda$, $m+n$ and $\beta$, and guarantees that the probability of type II error of the single MMD test $\db$, against the alternative $(p,q)$ defined in terms of the $L^2$-norm, is at most $\beta$.

	\theop{\Cref{proofphipower}}{\label{phipower}
		We assume $\max\left(\norm{p}_\infty,\norm{q}_\infty\right)\leq M$ for some $M>0$, 
		$\alpha\in(0,e^{-1})$, $\beta\in(0,1)$ and $B\in\N$ which satisfy $B\geq \frac{3}{\alpha^2}\big(\!\ln\!\big(\frac{8}{\beta}\big)+\alpha(1-\alpha)\big)$.
		We consider $\vl$ as defined in \Cref{phi definition} and let $\psi\coloneqq p-q$. Assume that $\LL\leq 1$.
		There exists a positive constant $C_3(M,d)$ such that if
		$$ 
			\norm{\psi}^2_2 ~\geq~ \norm{\psi-\psi * \varphi_\lambda}^2_2 ~+~ C_3(M,d)\frac{\lna}{\beta \mpn \sqrt{\LL}}, 
		$$
		then the probability of type II error satisfies $$\pii\leq \bb.$$
		}

	The main condition of \Cref{phipower} requires $\norm{p-q}_2^2$ to be greater than the sum of two quantities.
	The first one is the bias term $\norm{\psi-\psi * \varphi_\lambda}^2_2$ and the second one comes from the upper bounds in \Cref{boundvar,boundquantile} on the variance $\textrm{var}_{p\times q}\big(\mmdhv\big)$ and on the estimated conditional quantile $\qbv$.
	Now, we want to express the bias term $\norm{\psi-\psi * \varphi_\lambda}^2_2$ explicitly in terms of the bandwidths $\lambda$. 
	For this, we need some smoothness assumption on the difference of the probability densities. 

\subsection{Uniform separation rate of the single MMD test over a Sobolev ball}
	\label{usr}

	We now assume that $\psi\coloneqq p-q$ belongs to the Sobolev ball $\Sb$ defined in \Cref{sobolev}. 
	This assumption allows us to derive an upper bound on the uniform separation rate of the single MMD test in terms of the bandwidth $\lambda$ and of the sum of sample sizes $m+n$.

	\theop{\Cref{proofsobolevusr}}{\label{sobolevusr}
		We assume that $\alpha\in(0,e^{-1})$, $\beta\in(0,1)$, $s>0$, $R>0$, $M>0$ and $B\in\N$ satisfying $B\geq \frac{3}{\alpha^2}\big(\!\ln\!\big(\frac{8}{\beta}\big)+\alpha(1-\alpha)\big)$.
		Given that $\LL\leq 1$,
		the uniform separation rate of the test $\db$ over the Sobolev ball $\Sb$ can be upper bounded as follows
		$$
			\rho\!\left(\db, \Sb, \beta,M\right)^2 \leq C_4(M, d,s,R,\beta) \p{\sum_{i=1}^d \lambda_i^{2s} + \frac{\ln\!\left(\frac{1}{\alpha}\right)}{ \mpn \sqrt{\LL}} }
		$$
		for some positive constant $C_4(M, d,s,R,\beta)$.
	}

	The upper bound on the uniform separation rate $\rho\big(\db, \Sb, \beta,M\big)$ given by \Cref{sobolevusr} consists of two terms depending on the bandwidth $\lambda\in(0,\infty)^d$. 
	As the bandwidth $\lambda$ varies, there is a trade-off between those two quantities: increasing one implies decreasing the other.
	We can choose the optimal bandwidth $\lambda$ (depending on $m+n$, $d$ and $s$) in the sense that both terms have the same order with respect to the sum of sample sizes $m+n$.

	\corp{\Cref{proofsobolevusroptimal}}{\label{sobolevusroptimal}
		We assume that $\alpha\in(0,e^{-1})$, $\beta\in(0,1)$, $s>0$, $R>0$, $M>0$ and $B\in\N$ satisfying $B\geq \frac{3}{\alpha^2}\big(\!\ln\!\big(\frac{8}{\beta}\big)+\alpha(1-\alpha)\big)$.
		The test $\Delta^{\lambda^*\!, B}_\alpha$ for the choice of bandwidth
		$\lambda^*_i = \mpn^{-2/(4s+d)}$, $i=1,\dots,d$, is optimal in the minimax sense over the Sobolev ball $\Sb$, that is
		$$
			\rho\!\left(\Delta^{\lambda^*\!, B}_\alpha, \Sb, \beta, M\right) \leq C_5(M,d,s, R, \alpha, \beta)\, \mpn^{-2s / (4s+d)}
		$$
		for some positive constant $C_5(M,d,s, R, \alpha, \beta)$.
	}

	We have constructed the single MMD test $\Delta^{\lambda^*\!, B}_\alpha$ and proved that it is minimax optimal over the Sobolev ball $\Sb$ without any restriction on the sample sizes $m$ and $n$. However, it is worth pointing out that the optimality of the single MMD test hinges on the assumption that the smoothness parameter $s$ is known in advance, which is not realistic. Given this limitation, our next goal is to construct a test which does not rely on the unknown smoothness parameter $s$ of the Sobolev ball $\Sb$ and achieves the same minimax rate, up to an iterated logarithmic term, for all $s>0$ and $R>0$. This is the main topic of \Cref{agg} below.

\subsection{Non-asymptotic MMDAgg test aggregating multiple bandwidths}
	\label{agg}

	We propose to construct an aggregated test (MMDAgg) by combining multiple single MMD tests, which allows the test to be adaptive to the unknown the smoothness parameter of the Sobolev balls. 
	We use the powerful multiple testing correction of \citet[Equation 9]{romano2005stepwise}, for which we derive non-asymptotic level guarantees.
	Consider a finite collection $\Lambda$ of bandwidths in $(0,\infty)^d$ with an associated collection of positive weights\footnote{We stress that this differs from the notation often used in the literature (for example, for the independence aggregated test of \citealp{albert2019adaptive}) where the weights are defined  as $e^{-w_\lambda}$ rather than as $w_\lambda$.} $(w_\lambda)_{\lambda\in \Lambda}$, 
	which will determine the importance of each single MMD test over the others when aggregating all of them.
	We require that $\sum_{\lambda\in\Lambda} \ww  \leq 1$.
	For notational convenience, we let $\Lambda^{\!w}$ denote the collection of bandwidths $\Lambda$ with its associated collection of weights. 
	Intuitively, we want to define our aggregated MMDAgg test as the test which rejects the null hypothesis $\HH_0\colon p=q$ if one of the single MMD tests $\big(\Delta^{\lambda,B_1}_{u_\alpha \ww }\big)_{\lambda\in\Lambda}$ rejects the null hypothesis, where $ u_\alpha$ is defined as\footnote{Since $\alpha\in(0,1)$ and the function 
	$
		u \mapsto \PPP_{p\times p\times r}\big(\aamax{\lambda\in\Lambda}\!\big(\mmdhv\!-\!\widehat q_{1-u\ww }^{\,\lambda, B_1}\!\big(\Zbo\big|\Xm,\Yn\big)\big)\!>\!0\big)
	$
	is non-decreasing, tends to $0$ as $u$ tends to $0$, and tends to $1$ as $u$ tends to $\aamin{\lambda\in\Lambda}\ww^{-1}$, $u_\alpha$  is well-defined.} 
	$$
		u_\alpha = \sup\!\bigg\{\!u\!\in\!\Big(\!0,\aamin{\lambda\in\Lambda}\ww^{-1}\!\Big)\!: \ppr{\!\aamax{\lambda\in\Lambda}\!\p{\mmdhv\!-\!\widehat q_{1-u\ww }^{\,\lambda, B_1}\!\p{\Zbo\big|\Xm,\Yn}}\!>\!0} \!\leq\! \alpha\!\bigg\}
	$$
	to ensure that MMDAgg has level $\alpha$. 
	We stress that the data, as well as the choice of collections of bandwidths and weights, all affect the value of $u_\alpha$.
	In practice, the probability and the supremum in the definition of $u_\alpha $ cannot be computed exactly. 
	We can estimate the former using a Monte Carlo approximation and estimate the latter using the bisection method.
	We now explain this in more detail and provide a formal definition of our aggregated MMDAgg test.

	For the case of the estimator $\mmdhmn$ defined in \Cref{mmdmn}, we independently generate a permutation $\sigma^{(b,\ell)}\sim r $ of $\{1,\dots,m+n\}$ and compute $\widehat M_{\lambda,\ell}^{\,b}\coloneqq \widehat M_{\lambda}^{\,\sigma^{(b,\ell)}}$ as defined in \Cref{mmd permuted} for $\ell=1,2$, $b=1,\dots,B_\ell$ and $\lambda\in \Lambda$. 
	When working with the estimator $\mmdhmm$ defined in \Cref{mmdmm}, we independently generate $n$ i.i.d.\ Rademacher random variables $\epsilon^{(b,\ell)}=\big(\epsilon^{(b,\ell)}_1,\dots,\epsilon^{(b,\ell)}_n\big)\sim r $ and compute $\widehat M_{\lambda,\ell}^{\,b}\coloneqq \widehat M_{\lambda}^{\,\epsilon^{(b,\ell)}}$ as defined in \Cref{mmd bootstrap} for $\ell=1,2$, $b=1,\dots,B_\ell$ and $\lambda\in\Lambda$. 
	For consistency between the two procedures, we let $\Zbl^\ell \coloneqq \p{\mu^{(b,\ell)}}_{1\leq b \leq B_\ell}$ for $\ell=1,2$,
	where $\mu^{(b,\ell)}$ denotes either the permutation $\sigma^{(b,\ell)}$ or the Rademacher random variable $\epsilon^{(b,\ell)}$ for $\ell=1,2$ and $b=1,\dots,B_\ell$.
	With a slight abuse of notation, we refer to $\Zbo^1$ and $\Zbt^2$ simply as $\Zbo$ and $\Zbt$.
	For both estimators, we also let $\widehat{M}_{\lambda,1}^{\, B_1+1}\coloneqq\mmdh(\Xm, \Yn)$.
	We denote by 
	$
		\widehat{M}_{\lambda,1}^{\,\bullet 1}  \leq \dots \leq \widehat{M}_{\lambda,1}^{\,\bullet B_1+1}
	$
	the ordered elements $\big(\widehat{M}_{\lambda,1}^{\,b} \big)_{1\leq b \leq B_1+1}$.

	We use
	$\big(\widehat{M}_{\lambda,1}^{\,\bullet b} \big)_{1\leq b \leq B_1+1}$, which are computed using $\Zbo$, $\Xm$ and $\Yn$, to estimate the conditional $(1\!-\!a)$-quantile 
	$$
		\widehat q_{1-a}^{\,\lambda, B_1}\p{\Zbo\big|\Xm,\Yn} \coloneqq \widehat{M}_{\lambda,1}^{\,\bullet\ceil{(B_1+1)(1-a)}} 
	$$
	for any $a\in(0,1)$ as in \Cref{quantile}.
	As explained in \Cref{single test subsubsection}, $\widehat q_{1-a}^{\,\lambda, B_1}$ is defined such that an $a$-proportion of the test statistics $\big(\widehat{M}_{\lambda,1}^{\,b} \big)_{1\leq b \leq B_1+1}$ simulated under the null are greater than $\widehat q_{1-a}^{\,\lambda, B_1}$.
	By \Cref{level}, this ensures the single test with bandwidth $\lambda$ has non-asymptotic level $a$.

	We use $\big(\widehat{M}_{\lambda,2}^{\,\bullet b} \big)_{1\leq b \leq B_2}$, which are computed using $\Zbt$, $\Xm$ and $\Yn$, to estimate with a Monte Carlo approximation the probability
	\begin{equation}
		\label{Pu}
		\pqro{\aamax{\lambda\in\Lambda}\p{\mmdhv-\widehat q_{1-u
		\ww }^{\,\lambda, B_1}\p{\Zbo\big|\Xm,\Yn}}>0}
	\end{equation}
	which appears in the definition of $u_\alpha $. 
	We denote the approximated quantity by $\uu$, which is formally defined as
	\begin{align*}
		&\uuv\\
		\coloneqq\, &\sup\!\bigg\{\!u\!\in\!\!\Big(\!0,\aamin{\lambda\in\Lambda}\ww^{-1}\!\Big)\!\!:\! \frac{1}{B_2}\!\sum_{b=1}^{B_2}\!\one{\!\!\aamax{\lambda\in\Lambda}\p{\!\widehat M_{\lambda,2}^{\, b}\p{\!\mu^{(b,2)} \big|\Xm,\Yn\!}\!-\!\widehat q_{1-u \ww }^{\,\lambda, B_1}\!\p{\Zbo\big|\Xm,\Yn}\!}\!\!>\!0\!}\!\leq\! \alpha\!\bigg\} \\
		=\, &\sup\!\bigg\{\!u\!\in\!\!\Big(\!0,\aamin{\lambda\in\Lambda}\ww^{-1}\!\Big)\!\!:\! \frac{1}{B_2}\!\sum_{b=1}^{B_2}\!\one{\!\!\aamax{\lambda\in\Lambda}\p{\!\widehat M_{\lambda,2}^{\, b}-\widehat{M}_{\lambda,1}^{\,\bullet\ceil*{(B_1+1)(1-u \ww )}}}>0\!}\leq \alpha\!\bigg\}.
	\end{align*}
	Since the function $u\mapsto \frac{1}{B_2}\sum_{b=1}^{B_2}\mathbbm{1}\Big(\aamax{\lambda\in\Lambda}\p{\widehat M_{\lambda,2}^{\, b}-\widehat{M}_{\lambda,1}^{\,\bullet\ceil*{(B_1+1)(1-u \ww )}}}>0\Big)-\alpha$ is increasing, $\uu$ is actually the largest root of this function. As such, it can be computed in practice by using the bisection method for finding the root.
	We let\footnote{\label{Bfootnote}We use the condensed notation $B_{2:3}$ and $B_{1:3}$ to refer to $(B_2,B_3)$ and $(B_1,B_2,B_3)$, respectively.} $\wuu = \wuuv$ be the lower bound of the interval obtained by performing $B_3$ steps of the bisection method to approximate the supremum (\emph{i.e.} find the root) in the definition of $\uu$. 
	We then have
	$$
		u_\alpha^{B_2}
		\in
		\cc{
			{\widehat u}_{\alpha}^{B_{2:3}},\ 
			{\widehat u}_{\alpha}^{B_{2:3}} 
			+ 2^{-B_3}\,\aamin{\lambda\in\Lambda}\ww^{-1}
		}.
	$$
	We recall that the data, the collection of bandwidths, and the weights, all affect the value of the correction $u_\alpha$, and hence, also the value of its estimate ${\widehat u}_{\alpha}^{B_{2:3}}$.

	For $\alpha\in(0,1)$,
	we can then define our aggregated MMDAgg test\footnoteref{Bfootnote} $\dbb$ as rejecting the null hypothesis, that is $\dbbv = 1$, if one of the tests $\Big(\Delta_{\wuu \ww }^{\lambda, B_1}\Big)_{\lambda\in\Lambda}$ rejects the null hypothesis, that is
	$$
		\exists\, \lambda\in\Lambda : \mmdhv > \widehat q_{1-\wuuv \ww }^{\,\lambda, B_1}\!\p{\Zbo\big|\Xm,\Yn},
	$$
	or equivalently
	$$
		\exists\, \lambda\in\Lambda : \mmdhv > \widehat{M}_{\lambda,1}^{\,\bullet\ceil*{(B_1+1)\big(1-\wuuv \ww \big)}}\p{\Zbo\big|\Xm,\Yn}\!.
	$$

	The parameters of our MMDAgg test $\dbb$ are: its level $\alpha$, the finite collection $\Lambda^{\!w}$ of bandwidths with its associated weights, and the positive integers $B_1$, $B_2$ and $B_3$.
	We generate independent permutations or Rademacher random variables to obtain $\Zbo$ and $\Zbt$. 
	In practice, we are given realisations of $\Xm=(X_i)_{1\leq i\leq m}$ and $\Yn=(Y_j)_{1\leq j\leq n}$. Hence, we are able to compute $\dbbv$ to decide whether or not we should reject the null hypothesis $\HH_0\colon p=q$.
	This exact version of our aggregated MMDAgg test $\dbb$ can be implemented in practice with no further approximation. 
	We provide a detailed pseudocode of MMDAgg in \Cref{MMDAgg} and our code is available \href{https://github.com/antoninschrab/mmdagg-paper}{here}.
	In \Cref{efficientMMDAgg}, we further discuss how to efficiently compute the values $\widehat M_{\lambda,\ell}^{\,b}$ for $\ell=1,2$, $b=1,\dots,B_\ell$ and $\lambda\in \Lambda$ (corresponding to Step 1 of \Cref{MMDAgg}).

	\afterpage{

\begin{algorithm}
	\caption{MMDAgg $\dbb$}
	\label{MMDAgg}
	\begin{spacing}{1}
		\begin{algorithmic}
			\State \textbf{\underline{Inputs:}} 
			\State $\bullet$ samples $\Xm =(x_i)_{1\leq i \leq m}$ in $\R^d$ and $\Yn =(y_j)_{1\leq j \leq n}$ in $\R^d$ 
			\State $\bullet$ choice between permutations (\Cref{mmdmn}) or wild bootstrap (\Cref{mmdmm})
			\State $\bullet$ one-dimensional kernels $K_1,\dots,K_d$ satisfying the properties presented in \Cref{kernel} 
			\State $\bullet$ level $\alpha\in(0,e^{-1})$
			\State $\bullet$ finite collection of bandwidths $\Lambda$ in $(0,\infty)^d$
			\State $\bullet$ collection of positive weights $(w_\lambda)_{\lambda\in \Lambda}$ satisfying $\sum_{\lambda\in\Lambda} w_\lambda\leq 1$
			\State  $\bullet$ number of simulated test statistics $B_1$ to estimate the quantiles
			\State  $\bullet$ number of simulated test statistics $B_2$ to estimate the level correction
			\State  $\bullet$ number of iterations $B_3$ for the bisection method
			\State

			\State \textbf{\underline{Procedure:}}

			\State \textit{\underline{Step 1:} compute all simulated test statistics (see \Cref{efficientMMDAgg} for a more efficient Step 1)}
			\For {$\ell=1,2$ \textbf{and} $b=1,\dots,B_\ell$:}
					\State generate $\mu^{(b,\ell)}\sim r$ as in Sections \ref{permutations} \underline{or} \ref{bootstrap subsection} (permutations \underline{or} Rademacher)
					\For {$\lambda\in\Lambda:$}
						\State compute $\widehat M_{\lambda,\ell}^{\,b} \coloneqq \widehat M_{\lambda}^{\,\mu^{(b,\ell)}}$ as in Equations (\ref{mmd permuted}) \underline{or} (\ref{mmd bootstrap})
					\EndFor
			\EndFor
			\For {$\lambda\in\Lambda$:}
				\State compute $\widehat{M}_{\lambda,1}^{\, B_1+1}\coloneqq\mmdh(\Xm, \Yn)$ as in Equations (\ref{mmdmn}) \underline{or} (\ref{mmdmm})
				\State $\!\left(\widehat{M}_{\lambda,1}^{\,\bullet1},\dots,\widehat{M}_{\lambda,1}^{\,\bullet B_1+1}\right) =$ \texttt{sort\_by\_ascending\_order}$\left(\widehat{M}_{\lambda,1}^{\,1},\dots,\widehat{M}_{\lambda,1}^{\, B_1+1}\right)$

			\EndFor
			\State 

			\State \textit{\underline{Step 2:} compute $\widehat u_\alpha$ using the bisection method}
			\State $u_{\textrm{min}} \coloneqq 0$ \textbf{and} $u_{\textrm{max}} \coloneqq \aamin{\lambda\in\Lambda} w_\lambda^{-1}$
			\MRepeat $\ B_3$ \textbf{times}:
				\State compute $u \coloneqq  \frac{u_{\textrm{min}}+u_{\textrm{max}}}{2}$
				\State compute $P_u \coloneqq\frac{1}{B_2}\sum_{b=1}^{B_2}\one{\aamax{\lambda\in\Lambda}\p{\widehat M_{\lambda,2}^{\, b}- \widehat{M}_{\lambda,1}^{\,\bullet\ceil{(B_1+1)(1-uw_\lambda)}}}>0}$
				\If {${P}_u\leq \alpha$ \textbf{then} $u_{\textrm{min}} \coloneqq u$ \textbf{else} $u_{\textrm{max}} \coloneqq u$}
				\EndIf
			\EndRepeat
			\State $\widehat{u}_\alpha \coloneqq u_{\textrm{min}}$
			\State

			\State \textit{\underline{Step 3:} output test result}
			\If {$\widehat{M}_{\lambda,1}^{\, B_1+1} > \widehat{M}_{\lambda,1}^{\,\bullet\ceil{(B_1+1)(1-\widehat u_\alpha w_\lambda)}}$ \textbf{for some} $\lambda\in\Lambda$:}
				\State \textbf{return} 1 (reject $\HH_0$)
			\Else:
				\State \textbf{return} 0 (fail to reject $\HH_0$)
			\EndIf

			\State
			\State \textbf{\underline{Time complexity:}}\protect\footnotemark\ $\mathcal{O}\p{\abs{\Lambda}(B_1+B_2) (m+n)^2}$
			\State
			\State \textbf{\underline{Space complexity:}} $\mathcal{O}\p{(m+n)^2 + (B_1+B_2) (m+n)}$
			\vspace{-2cm}
		\end{algorithmic}
	\end{spacing}
\end{algorithm}
{\footnotetext{The time complexity is actually
$\mathcal{O}\p{\abs{\Lambda} (B_1+B_2) (m+n)^2+\abs{\Lambda}B_1 \ln(B_1)
+\abs{\Lambda}B_2B_3}$
which under the reasonable assumption $m+n>\max\big(\sqrt{\ln(B_1)},\sqrt{B_3}\big)$ 
gives $\mathcal{O}\p{\abs{\Lambda} (B_1+B_2) (m+n)^2}$.
}}}

	The only conditions we have on our weights $(w_\lambda)_{\lambda\in \Lambda}$ for the collection of bandwidths $\Lambda$ are that they need to be positive and to satisfy $\sum_{\lambda\in \Lambda}w_\lambda\leq 1$. 
	We now explain why this condition on the sum of the weights is not necessarily required.
	In general, the two aggregated tests with weights $(w_\lambda)_{\lambda\in \Lambda}$ and with scaled weights $(w_\lambda')_{\lambda\in \Lambda}$ where $w_\lambda'\coloneqq \frac{w_\lambda}{\sum_{\lambda\in \Lambda}w_\lambda}$ for $\lambda\in \Lambda$ are exactly the same.
	This is due to the way the correction of the levels of the single MMD tests is performed. 
	In particular, making the dependence of $\wuu$ on either $\Lambda^{\!w}$ or $\Lambda^{\!w'}$ explicit, we have
	$$
		{\widehat u}_{\alpha}^{B_{2:3}, \Lambda^{\!w'}}\big(\Zbt\big|\Xm,\Yn,\Zbo\big)
		=
		{\widehat u}_{\alpha}^{B_{2:3}, \Lambda^{\!w}}\big(\Zbt\big|\Xm,\Yn,\Zbo\big)\sum_{\lambda\in \Lambda}w_\lambda,
	$$
	and so 
	$
		{\widehat u}_{\alpha}^{B_{2:3}, \Lambda^{\!w'}}
		w_\lambda'
		=
		{\widehat u}_{\alpha}^{B_{2:3}, \Lambda^{\!w}}
		w_\lambda
	$, which implies that 
	\begin{equation}
		\label{weight scaling}
		{\Delta}^{\Lambda^{\!w'}\!, B_{1:3}}_\alpha\!\left(\Xm,\Yn,\Zbo,\Zbt\right)
		=
		{\Delta}^{\Lambda^{\!w}\!, B_{1:3}}_\alpha\!\left(\Xm,\Yn,\Zbo,\Zbt\right).
	\end{equation}	

	Consider some $u\in\big(0,\textrm{min}_{\lambda\in\Lambda}\ww^{-1}\big)$. Note that if a single MMD test $\Delta^{\lambda,B_1}_{u \ww}$ has a large associated weight $w_\lambda$, then its adjusted level $u \ww $ is bigger and so the estimated conditional quantile $\widehat q^{\,\lambda,B_1}_{1- u \ww }$ is smaller, which means that we reject this single test more often. 
	Recall that if a single MMD test rejects the null hypothesis, then the aggregated MMDAgg test necessarily rejects the null as well.
	It follows that a single test $\Delta^{\lambda,B_1}_{u \ww}$ with large weight $w_\lambda$ is viewed as more important than the other tests in the aggregated procedure.
	When running an experiment, putting weights on the bandwidths of the single MMD tests can be seen as incorporating prior knowledge about which bandwidths might be better suited to this specific experiment. 
	The choice of prior, or equivalently of weights, is further explored in \Cref{weighting strategies}.

	As presented in \Cref{single}, the $p$-value of one of the single MMD tests can be computed as
	\begin{equation*}
		p_{\textrm{val}}^\lambda \coloneqq \frac{1}{B_1+1} \left(1 + \sum_{b=1}^{B_1} \one{\widehat M_{\lambda,1}^{\,b}\geq \widehat M_{\lambda,1}^{\,B_1+1}}\right)
	\end{equation*}
	and, with its adjusted level $\wuu \ww$, it satisfies the property that
	\begin{equation*}
		p_{\textrm{val}}^\lambda \,\leq\, \wuu \ww
		\quad
		\Longleftrightarrow
		\quad
		\mmdhv \,>\, \widehat q_{1-\wuu \ww }^{\,\lambda, B_1}.
	\end{equation*}
	Hence, our aggregated MMDAgg test can also be expressed in terms of $p$-values as
	\begin{align*}
		\dbbv 
		&= \mathbbm{1}\p{\mmdhv \,>\, \widehat q_{1-\wuu \ww }^{\,\lambda, B_1}
		\text{ for some } \lambda \in \Lambda} \\
		&= \mathbbm{1}\p{p_{\textrm{val}}^\lambda \,\leq\, \wuu \ww \text{ for some } \lambda \in \Lambda}.
	\end{align*}
	We now show that MMDAgg indeed has non-asymptotic level $\alpha$.
	We emphasize the non-asymptotic nature of our aggregated test, which allows for the use of MMDAgg even in settings with small fixed sample sizes, where other asymptotic tests (such as the original MMD test of \citealp{gretton2012kernel}) fail to control correctly the probability of type I error.

	\propp{\Cref{prooflevelagg}}{\label{levelagg} 
		Consider $\alpha\in(0,1)$ and $B_1,B_2,B_3\in\Nz$.
		For a collection $\Lambda$ of bandwidths in $(0,\infty)^d$ and a collection of positive weights $(w_\lambda)_{\lambda\in \Lambda}$ satisfying $\sum_{\lambda \in \Lambda} \ww  \leq 1$, the MMDAgg test $\dbb$ has non-asymptotic level $\alpha$, that is 
		$$
			\pprr{\dbbv=1} \leq \alpha
		$$
		for all probability density functions $p$ on $\R^d$.
	}

\subsection{Uniform separation rate of MMDAgg over Sobolev balls}
	\label{usragg}

	In this section, we compute the uniform separation rate of our MMDAgg test $\dbb$ over the Sobolev ball $\Sb$. We then present a collection $\Lambda^{\!w}$ of bandwidths and associated weights for which our aggregated test $\dbb$ is almost optimal in the minimax sense.

	First, as part of the proof of \Cref{sobolevusragg} in \Cref{bound aggregated power}, we have shown that the following bound holds
	$$
		\pqrr{\dbbv = 0} \leq \frac{\beta}{2} + \aamin{\lambda\in\Lambda} \ {\pqro{{\Delta}^{\lambda, B_1}_{\alpha \ww /2}\!\left(\Zbo\big|\Xm,\Yn\right)=0}}.
	$$
	This means that we can control the probability of type II error of our MMDAgg test $\dbb$ by controlling the smallest probability of type II error of the single MMD tests 
	$\big(\Delta^{\lambda,B_1}_{\alpha \ww /2}\big)_{\lambda\in\Lambda}$ with adjusted levels.
	Hence, given a collection $\Lambda$ of bandwidths with its associated weights $(w_\lambda)_{\lambda\in \Lambda}$, if for some $\lambda\in\Lambda$ the single MMD test 
	$\Delta^{\lambda,B_1}_{\alpha \ww /2}$ has probability of type II error upper bounded by $\beta/2\in(0,0.5)$, then the probability of type II error of our aggregated MMDAgg test $\dbb$ is at most $\beta$. 
	Intuitively, this means that even if our collection of single MMD tests consists of only one `good' test (in the sense that it has high power with adjusted level) and many other `bad' tests (in the sense that they have low power with adjusted levels), MMDAgg would still have high power. 
	This is because when the `good' MMD test rejects the null hypothesis, MMDAgg also necessarily rejects it.
	Another point of view on this is that we do not lose any power by testing a wider range of bandwidths as long as 
	the weight of the `best' test remains the same.

	The uniform separation rate of our MMDAgg test $\dbb$ over the Sobolev ball $\Sb$ is then at most twice the lowest of the uniform separation rates of the single MMD tests $\big(\Delta_{\alpha w_\lambda/2}^{\lambda,B_1}\big)_{\lambda\in\Lambda}$.
	Combining this result with \Cref{sobolevusr}, we obtain the following  upper bound on the uniform separation rate of MMDAgg $\dbb$ over the Sobolev ball $\Sb$. 

	\begin{samepage}
		\theop{\Cref{proofsobolevusragg}}{\label{sobolevusragg}
			Consider a collection $\Lambda$ of bandwidths in $(0,\infty)^d$ such that $\LL\leq 1$ for all $\lambda\in \Lambda$ and a collection of positive weights $(w_\lambda)_{\lambda\in \Lambda}$ such that $\sum_{\lambda \in \Lambda} \ww  \leq 1$.
			We assume $\alpha\in(0,e^{-1})$, $\beta\in(0,1)$, $s>0$, $R>0,$ $M>0$ and $B_1,B_2,B_3\in\N$ satisfying 
			$B_1\geq \big(\mathrm{max}_{\lambda\in\Lambda}\, \ww^{-2}\big) \frac{12}{\alpha^2}\big(\log\big(\frac{8}{\beta}\big)+\alpha(1-\alpha)\big)$,
			$B_2\geq \frac{8}{\alpha^2}\ln\!\big(\frac{2}{\beta}\big)$
			and 
			$B_3 \geq \log_2\!\big(\frac{4}{\alpha}\,\mathrm{min}_{\lambda\in\Lambda}\ww^{-1}\big)$.
			The uniform separation rate of the aggregated MMDAgg test $\dbb$ over the Sobolev ball $\Sb$ can be upper bounded as follows
			$$
				\rho\!\left(\dbb, \Sb, \beta,M\right)^2 \leq C_6(M, d,s,R,\beta)\, \aamin{\lambda\in\Lambda}\p{\sum_{i=1}^d \lambda_i^{2s} + \frac{\ln\!\left(\frac{1}{\alpha}\right)+\ln\!\left(\frac{1}{w_\lambda}\right)}{ \mpn \sqrt{\LL}}}
			$$
			for some positive constant $C_6(M, d,s,R,\beta)$.
		}
	\end{samepage}

	We recall from \Cref{sobolevusroptimal} that the optimal choice of bandwidth $\lambda^*_i = \mpn^{-2/(4s+d)}$, $i=1,\dots,d$, for the single MMD test $\Delta^{\lambda^*\!,B}_\alpha$ leads to a uniform separation rate over the Sobolev ball $\Sb$ of order $\mpn^{-2s / (4s+d)}$ which is optimal in the minimax sense. However, this choice depends on the unknown smoothness parameter $s$ and so the test cannot be run in practice with this bandwidth. 
	We now propose a specific choice of collection $\Lambda^{\!w}$ of bandwidths and associated weights, which does not depend on $s$, and derive the uniform separation rate over the Sobolev ball $\Sb$ of our aggregated MMDAgg test $\dbb$ using that collection.
	Intuitively, the main idea is to construct a collection of bandwidths which includes a bandwidth (denoted $\lambda^*$) with the property that
	$$
		\frac{1}{a} \p{\frac{m+n}{\ln(\ln(m+n))}}^{-2/(4s+d)} \leq \lambda_i^* \leq \p{\frac{m+n}{\ln(\ln(m+n))}}^{-2/(4s+d)}
	$$
	for some $a>1$ and for $i=1,\dots,d$.
	The extra iterated logarithmic term comes from the additional weight term $\ln\big(\frac{1}{w_\lambda}\big)$ in \Cref{sobolevusragg}.

	\corp{\Cref{proofsobolevusraggoptimal}}{\label{sobolevusraggoptimal}
		We assume $\alpha\in(0,e^{-1})$, $\beta\in(0,1)$, $s>0$, $R>0$, $M>0$, $m+n>15$ so that $\ln(\ln(m+n))>1$ and $B_1,B_2,B_3\in\N$ satisfying 
		$B_1\geq \frac{3}{\alpha^2}\big(\!\ln\!\big(\frac{8}{\beta}\big)+\alpha(1-\alpha)\big)$,
		$B_2\geq \frac{8}{\alpha^2}\ln\!\big(\frac{2}{\beta}\big)$
		and 
		$B_3 \geq \log_2\!\big(\frac{2\pi^2}{3\alpha}\big)$.
		We consider our aggregated MMDAgg test $\dbb$ with the collection of bandwidths  
		$$
			\Lambda \coloneqq \Big\{\big(2^{-\ell},\dots,2^{-\ell}\big) \in (0,\infty)^d: \ell \in \Big\{1,\dots, \Big\lceil\frac{2}{d}\log_2\!\Big(\frac{m+n}{\ln(\ln(m+n))}\Big)\Big\rceil\Big\}\Big\}
		$$
		and the collection of positive weights 
		$
			w_\lambda \coloneqq \frac{6}{\pi^2\,\ell^2}
		$
		so that $\sum_{\lambda \in \Lambda} \ww  \leq 1$ for any sample sizes $m$ and $n$. 
		The uniform separation rate of the MMDAgg test $\dbb$ over the Sobolev ball $\Sb$ then satisfies
		$$
			\rho\!\left(\dbb, \Sb, \beta,M\right) \leq C_7(M, d,s,R,\alpha,\beta)\,
			\left(\frac{m+n}{\ln(\ln(m+n))}\right)^{-2s / (4s+d)}
		$$
		for some positive constant $C_7(M, d,s,R,\alpha,\beta)$.
		This means that the MMDAgg test $\dbb$, which does not depend on $s$ and $R$, is optimal in the minimax sense up to an iterated logarithmic term over the Sobolev balls $\Sbb$; the MMDAgg test $\dbb$ is minimax adaptive. 
	}

	Note that the choice of using negative powers of 2 for the bandwidths in \Cref{sobolevusraggoptimal} is arbitrary. 
	The result holds more generally using negative powers of $a$ for any real number $a>1$.

	With the specific choice of bandwidths and weights of \Cref{sobolevusraggoptimal}, we have proved that the uniform separation rate of the proposed aggregated MMDAgg test is upper bounded by $((m+n)/\ln(\ln(m+n)))^{-2s / (4s+d)}.$ Comparing this with the minimax rate~$(m+n)^{-2s/(4s+d)}$, we see that MMDAgg attains rate optimality over the Sobolev ball $\Sb$, up to an iterated logarithmic factor, and more importantly, the aggregated test does not depend on the prior knowledge of the smoothness parameter $s$.
	Our MMDAgg test is minimax adaptive over the Sobolev balls $\Sbb$.


\section{Related work}
\label{discussion}

In this section, we compare our results to a number of different adaptive kernel hypothesis testing approaches.

\citet{fromont2012kernels,fromont2013two} construct a two-sample aggregated test in a framework in which the sample sizes follow independent Poisson processes.
They use a different kernel-based estimator which corresponds to an unscaled version of the classical quadratic-time MMD estimator of \citet[Lemma 6]{gretton2012kernel}.
In their Poisson setting, they derive uniform separation rates for their aggregated test which is minimax adaptive over Sobolev balls, and over anisotropic Nikol'skii-Besov balls, up to an iterated logarithmic term.
The quantiles they consider are estimated with a wild bootstrap.
They also have an additional assumption on the kernel \citep[condition in the Fourier domain;][Equation 3.7]{fromont2013two}, which we do not require.

\citet{albert2019adaptive} consider the problem of testing whether two random vectors are dependent and use the kernel-based Hilbert-Schmidt Independence Criterion (HSIC---\citealp{gretton2005measuring}) as a dependence measure. 
Similarly to our work, they propose a non-asymptotic minimax adaptive test which aggregates single (HSIC) tests, and provide theoretical guarantees: upper bounds for the uniform separation rate of testing over Sobolev and Nikol'skii balls. 
In their independence testing setting,
the information about the problem is encoded in the joint distribution over pairs of variables, with the goal of determining whether this  is equal to the product of the marginals. 
This differs from the two-sample problem we consider, where we have samples from two separate distributions.

\citet{albert2019adaptive} define their single HSIC test using the theoretical quantile of the statistic under the null hypothesis, which is an unknown quantity in practice.
To implement the test, they propose a deterministic upper bound on the theoretical quantile \citep[Proposition~3]{albert2019adaptive}.
This upper bound  holds in the two-sample case under the 
 restrictive assumption 
$\mpn\sqrt{\LL}>\lna$
(this condition is adapted to the two-sample setting from their condition $n\sqrt{\lambda_1\dots \lambda_p \mu_1 \dots \mu_q}>\lna$ for  independence testing).
If the bandwidth is small (as it can be in the case of the optimal bandwidth $\lambda^*$ in the proof of \Cref{sobolevusraggoptimal}), then this condition implies that the results would hold only for very large sample sizes. 

By contrast with the above bound, we use a wild bootstrap or permutations to approximate the theoretical quantiles. 
While the theoretical quantiles are real numbers given data, our estimated quantiles are random variables given data.
This means that instead of having a deterministic upper bound on the theoretical quantiles \citep[Proposition~3]{albert2019adaptive}, we have an upper bound on our estimated conditional quantiles which holds with high probability as in \Cref{boundquantile}.
Our use of an estimated threshold in place of a deterministic upper bound has an important practical consequence: 
it allows us to drop  the assumption $\mpn\sqrt{\LL}>\lna$ entirely.

Another difference is how the level correction of the single MMD tests is performed. 
The  aggregated test of \citet{albert2019adaptive} involves a theoretical value $u_\alpha$ which cannot be computed in practice, we incorporate directly in our test a Monte Carlo approximation, using either a wild bootstrap or permutations, to estimate the probability under the null hypothesis, and use the bisection method to approximate the supremum. 
We stress that our theoretical guarantees of minimax optimality (up to an iterated logarithmic term) hold for our aggregated MMDAgg test which can be implemented without any further approximations. 
Finally, while the results of \citet{albert2019adaptive} hold only for the Gaussian kernel, ours are more general and hold for any product of one-dimensional characteristic translation invariant kernels which are absolutely and square integrable. 

\citet[Section~7]{kim2020minimax} propose an adaptive two-sample test for testing equality between two H\"older densities supported on the real $d$-dimensional unit ball. 
Instead of testing various bandwidths or kernels, they discretise the support in bins of equal sizes and aggregate tests with varying bin sizes. 
Each single test is a multinomial test based on the discretised data. 
Their strategy and the function class they use are both different from the one we consider, but they derive a similar upper bound on the uniform separation rate of testing over H\"older densities.
\citet[Proposition~8.4]{kim2020minimax} also mention the setting considered by \citet{albert2019adaptive} and prove an equivalent version of our \Cref{phipower} for single tests, using permutations for the Gaussian kernel. 
We consider both the permutation-based and wild bootstrap procedures, and our results hold more generally for a wide range of kernels.
With those aforementioned differences, \citet[Example~8.5]{kim2020minimax} anticipate that one can use a similar reasoning to  \citet{albert2019adaptive} to obtain minimax optimality of the single MMD tests. We provide the full statement and proof of this result in our more general setting.

\cite{li2019optimality} present goodness-of-fit, two-sample and independence aggregated asymptotic tests and also establish the minimax rates over Sobolev balls for these three settings.
Their tests use the Gaussian kernel and heavily rely on the asymptotic distribution of the test statistic, while our test  is non-asymptotic and is not limited to a particular choice of kernel.
Their tests are adaptive over Sobolev balls (which they define in a slightly different way than in our case) provided that the smoothness parameter satisfies $s \geq d/4$. We do not have such a restriction.
We also note that they assume that the two densities belong to a Sobolev ball, rather than assuming only that the difference of the densities lies in a Sobolev ball.
We also point out the work of \citet{tolstikhin2016minimax} who derive lower bounds for MMD estimation based on finite samples for any radial universal kernel \citep{sriperumbudur2011universality}.
They establish the minimax rate optimality of the MMD estimators \citep[$V$-statistic and $U$-statistic;][]{lee1990ustatistic}.

\citet{gretton2012optimal}
address kernel adaptation for the linear-time MMD, where the test statistic is computed as a running average (this results in a statistic with greater variance, but allows the  processing of larger sample sizes).
They  propose to choose the kernel by  splitting the data, and 
using one part to select the bandwidth which maximises the estimated ratio of the Maximum Mean Discrepancy to the standard deviation.
They show that maximizing this criterion for the linear-time setting corresponds to maximizing the  asymptotic power of the test.
The test is then performed on the remaining part of the data. 
\citet{sutherland2016generative} and \citet{liu2020learning} address kernel adaptation for the quadratic-time MMD using the same sample-splitting strategy, and show that the ratio of the MMD to the standard deviation under the alternative can again be used as a good proxy for test power.
\citet{liu2020learning} in particular
propose a regularized estimator for the variance under the alternative hypothesis, which admits a convenient closed-form expression.
Generally, kernel choice by sample splitting gives better results than the median heuristic, as the former is explicitly tuned to optimize the asymptotic power (or a proxy for it).
The price to pay for this increase in performance, however, is that we cannot use all the data for the test. 
In cases where we have access to almost unlimited data this clearly would not be a problem, but in cases where we have a restricted number of samples and work in the non-asymptotic setting, 
the loss of data to kernel selection might actually result in a net reduction in power, even after kernel adaptation.
For better data efficiency, \citet{kubler2022witness} propose an MMD test which uses held-out data not only for kernel selection, but also for choosing weights and test locations for the MMD witness function.
In later work, leveraging recent advances in supervised learning and also relying on sample splitting, \citet{kubler2022automl} construct a test which learns the witness function directly by training, for a given amount of time (\emph{i.e.} one minute), an AutoGluon model \citep{erickson2020autogluon} which can be, for example, a neural network.

\citet{kubler2020learning} propose another approach to an MMD adaptive two-sample test which does not require data splitting. 
Using all the data, they select the linear combination of test statistics with different bandwidths (or even different kernels) which is optimal in the sense that it maximises a power proxy, they then run their test using again all the data.
Using the post-selection inference framework \citep{fithian2014optimal,lee2016exact}, they are able to correctly calibrate their test to account for the introduced dependencies. 
This framework requires asymptotic normality of the test statistic under the null hypothesis, however, and hence they are by design restricted to using the linear-time MMD estimate.
We observe in our experiments that using this estimate results in a significant loss in power when compared to tests which use the quadratic-time statistic.
\citet{yamada2018post} also use post-selection inference  to obtain a feature selection method based on the MMD, where the chosen features best distinguish the samples.

In a different setting, \citet{wynne2022kernel} study the efficiency of MMD-based tests when dimension increases, and propose an MMD-based two-sample test for Functional Data Analysis, a framework in which the samples consist of functions rather than of data points.
In this setting, \citet{wynne2021statistical} study the connections between kernel mean embeddings and statistical depth (i.e. how representative a point is from a given measure).


\section{Experiments}
\label{experiments}

For our aggregated MMDAgg test, we first introduce in \Cref{weighting strategies} four weighting strategies and a family of collections of bandwidths motivated by \Cref{sobolevusraggoptimal}.
Those collections depend on some parameters which would usually need to be chosen by the user, by contrast, we introduce in \Cref{collection bandwidths} a parameter-free adaptive collection for MMDAgg, which we recommend using in practice.
We then present in \Cref{other tests} some other state-of-the-art MMD-based two-sample tests we will compare ours to.
In \Cref{experimental details}, we provide details about our experimental procedure. 
We show that our aggregated MMDAgg test obtains high power on both synthetic and real-world datasets in \Cref{exp power synthetic,exp power mnist}, respectively.
In \Cref{continuous limit}, we observe that MMDAgg retains power even in the continuous limit of the collection of bandwidths.
We show in \Cref{shift experiment} that, on image shift experiments, MMDAgg matches the power of tests using neural networks, even for large sample sizes.
Finally, in \Cref{results additional experiments}, we briefly report the results from the additional experiments presented in \Cref{experiments appendix}.

\subsection{Weighting strategies and fixed bandwidth collections for MMDAgg}
	\label{weighting strategies}

	The positive weights $(w_\lambda)_{\lambda\in \Lambda}$ for the collection of bandwidths $\Lambda$ are required to satisfy $\sum_{\lambda\in \Lambda}w_\lambda \leq 1$.
	As noted in \Cref{weight scaling}, rescaling all the weights to ensure that $\sum_{\lambda\in \Lambda}w_\lambda = 1$ does not change the output of our aggregated test $\dbbv$.
	For this reason, we propose weighting strategies for which
	$\sum_{\lambda\in \Lambda}w_\lambda = 1$ holds.

	For any collection $\Lambda$ of $N$ bandwidths, one can use \textit{uniform} weights which we define as 
	$$
		w_\lambda^{\texttt{u}} \coloneqq \frac{1}{N}
		\quad \text{for }\  \lambda\in \Lambda.
	$$
	Using uniform weights should be prioritised if the user does not have any useful prior information to incorporate in the test.
	The choice of weights is entirely up to the user; 
	the weights can be designed to reflect any given prior belief about the location of the `best' bandwidths in the collection.
	Nonetheless, we also present three standard weighting strategies for incorporating prior knowledge when dealing with a more structured collection of bandwidths.

	Consider the case where we have some reference bandwidth $\lambda_{ref}\in(0,\infty)^d$ and we are interested in aggregating scaled versions of it, that is, we have an ordered collection of $N$ bandwidths defined as $\lambda^{(i)}\coloneqq c_i \lambda_{ref}$, $i=1,\dots,N$,
	for positive constants $c_1<\dots<c_N$.
	If we have no prior knowledge, then we would simply use the aforementioned uniform weights. 
	Suppose we believe that, if the two distributions differ, then this difference would be better captured by the smaller bandwidths in our collection, in that case we would use \textit{decreasing} weights
	$$
		w_{\lambda^{(i)}}^{\texttt{d}} \coloneqq \frac{1}{i} \p{\sum_{\ell=1}^N \ell^{-1}}^{-1}
		\quad \text{for }\  i=1,\dots,N.
	$$
	\begin{figure}[t]
		\centering
		\includegraphics[width=\textwidth]{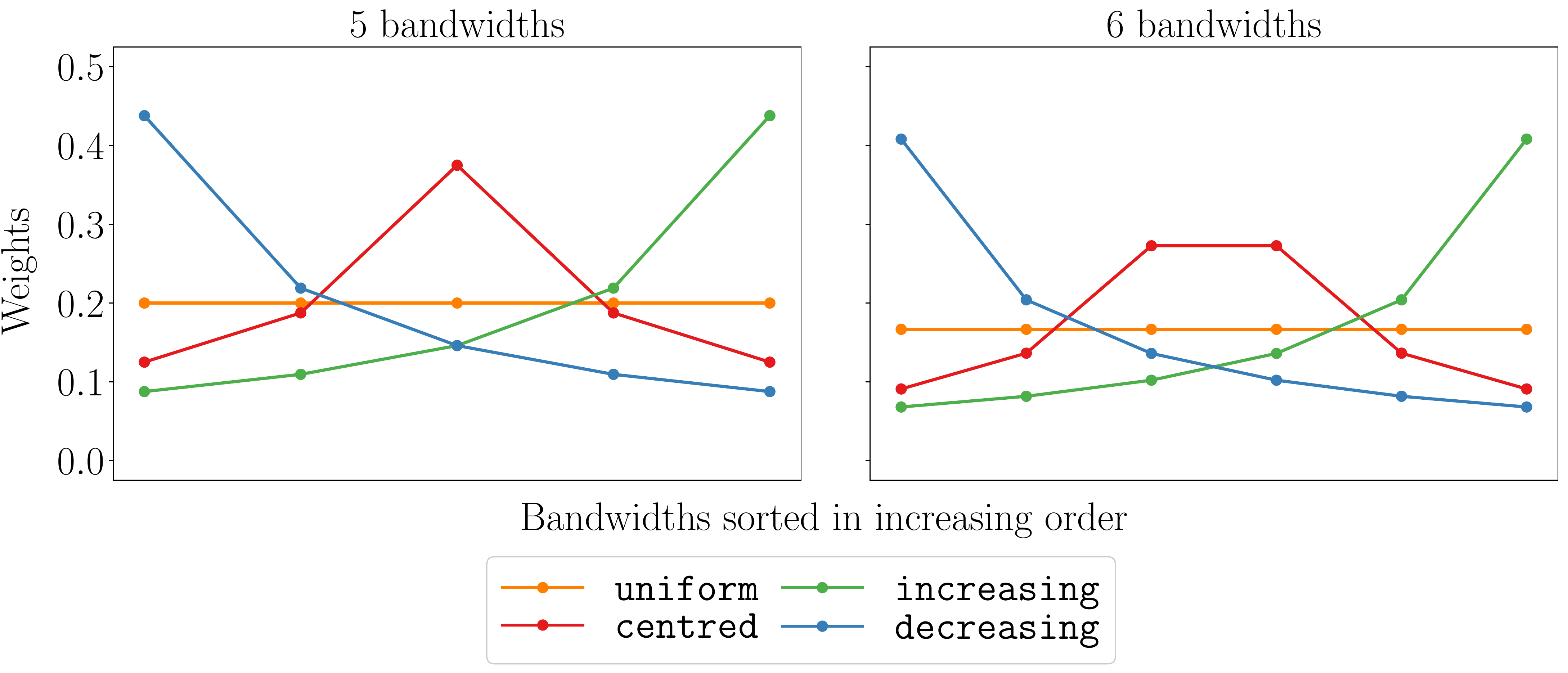}
		\caption{Weighting strategies.}
		\label{fig:weights}
	\end{figure}%
	On the contrary, if we think that the larger bandwidths in our collection are well suited to capture the difference between the two distributions, if it exists, then we would use \textit{increasing} weights
	$$
		w_{\lambda^{(i)}}^{\texttt{i}} \coloneqq \frac{1}{N+1-i} \p{\sum_{\ell=1}^N \ell^{-1}}^{-1}
		\quad \text{for }\  i=1,\dots,N.
	$$
	Finally, if our prior knowledge is that the bandwidths in the middle of our ordered collection are the most likely to detect the potential difference between the two densities, then we would use \textit{centred} weights which, for $N$ odd, are defined as 
	$$
		w_{\lambda^{(i)}}^{\texttt{c}} \coloneqq \frac{1}{\abs{\frac{N+1}{2}-i}+1}
		\p{\sum_{\ell=1}^N \p{\abs{\frac{N+1}{2}-\ell}+1}^{-1}}^{-1}
		\quad \text{for }\  i=1,\dots,N,
	$$
	and, for $N$ even, as 
	$$
		w_{\lambda^{(i)}}^{\texttt{c}} \coloneqq \frac{1}{\abs{\frac{N+1}{2}-i}+\frac{1}{2}}
		\p{\sum_{\ell=1}^N \p{\abs{\frac{N+1}{2}-\ell}+\frac{1}{2}}^{-1}}^{-1}
		\quad \text{for }\  i=1,\dots,N.
	$$
	All those weighting strategies are inspired from the weights of \Cref{sobolevusraggoptimal} which are defined as
	$w_{\lambda^{(i)}} \coloneqq i^{-2} \p{\sum_{\ell=1}^\infty \ell^{-2}}^{-1}$
	for $i\in\Nz$, where the square exponent is required in order to have a convergent series.
	However, in practice, using square exponents in our weights would assign extremely small weights to some of the bandwidths in our collection.
	This would be almost equivalent to disregarding those bandwidths in our aggregated MMDAgg test, which is not a desired property since if we are not interested in testing some bandwidths, then we would simply not include them in our collection.
	For this reason, we have defined our weighting strategies without the square exponent.
	We provide visualisations of our four weighting strategies for collections of 5 and 6 bandwidths in \Cref{fig:weights}.

	In our experiments, we refer to our aggregated MMDAgg test with those four weighting strategies as: \mmdu{}, \mmdd{}, \mmdi{} and \mmdc{}.
	For these, we use $B_1=500$ simulated test statistics to estimate the quantiles, $B_2=500$ simulated test statistics to estimate the probability in \Cref{Pu} for the level correction, and $B_3=100$ iterations for the bisection method.
	Motivated by \Cref{sobolevusraggoptimal}, those tests are used with collections of bandwidths of the form
	\begin{equation}
		\label{collection}
		\Lambda(\ell_-, \ell_+) \coloneqq \Big\{2^\ell \lambda_{med} \in (0,\infty)^d: \ell \in \big\{\ell_-,\dots, \ell_+\big\}\Big\}
	\end{equation}
	for $\ell_-, \ell_+ \in \mathbb{Z}$ such that $\ell_- < \ell_+$, where the median bandwidth $\lambda_{med}$ is
	$$
		\p{\lambda_{med}}_i \coloneqq \textrm{median}\!\acc{\abs{w_i-w_i'}:w,w'\in \Xm\cup \Yn, w\neq w'}
	$$
	for $i=1,\dots,d$. 
	We note that, for each experiment, we have chosen the values $\ell_-$ and $\ell_+$ which highlight the differences between the four weighting strategies. 
	In practice, it is not clear how to choose those values, instead, we recommend using the adaptive parameter-free collection of bandwidths introduced in \Cref{collection bandwidths} with uniform weights.

\subsection{Adaptive parameter-free collection of bandwidths for MMDAgg}
	\label{collection bandwidths}
	For radial basis function kernels (\emph{i.e.} kernels $k(x,y)$ which can be written as a function of $\norm{x-y}$ for some norm $\norm{\cdot}$), we recommend using a collection of bandwidths which, intuitively, discretises the interval between the smallest and the largest of the inter-sample distances\footnote{In practice, $D$ can be computed using at most 500 samples from $\Xm$ and 500 samples from $\Yn$.}
	$$
		D \coloneqq \big\{\norm{x-y} : x\in\Xm,y\in\Yn\big\}.
	$$
	More formally, we use
	$$
		\Lambda = \left\{\p{\frac{2\lambda_{max}}{\lambda_{min}/2}}^{\!(i-1)/(N-1)} \lambda_{min}/2 \,:\, i = 1,\dots, N\right\}
	$$
	which is a discretisation of the interval $[\lambda_{min}/2,\, 2\lambda_{max}]$ using $N$ points.
	We let $\lambda_{min}$ be the minimum value in $D$. 
	If this value is smaller than $0.1$, we  instead use the 5\% smallest value in $D$ for $\lambda_{min}$, if this quantity is still smaller than $0.1$, we use $\lambda_{min}=0.1$.
	For $\lambda_{max}$, we use the maximum value of $D$ or $0.3$ if this maximum is smaller than $0.3$. 
	In practice, we recommend using $N=10$ points, as can be observed in \Cref{fig:change} the power remain the same when increasing $N$ to be larger (\emph{i.e.} $N=100$ or $N=1000$).
	Since in general we might not have prior information about the location of well suited bandwidths, we recommend using the proposed collection of bandwidths with uniform weights as defined in \Cref{weighting strategies}.
	We use $B_1=2000$ and $B_2=2000$ simulated test statistics to estimate the quantiles and the probability in \Cref{Pu} for the level correction, respectively, and use $B_3=50$ steps of bisection method.
	We refer to this test as \texttt{MMDAgg}$^\star$ and emphasize the fact that it is run with exactly the same parameters across all experiments.

\subsection{State-of-the-art MMD-based two-sample tests}
	\label{other tests}

	\citet{gretton2012kernel} first suggested using the median heuristic to choose the bandwidth of the MMD test with the Gaussian kernel\footnote{\citet{gretton2012kernel} actually consider the unnormalised Gaussian kernel without the $\big(\LL \pi^{d/2}\big)^{-1}$ term, but as pointed out in \Cref{kernelfootnote}, this does not affect the output of the test.
	} corresponding to $K_i(u)\coloneqq \frac{1}{\sqrt{\pi}}\exp(-u^2)$ for $u\in\R$, $i=1,\dots,d$, so that
	$$
		\kk(x,y) 
		\coloneqq \prod_{i=1}^d \frac{1}{\lambda_i}K_i\p{\frac{x_i-y_i}{\lambda_i}}
		=\frac{1}{\LL \pi^{d/2}} \exp\p{-\sum_{i=1}^d\p{\frac{x_i-y_i}{\lambda_i}}^2}.
	$$
	They proposed to set the bandwidth to be equal to 
	$$
		\lambda_i \coloneqq \textrm{median}\!\acc{\norm{w-w'}_2:w,w'\in \Xm\cup \Yn, w\neq w'}
	$$
	for $i=1,\dots,d$.
	To generalise this approach to our case where $K_1,\dots,K_d$ need not all be the same, as explained in \Cref{weighting strategies}, we can in a similar way set the bandwidth\footnote{Note that those two ways of setting the bandwidths are not equivalent for the Gaussian kernel but they are each equally valid.} to 
	$$
		\lambda_i \coloneqq \textrm{median}\!\acc{\abs{w_i-w_i'}:w,w'\in \Xm\cup \Yn, w\neq w'}
	$$
	for $i=1,\dots,d$. With this specific definition for the bandwidth, we use the notation $\lambda_{med} \coloneqq (\lambda_1,\dots,\lambda_d)$.
	We refer to the single MMD test with the bandwidth $\lambda_{med}$ as \texttt{median} in our experiments.

	Another common approach for selecting the bandwidth was first introduced by \citet{gretton2012optimal} for the single MMD test using the linear-time MMD estimator. 
	The method was then extended to the case of the quadratic-time MMD estimator by \citet{sutherland2016generative}.
	It consists in splitting the data in two parts and in using the first part to select the bandwidth which maximises the asymptotic power of test, or equivalently the estimated ratio \citep[Equations 4 and 5]{liu2020learning} 
	\begin{equation}
		\label{estimated ratio}
		\frac{\mmdh(\Xn,\Yn)}{\widehat\sigma_{\lambda}(\Xn,\Yn)}
	\end{equation}
	where 
	$$
		\widehat\sigma_{\lambda}^2(\Xn,\Yn) \coloneqq \frac{4}{n^3}\sum_{i=1}^n\p{\sum_{j=1}^n \h(X_i,X_j,Y_i,Y_j) }^2-\frac{4}{n^4}\p{\sum_{i=1}^n\sum_{j=1}^n \h(X_i,X_j,Y_i,Y_j) }^2 +10^{-8}
	$$
	is a regularised positive estimator of the asymptotic variance of the quadratic-time estimator $\mmdh(\Xn,\Yn)$ under the alternative hypothesis $\HH_a$ for $m=n$.
	In our experiments, we select the bandwidth of the form $c\lambda_{med}$ for various positive values of $c$
	which maximises the estimated ratio. 
	The single MMD test with the selected bandwidth is then performed on the second part of the data.
	In our experiments, we refer to this test as \texttt{split}.

	Another interesting test to compare ours to, is the one which uses new data to choose the optimal bandwidth. 
	This corresponds to running the above test which uses data splitting on twice the amount of data. 
	In some sense, this represents the best performance the single MMD test can achieve as the test is run on the whole dataset with an optimal choice of bandwidth.
	As such, it is interesting to observe the difference in power between MMDAgg and this oracle test which uses extra data.
	In our experiments, we denote this test as \texttt{oracle}.

	A radically different approach to constructing an MMD adaptive two-sample test was recently presented by \cite{kubler2020learning}.
	They work in the asymptotic regime and require asymptotic normality under the null hypothesis of their MMD estimator, so they are restricted to using the linear-time estimator.
	Using all of the data, they compute the linear-time MMD estimates for several kernels (or several bandwidths of a kernel), they then select the linear combination of these which maximises a proxy for asymptotic power, and compare its value to their test threshold.
	They do not split the data but they are able to correctly calibrate their test for the introduced dependencies.
	For this, they prove and use a generalised version of the post-selection inference framework \citep{fithian2014optimal,lee2016exact} which holds for uncountable candidate sets (i.e.\ all linear combinations).
	In our experiments, we compare our aggregated MMDAgg test to their one-sided test (OST---\citealp{kubler2020learning}) for which we use their implementation. 
	This test is referred to as \ost{} in our experiments.

	In later work, \citet{kubler2022automl} propose an AutoML (Automated Machine Learning) test with an implementation which is essentially parameter-free. 
	Their test relies on sample splitting, on cross-validation, and on permuting the data, the witness function of the test is learnt by training an AutoGluon model \citep{erickson2020autogluon} for some prescribed amount of time (one minute by default). 
	Depending on the time limit and on the compute available, a different model will be chosen automatically.
	While such a black-box approach can certainly be convenient for everyday users, this convenience comes at the expense of reproducibility (even on the same machine it can take different amounts of time to train identical models).
	We refer to this test in our experiments as \texttt{AutoML}.

\subsection{Experimental procedure}
	\label{experimental details}

	To compute the median bandwidth
	$$
		(\lambda_{med})_i \coloneqq \textrm{median}\!\acc{\abs{w_i-w_i'}:w,w'\in \Xm\cup \Yn, w\neq w'}
	$$
	for $i=1,\dots,d$, we use at most $1000$ randomly selected data points from $\Xm$ and at most $1000$ from $\Yn$, since the median is robust, this is sufficient to get an accurate estimate of it. 
	Moreover, we use a threshold so that the bandwidth is not smaller than 0.0001. 
	This avoids division by 0 in some settings where one component of the data points is always the same value, as it can be the case for the problem considered in \Cref{exp power mnist} which uses the MNIST dataset, where the pixel in one corner of the images is always black for every digit.

	We run all our experiments with the Gaussian kernel 
	$$
		\kk(x,y) 
		\coloneqq \prod_{i=1}^d \frac{1}{\lambda_i}K_i\p{\frac{x_i-y_i}{\lambda_i}}
		=\frac{1}{\LL \pi^{d/2}} \exp\p{-\sum_{i=1}^d\p{\frac{x_i-y_i}{\lambda_i}}^2}
	$$
	for $K_i(u)\coloneqq \frac{1}{\sqrt{\pi}}\exp(-u^2)$ for $u\in\R$, $i=1,\dots,d$, and with the Laplace kernel
	$$
		\kk(x,y) 
		\coloneqq \prod_{i=1}^d \frac{1}{\lambda_i}K_i\p{\frac{x_i-y_i}{\lambda_i}}
		=\frac{1}{\LL 2^d} \exp\p{-\sum_{i=1}^d\abs{\frac{x_i-y_i}{\lambda_i}}}
	$$
	for $K_i(u) \coloneqq \frac{1}{2} \exp\p{-\abs{u}}$ for $u\in\R$, $i=1,\dots,d$.
	As mentioned in \Cref{kernelfootnote}, MMDAgg does not depend on the scaling of the kernels. Hence, in our implementation we drop the scaling terms in front of the exponential functions, which is numerically more stable.

	We use a wild bootstrap for all our experiments, except for the one in \Cref{comparison permutations wild bootstrap} where we compare using the permutation-based and wild bootstrap procedures, and for the one in \Cref{different sample sizes} where we must use permutations as we consider different sample sizes $m\neq n$.
	We use level $\alpha = 0.05$ for all our experiments. 

	We run all our tests on three different types of data: 1-dimensional and 2-dimensional perturbed uniform distributions, and the MNIST dataset. 
	Those are introduced in \Cref{exp power synthetic,exp power mnist}, respectively. 
	We use sample sizes $m=n=500$ for the 1-dimensional perturbed uniform distributions and for the MNIST dataset, and use larger sample sizes $m=n=2000$ for the case of the 2-dimensional perturbed uniform distributions.

	For the \spl{} and \ora{} tests, we use two equal halves of the data, and \ora{} is run on twice the sample sizes.
	We choose the bandwidth which maximises the estimated ratio presented in \Cref{estimated ratio} out of the collection $\big\{c\lambda_{med}: c \in\{0.1,0.2,\dots,0.9,1\}\big\}$ when considering perturbed uniform distributions, and when considering the MNIST dataset we select it out of the collection $\big\{2^c\lambda_{med}: c \in\{10,11,\dots,19,20\}\big\}$.
	Similarly to our aggregated tests of \Cref{weighting strategies}, for the \med{}, \spl{} and \ora{} tests, we use $B=500$ simulated test statistics to estimate the quantile.

	For \mmdagg{}, we also use either the Gaussian or the Laplace kernel with $N=10$ bandwidths, we refer to those tests as \mmdaggg{} and \mmdaggl{}. 
	We also consider aggregating over different types of kernels, in particular, we can use both the Gaussian and Laplace kernels, each with $N=10$ bandwidths. This gives a collection consisting of $2N =20$ kernels, over which \mmdagglg{} aggregates.
	Finally, we propose to aggregate 12 kernels (each with $N=10$ bandwidths): the Gaussian kernel, the inverse multiquadric (IMQ) kernel, and the Mat\'ern kernels with the $\ell^1$ and $\ell^2$ distances for $\nu=0.5,1.5,2.5,3.5,4.5$ (the Laplace kernel is the Mat\'ern kernel with $\ell^1$ distance and $\nu=0.5$).
	This test aggregates over $12N=120$ kernels, we refer to it as \mmdagga{}. 
	Note that in \Cref{fig:change}, we consider $N=1000$ bandwidths, which means for example that the \mmdagga{} aggregates over $12N=12000$ kernels, and we observe that it retains its high power.

	To estimate the power in our experiments, we average the test outputs of 500 repetitions, that is, 500 times, we sample some new data and run the test. 
	We sample new data for each test with different parameters, except when we compare using either a wild bootstrap or permutations, in which case we use the same samples. 
	All our experiments are reproducible and our code is available \href{https://github.com/antoninschrab/mmdagg-paper}{here}. 

\subsection{Power experiments on synthetic data}
	\label{exp power synthetic}

	As explained in \Cref{appendix lower}, a lower bound on the minimax rate of testing over the Sobolev ball $\Sb$ can be obtained by considering a $d$-dimensional uniform distribution and a perturbed version of it with $P\in\Nz$ perturbations. 
	As presented in \Cref{f_theta}, the latter has density 
	\begin{equation}
		\label{f_theta bis}
		f_\theta (u) \coloneqq \mathbbm{1}_{[0,1]^d}(u) + c_d P^{-s} \!\sum_{\nu\in\{1,\dots,P\}^d} \theta_\nu \prod_{i=1}^dG\left(Pu_i-\nu_i\right), \quad u\in\R^d		
	\end{equation}
	\begin{figure}[t]
		\centering
		\includegraphics[width=\textwidth]{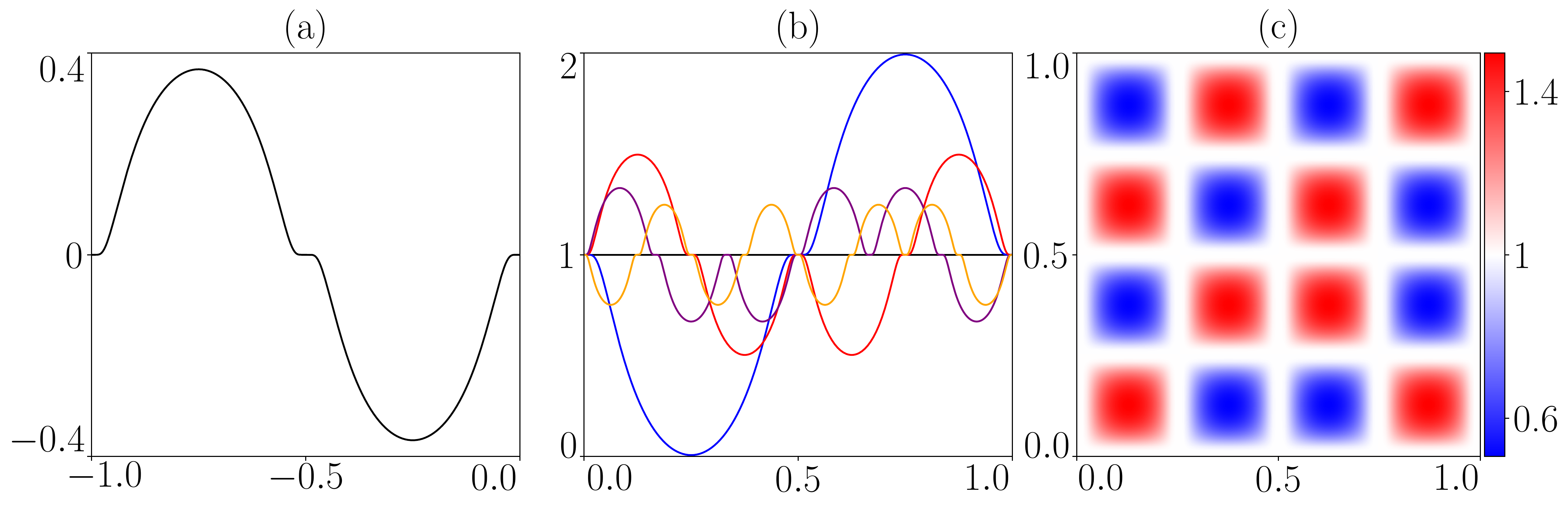}
		\captionsetup{format=hang}
		\caption{
		(a) Function $G$,
		(b) 1-dimensional uniform distribution with 0, 1, 2, 3 and 4 perturbations,
		(c) 2-dimensional uniform distribution with 2 perturbations.
		}
		\label{fig:uniform}
	\end{figure}%
	where $\theta= (\theta_\nu)_{\nu\in\{1,\dots,P\}^d}\in\{-1,1\}^{P^d}$, that is, $\theta$ is a vector of length $P^d$ with entries either $-1$ or $1$, and it is indexed by the $P^d$ $d$-dimensional elements of $\{1,\dots,P\}^d$, and
	$$
		G(u) \coloneqq \exp\left(-\frac{1}{1-(4u+3)^2} \right) \mathbbm{1}_{\left(-1,-\frac{1}{2}\right)}(u)
		- \exp\left(-\frac{1}{1-(4u+1)^2} \right) \mathbbm{1}_{\left(-\frac{1}{2},0\right)}(u), \quad u\in\R.
	$$
	\begin{figure}[t]
		\centering
		  \includegraphics[width=\textwidth]{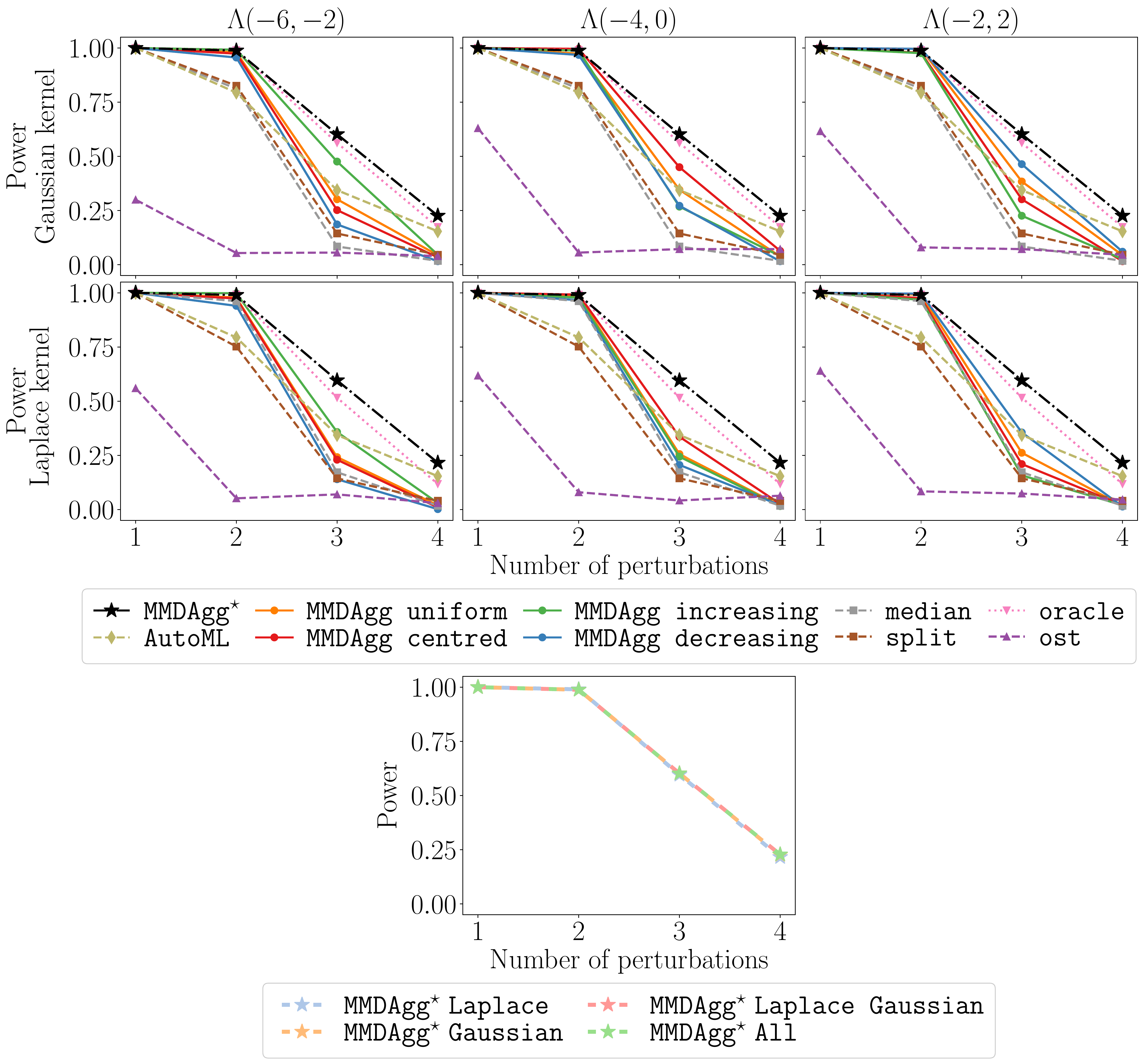}
		\captionsetup{format=hang}
		\caption{Power experiments with 1-dimensional perturbed uniform distributions using sample sizes $m=n=500$ with a wild bootstrap.}
		\label{fig:perturbed_uniform_1d}
	\end{figure}%
	We have added a scaling factor $c_d$ to emphasize the effect of the perturbations, in our experiments we use $c_1 = 2.7$ and $c_2 = 7.3$. 
	Those values were chosen to ensure that the densities with one perturbation remain positive on $[0,1]^d$.
	The uniform density with $P$ perturbations for $P=0,1,2,3,4$ when $d=1$ and for $P=2$ when $d=2$, as well as the function $G$, are plotted in \Cref{fig:uniform}.
	As shown by \citet{li2019optimality}, for $P$ large enough, the difference between the uniform density and the perturbed uniform density lies in the Sobolev ball $\Sb$ for some $R>0$.
	In our experiments, we choose the smoothness parameter of the perturbed uniform density defined in \Cref{f_theta bis} to be equal to $s=1$. 
	For each of the 500 repetitions used to estimate the power of a test, we sample uniformly a new value of the parameter $\theta\in\{-1,1\}^{P^d}$ for the perturbed uniform density.

	\begin{figure}[t]
		\centering
		  \includegraphics[width=\textwidth]{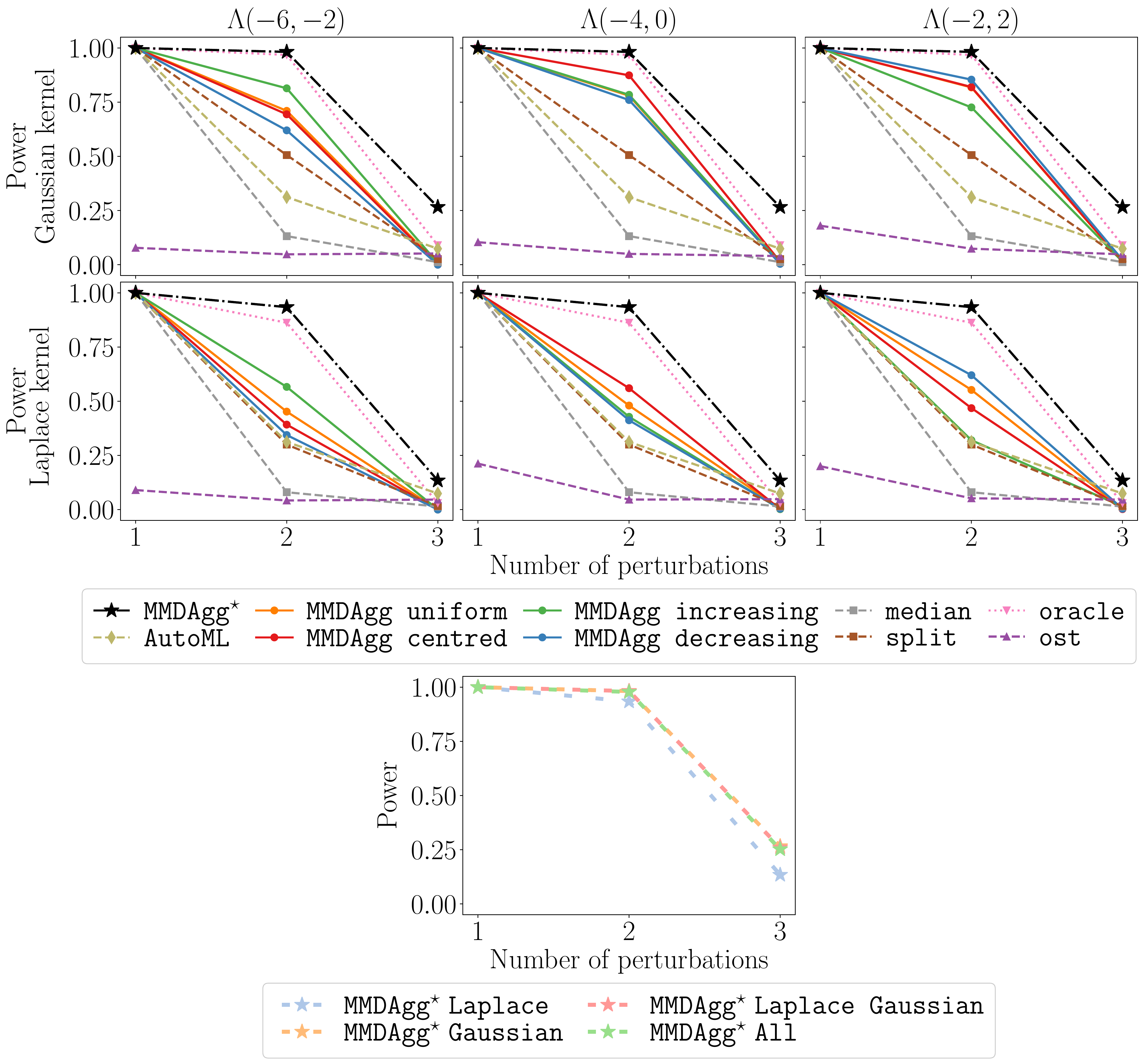}
		\captionsetup{format=hang}
		\caption{Power experiments with 2-dimensional perturbed uniform distributions using sample sizes $m=n=2000$ with a wild bootstrap.}
		\label{fig:perturbed_uniform_2d}
	\end{figure}

	In \Cref{fig:perturbed_uniform_1d}, we consider testing $n=500$ samples drawn from the $1$-dimensional uniform distribution against $m=500$ samples drawn from a $1$-dimensional uniform distribution with $P=1,2,3,4$ perturbations.
	We consider the same setting but in two dimensions with sample sizes $m=n=2000$ with up to three perturbations 
	in \Cref{fig:perturbed_uniform_2d}.
	For the MMDAgg tests of \Cref{weighting strategies} and for the \ost{} test, we use the collections of bandwidths $\Lambda(-6,-2)$, $\Lambda(-4,0)$ and $\Lambda(-2,2)$ as defined in \Cref{collection}.
	As the number of perturbations increases, it becomes harder to distinguish the two distributions, this translates into a decrease in power for all the tests.
	Even though we consider more samples for the 2-dimensional case, the performance of all the tests degrades significantly with dimension as detecting the perturbations becomes considerably more challenging.

	For all the settings considered in \Cref{fig:perturbed_uniform_1d,fig:perturbed_uniform_2d}, we observe that \mmdagg{} always performs the best, with power slightly higher than the one of \ora{} which has access to extra data to select an optimal bandwidth.
	All other tests achieve significantly lower test power.
	This validates our theoretical results that our aggregated MMDAgg test is minimax optimal in settings such as the one considered in this experiment where the difference in densities lies in a Sobolev ball of unknown smoothness.
	In particular, with its adaptive collection of bandwidths, \mmdagg{} obtains higher power than the four other aggregated tests with specifically chosen collections.
	In the 1-dimensional case, \automl{} performs similarly to our four aggregated tests, while in the 2-dimensional case with two perturbations it obtains much lower power.
	Our four tests with different weighting strategies outperform the three other tests \med{}, \spl{} and \ost{} in most settings, and always at least match the power of the best of those three.
	The \ost{} test, which is restricted to using the linear-time MMD estimate, obtains very low power compared to all the other tests using the quadratic-time estimate. 

	In all the experiments in \Cref{fig:perturbed_uniform_1d,fig:perturbed_uniform_2d}, using the Gaussian rather than the Laplace kernel results in higher power for our four aggregated tests, the difference is small but notable for the $1$-dimensional case while it is large for the $2$-dimensional case.
	For \mmdagg{}, there is no difference in \Cref{fig:perturbed_uniform_1d} and a small one in \Cref{fig:perturbed_uniform_2d}, as can be seen in the bottom plots.
	In one dimension, the \med{} test performs significantly better when using the Laplace kernel and outperforms the \spl{} test, which is not the case in all the other settings.

	In the bottom plots of \Cref{fig:perturbed_uniform_1d,fig:perturbed_uniform_2d}, we observe that by aggregating over both Gaussian and Laplace kernels, \mmdagglg{} obtains the highest power achieved by either \mmdaggl{} or \mmdaggg{}. 
	When adding 10 other kernels to the collection (each with $10$ bandwidths), \mmdagga{} retains the same high power, we do not observe a cost in power for considering more kernels. 
	This is only possible due the way we perform the level correction in \Cref{agg}.

	We now discuss the relation between the four weighting strategies for MMDAgg.
	Recall from \Cref{agg} that a single MMD test with larger associated weight is viewed as more important than one with smaller associated weight in the aggregated procedure. 
	Recall from \Cref{weighting strategies} that \mmdu{} puts equal weights on every bandwidths, that \mmdc{} puts the highest weight on the bandwidth in the middle of the collection, and that \mmdi{} puts the highest weight on the biggest bandwidth while \mmdd{} puts it on the smallest bandwidth.
	This allows us to interpret our results. 
	
	First, let's consider the case of the collection of bandwidths $\Lambda(-6,-2)$ for both one and two dimensions. 
	We observe that \mmdi{} has the highest power and \mmdd{} the lowest of the four aggregated tests, this means that putting the highest weight on the biggest bandwidth performs the best while putting it on the smallest bandwidth performs the worst.
	We can deduce that the most important bandwidth in our collection is the biggest one, which suggests that we should consider a collection consisting of larger bandwidths, say $\Lambda(-4,0)$. 
	In this case, \mmdc{} now obtains the highest power of our four weighting strategies. 
	We can infer that the optimal bandwidth is close to the bandwidths in the middle of our collection.
	When considering a collection of even larger bandwidths $\Lambda(-2,2)$, we see the opposite trends to ones observed using $\Lambda(-6,-2)$; \mmdd{} and \mmdi{} are performing the best and worst of our four tests, respectively. 
	This suggests that a collection consisting of smaller bandwidths than $\Lambda(-2,2)$ might be more appropriate. 

	So, comparing our aggregated tests with different weighting strategies gives us some insights on whether the collection we have considered is appropriate, or consists of bandwidths which are either too small or too large.
	The uniform weighting strategy does not perform the best but it is more robust to changes in the collection of bandwidths than the other strategies. 
	Of course, in practice, if we have access to a limited amount of data, one cannot run a hypothesis test with some parameters, observe the results and then modify those parameters to run the test again. 
	Nonetheless, the interpretation of the results of our different weighting strategies remains an appealing feature of our tests.
	In practice, we recommend using the parameter-free test \mmdagglg{} with its collection of bandwidths chosen adaptively.

\subsection{Power experiments on the MNIST dataset}
	\label{exp power mnist}

	\begin{figure}[t]
		\centering
		\includegraphics[width=\textwidth]{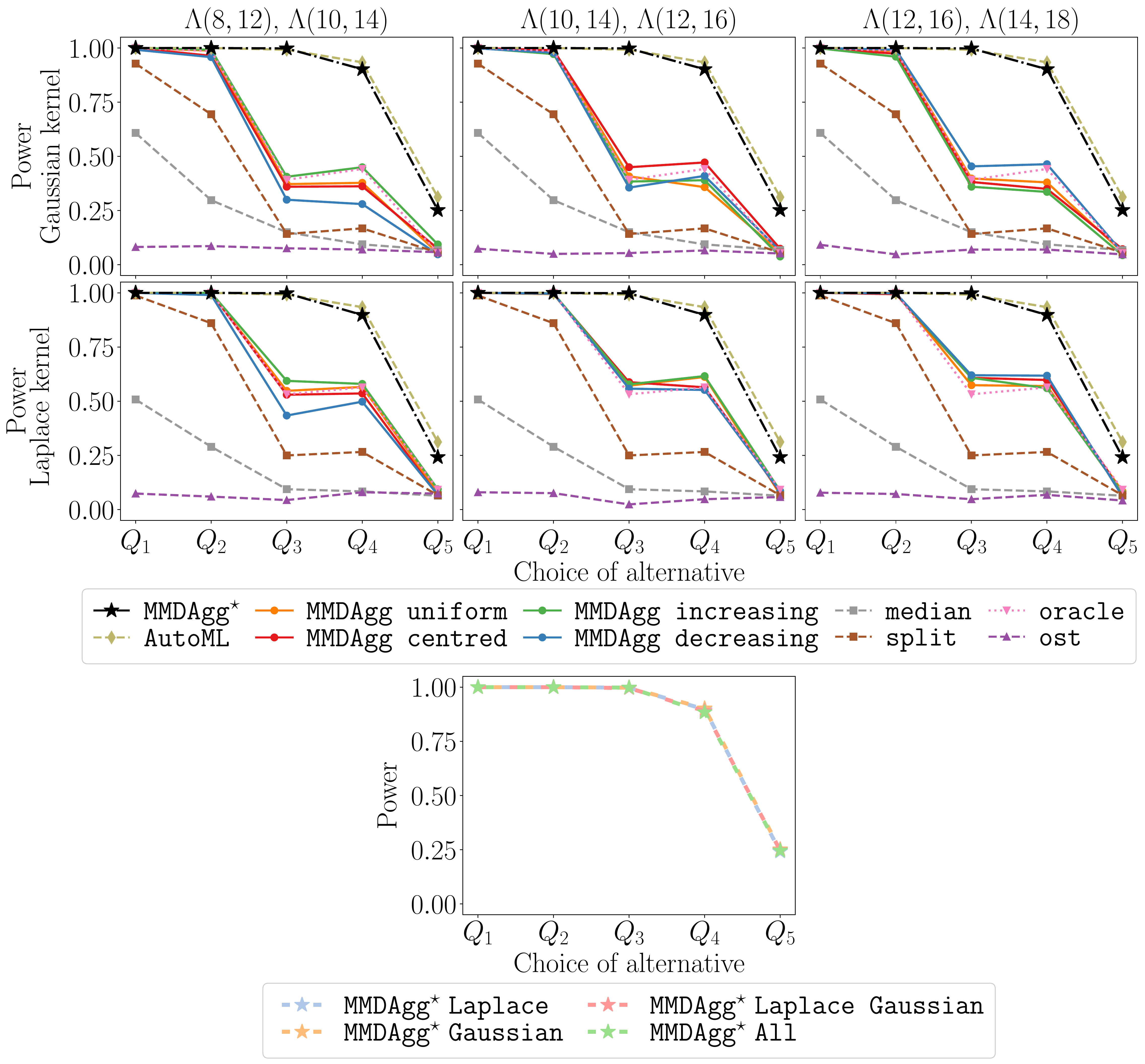}
		\captionsetup{format=hang}
		\caption{Power experiments with the MNIST dataset using sample sizes $m=n=500$ with a wild bootstrap.}
		\label{fig:mnist}
	\end{figure}

	Motivated by the experiment considered by \citet{kubler2020learning}, we consider the MNIST dataset \citep{lecun2010mnist} down-sampled to $7\times 7$ images. 
	In \Cref{fig:mnist}, we consider 500 samples drawn with replacement from the set $\mathcal{P}$ consisting all 70\,000 images of digits
	$$
		\mathcal{P}\colon\ 0,1,2,3,4,5,6,7,8,9.
	$$
	We test these against 500 samples drawn with replacement against one of the sets
	\begin{align*}
		Q_1\colon &1,3,5,7,9, \\
		Q_2\colon &0,1,3,5,7,9, \\
		Q_3\colon &0,1,2,3,5,7,9, \\
		Q_4\colon &0,1,2,3,4,5,7,9, \\
		Q_5\colon &0,1,2,3,4,5,6,7,9,
	\end{align*}
	of respective sizes 
	$35\,582$, 
	$42\,485$,
	$49\,475$,
	$56\,299$ and
	$63\,175$.
	While the samples are in dimension 49, distinguishing images of different digits reduces to a lower-dimensional problem.
	We consider the Gaussian kernel with the collections of bandwidths $\Lambda(8,12)$, $\Lambda(10,14)$ and $\Lambda(12,16)$ and the Laplace kernel with $\Lambda(10,14)$, $\Lambda(12,16)$ and $\Lambda(14,18)$.
	As $i$ increases, distinguishing $\mathcal{P}$ from $Q_i$ becomes a more challenging task, which results in a decrease in power.

	In the experiments presented in \Cref{fig:mnist} on image data, \mmdagg{} and \automl{} achieve the same power and outperform by far all the other tests. 
	This could be due to the fact that the collections for the other aggregated tests consist of very large bandwidths (from $2^8\lambda_{med}$ to $2^{18}\lambda_{med}$). 
	Nonetheless, we observe that \texttt{split} and \texttt{oracle}, which consider smaller bandwidths, also perform poorly compared to \mmdagg{} and \automl{}.

	The four aggregated tests with different weighting strategies often match, or even slightly beat, the performance of \ora{} which uses extra data to select an optimal bandwidth. 
	Moreover, they outperform significantly the two adaptive tests \spl{} and \ost{}, as well as the \med{} test.

	For \mmdagg{}, using either the Laplace or Gaussian kernel leads to the same performance in this experiment. 
	Furthermore, \mmdagg{} retains its high power when considering many more kernels as well, as can be seen in the bottom figure of \Cref{fig:mnist} with \mmdagga{}.
	Contrary to the previous experiments, we observe that using the Laplace kernel rather than the Gaussian one results in substantially higher power for the four aggregated tests with different weighting strategies.
	We recall that in the experiments of \Cref{fig:perturbed_uniform_1d,fig:perturbed_uniform_2d}, we observed that using a Gaussian kernel leads to higher power.
	This illustrates that the optimal choice of kernel varies depending on the type of data, as such in practice we recommend using \mmdagglg{} since it is observed to achieve the highest power obtained by either \mmdaggl{} or \mmdaggg{} in those three experiments. 
	The test \mmdagga{} also obtains the same power, but as it considers 12 types of kernels, each with 10 bandwidths, it is computationally more expensive.

	The pattern we observed in \Cref{fig:perturbed_uniform_1d,fig:perturbed_uniform_2d} of having \mmdi{} and \mmdd{} obtaining the highest and lowest power of our four aggregated tests for the collections of smaller and larger bandwidths, respectively, still holds to some extent in \Cref{fig:mnist} but the differences are less significant. 
	For the collection of bandwidths $\Lambda(12,16)$ with the Laplace kernel, \mmdc{} does not perform the best, it obtains slightly less power than \mmdu{} and \mmdi{} which have almost equal power. 
	Following our interpretation, this simply means that, while $\Lambda(12,16)$ is an appropriate choice of collection, the optimal bandwidth might be slightly larger than $2^{14}\lambda_{med}$.

	Note that, except \mmdagg{} and \automl{}, each test obtains similar power when trying to distinguish $\mathcal{P}$ from either $Q_3$ or $Q_4$. 
	Recall that $Q_3$ consists of images of all the digits except 4, 6 and 8 while $Q_4$ consists of images of all of them except 6 and 8.
	One possible explanation could be that these tests distinguish $\mathcal{P}$ from $Q_3$ mainly by detecting if images of the digit 6 appear in the sample, this would explain why we observe similar power for $Q_3$ and $Q_4$, and why the power for	$Q_5$ (consisting of every digit except 8) drops significantly.

	We also sometimes observe in \Cref{fig:mnist} that our aggregated tests with different weighting strategies obtain slightly higher power for $Q_4$ than for $Q_3$, which might at first seem counter-intuitive. 
	This could be explained by the fact that the optimal bandwidths for distinguishing $\mathcal{P}$ from $Q_3$ and from $Q_4$ might be very different, and that the choice of collections of bandwidths presented in \Cref{fig:mnist} are slightly better suited for distinguishing $\mathcal{P}$ from $Q_4$ than from $Q_3$.
	While it is also the case in \Cref{fig:perturbed_uniform_1d,fig:perturbed_uniform_2d} that the alternatives with different number of perturbations require different bandwidths to be detected, it looks like in that case considering a collection of five bandwidths which are powers of 2 is enough to adapt to those differences.
	For the MNIST experiment in \Cref{fig:mnist}, it seems that the differences between the optimal bandwidths for $Q_3$ and $Q_4$ are more important.
	Using \mmdagg{} with its adaptive parameter-free collection of bandwidths solves this problem.
	An advantage of our aggregated MMDAgg tests is that, even if we fix the collection of bandwidths, they are able to detect differences at various lengthscales, this is not the case for the \med{} and \spl{} tests as those select some specific bandwidth and are only able to detect the differences at the corresponding lengthscale.

\subsection{Power experiment: continuous limit of the collection of bandwidths}
	\label{continuous limit}

	\begin{figure}[!b]
		\centering
		\includegraphics[width=\textwidth]{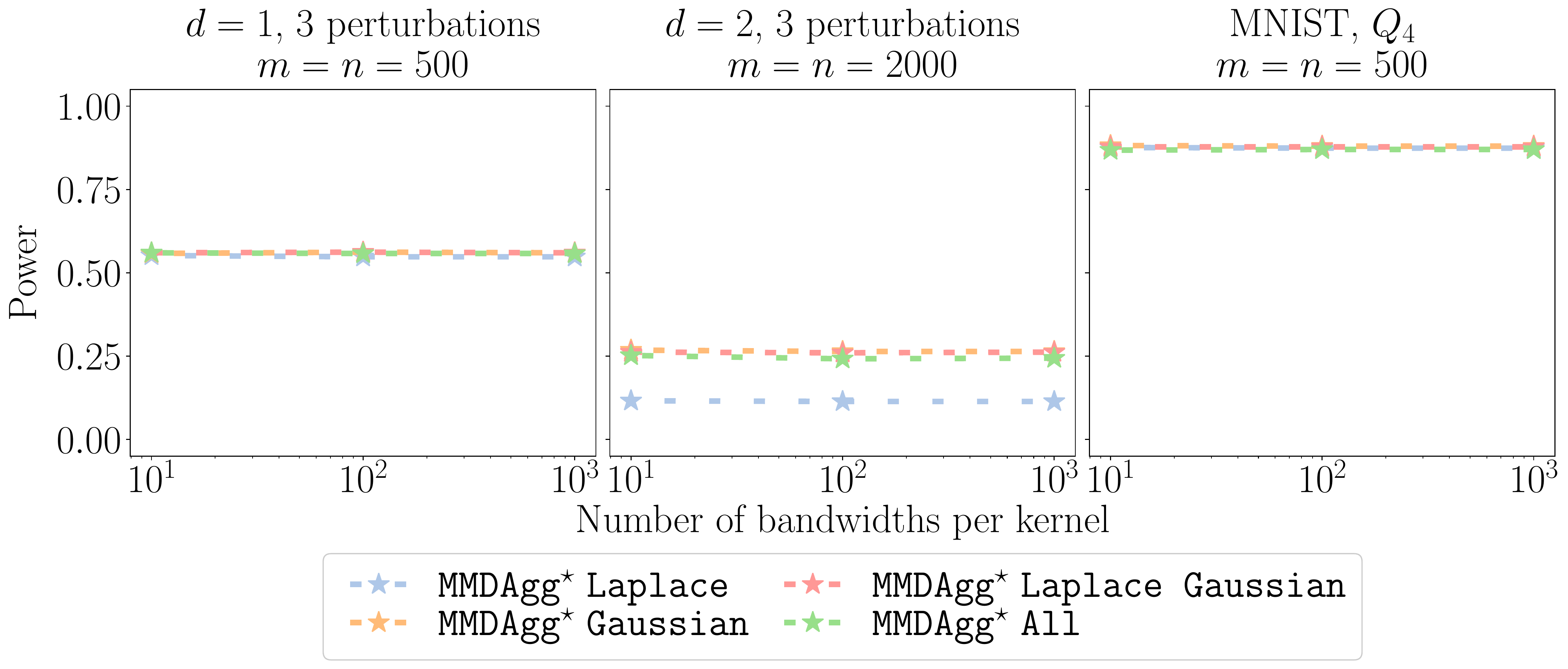}
		\captionsetup{format=hang}
		\caption{Power experiments varying the size of the collections of bandwidths using perturbed uniform $d$-dimensional distributions and the MNIST dataset with a wild bootstrap. 
		For \mmdagglg{}, two kernels are aggregated, each with a varying number of bandwidths.
		For \mmdagga{}, 12 kernels are considered with a varying number of bandwidths (up to 12000 kernels are aggregated).
		}
		\label{fig:change}
	\end{figure}

	Our collection of bandwidths for \mmdagg{} is a discretisation of an interval using $N=10$ points, which we formally introduced in \Cref{collection bandwidths}.
	As we increase the number of points $N$, the discretisation becomes finer and in the limit as $N\to\infty$ it corresponds to the whole continuous interval.
	In \Cref{fig:change}, we consider the three experiments of \Cref{fig:perturbed_uniform_1d,fig:perturbed_uniform_2d,fig:mnist} presented in \Cref{exp power synthetic,exp power mnist} for \mmdaggg{}, \mmdaggl{}, \mmdagglg{}, and \mmdagga{}, with collections of sizes $N$, $N$, $2N$ and $12N$, respectively (details in \Cref{experimental details}).
	We vary the number of bandwidths per kernel $N$ to be 10, 100 and 1000.
	In particular, this means for example that \mmdagga{} with $N=1000$ aggregates over $12N=12000$ kernels.
	For the one-dimensional uniform setting, we use three perturbations and sample sizes 500.
	The 2-dimensional case is considered with three perturbations and $m=n=2000$.
	For the MNIST experiment, we use $Q_4$ as an alternative (every digit except 8 and 6) with sample sizes 500.

	As in the previous experiments, in \Cref{fig:change}, the MMDAgg test aggregating both Laplace and Gaussian kernels obtains the highest power achieved by either \mmdaggl{} or \mmdaggg{}.
	Considering more types of kernels with \mmdagga{} does not change the test power.
	Since it is computationally more expensive, we recommend using \mmdagglg{} in practice.

	In \Cref{fig:change}, we observe for all four tests that the test power remains the same when increasing the number of bandwidths per kernel from 10 to 1000.
	The case $N=1000$ simulates the continuous limit of the collection of bandwidths, that is, when the full interval is considered without discretisation.
	First, this shows that our MMDAgg test retains high power even when aggregating up to 12000 kernels, which is only possible due to the way we perform the level correction.
	Second, this illustrates that the power of the aggregated test with a continuous collection of bandwidths, can also be achieved by the less computationally expensive MMDAgg test with a discretisation of $N=10$ points.

\subsection{Power experiment for image shift detection}
	\label{shift experiment}

	\begin{table}[!t]
	\captionsetup{format=hang}
    \caption{Image shift detection experiment of \citet{rabanser2019failing}. The numbers reported correspond to the test power averaged over 60 alternatives (each repeated 5 times): 10 different shift types applied on either 10\%, 50\% or 100\% of the image samples drawn from either the MNIST or CIFAR-10 datasets.}
    \label{tab:shift_results}
	\centering
	\setlength\tabcolsep{5pt}
	\begin{tabular}{cccccccccc}
	\toprule

    \multicolumn{2}{c}{\multirow{2}{*}[-2px]{Test}} & \multicolumn{8}{c}{Number of samples} \\
    \cmidrule(r){3-10}
    & & 10 & 20 & 50 & 100 & 200 & 500 & 1000 & 10000\\
    \midrule

    \multicolumn{2}{c}{\mmdaggl} & 0.21 & 0.29 & 0.40 & 0.44 & 0.47 & 0.56 & 0.67 & 0.83 \\
    \multicolumn{2}{c}{\mmdaggg} & 0.19 & 0.26 & 0.34 & 0.42 & 0.42 & 0.51 & 0.62 & 0.75 \\
    \multicolumn{2}{c}{\mmdagglg} & 0.21 & 0.27 & 0.37 & 0.43 & 0.45 & 0.54 & 0.65 & 0.80 \\
    \multicolumn{2}{c}{\mmdagga} & 0.21 & 0.27 & 0.37 & 0.43 & 0.46 & 0.55 & 0.66 & 0.80 \\
    \midrule

    \multicolumn{2}{c}{\texttt{MMDAgg uniform (Laplace)}} & 0.20 & 0.28 & 0.40 & 0.43 & 0.46 & 0.52 & 0.58 & 0.79 \\
    \multicolumn{2}{c}{\texttt{MMDAgg uniform (Gaussian)}} & 0.15 & 0.23 & 0.33 & 0.35 & 0.38 & 0.44 & 0.48 & 0.69 \\
    \midrule

    \multicolumn{2}{c}{\texttt{AutoML (raw)}} & 0.17&    0.24    &0.37   &0.46   &0.50   &0.62   &0.67   &0.87\\
    \multicolumn{2}{c}{\texttt{AutoML (pre)}} &0.18  &0.29  &0.42   &0.47  &0.47   &0.64   &0.65   &0.72\\
    \multicolumn{2}{c}{\texttt{AutoML (class)}}&0.19 &0.19   &0.38   &0.46   &0.52  &0.61   &0.67   &0.87\\
    \multicolumn{2}{c}{\texttt{AutoML (bin)}}&0.03   &0.14   &0.31   &0.43   &0.49   &0.51   &0.59   &0.86\\

    \bottomrule
    \end{tabular}
	\end{table}

	In this section, we consider the experiment of \citet[Table 1a]{rabanser2019failing} on image shift detection on the MNIST \citep{lecun2010mnist} and CIFAR-10 \citep{krizhevsky2009learning} datasets.
	Ten different types of shifts are applied to either 10\%, 50\% or 100\% of the samples, those include 
	adversarial shifts, 
	class knock-outs, 
	injecting Gaussian noise, 
	and combining rotations/translations/zoom-ins,
	with different shift strengths (see Section 4 of \citealt{rabanser2019failing} for details).

	We report in \Cref{tab:shift_results} the test power obtained by our four \mmdagg{} tests, by our \texttt{MMDAgg uniform} test with Laplace kernels and with Gaussian kernels, as well as by four versions of the \texttt{AutoML} test of \citet[Table 1a]{kubler2022automl}. 
	We observe that the \mmdagg{} tests outperform the \texttt{MMDAgg uniform} tests. 
	For \mmdagg{}, using the Laplace kernel performs the best, aggregating both Laplace and Gaussian kernels performs better than using only Gaussian kernels and almost as well as using only Laplace kernels, aggregating many more kernels with \mmdagga{} results in the same power obtained by \mmdagglg{}.

	Despite using off-the-shelf kernels which are not specifically designed for images, our aggregated \mmdagg{} tests are still performing within only a few percentage points of state-of-the-art tests based on training models (\emph{e.g.} neural networks) which excel on image data, such as the \texttt{AutoML} test.

\subsection{Overview of additional experimental results}
	\label{results additional experiments}

	We consider additional experiments in \Cref{experiments appendix}, we briefly summarize them here.
	In \Cref{exp level}, we verify that all the tests we consider have well-calibrated levels.
	We then consider widening the collection the bandwidths for our tests \mmdu{} and \mmdc{} in \Cref{widen}, and observe that the associated cost in the power is relatively small.
	We verify in \Cref{comparison permutations wild bootstrap} that using a wild bootstrap or permutations results in similar performance; the difference is of non-significant order and is not biased towards one or the other. 
	In \Cref{different sample sizes}, we show that if one of the sample sizes is fixed to a small number, we cannot obtain high power even if we take the other sample size to be very large.
	Finally, we increase the sample sizes for the \ost{} test in \Cref{ost increase} and observe that we need extremely large sample sizes to match the performance of \mmdu{} with 500 or 2000 samples.


\section{Conclusion and future work}
\label{conclusion}

We have constructed a two-sample hypothesis test, called MMDAgg, which aggregates multiple MMD tests using different kernels/bandwidths.
Our test is adaptive over Sobolev balls and does not require data splitting.
We have proved that MMDAgg is optimal in the minimax sense over Sobolev balls up to an iterated logarithmic term, for any product of one-dimensional translation invariant characteristic kernels which are absolutely and square integrable.
This optimality result also holds under two popular strategies used in estimating the test thresholds, namely the wild bootstrap and  permutation procedures.
In practice, we propose four weighting strategies which allow the user to incorporate prior knowledge about the collection of bandwidths.
We also introduce a parameter-free adaptive collection of bandwidths, which we recommend using in practice while aggregating over both Gaussian and Laplace kernels, each with multiple bandwidths.
This adaptive collection is the discretisation of an interval and in practice we found that using ten bandwidths per kernel performs as well as using the whole continuous interval in the limit.
Our MMDAgg test obtains significantly higher power than other state-of-the-art MMD-based two-sample tests in synthetic settings where the smoothness Sobolev assumption is satisfied, this empirically validates our theoretical result of minimax optimality and adaptivity over Sobolev balls.
In experiments on image data, we observe that MMDAgg almost matches the power of much more complex two-sample tests relying on training models, such as  neural networks, to detect the difference in distributions.

We now discuss three research directions based on this current work, two of which have been explored by \citet{schrab2022ksd,schrab2022efficient}.

First, it would be interesting to consider the two-sample kernel-based test of \citet{jitkrittum2016interpretable}, who use adaptive features (in the data space or in  the  Fourier domain) to construct a linear-time test with good test power. 
\citet{jitkrittum2016interpretable} require setting aside part of the data to select the kernel bandwidths and the feature locations, by maximizing a proxy for test power. 
They then perform the test on the remaining  data.  
It would be of interest to develop an approach to learning such adaptive interpretable features without data splitting.
Adapting the current results of this work to that setting remain an open challenge.

Second, aggregated tests that  are adaptive over Sobolev balls have been constructed for several alternative testing scenarios.
The independence testing problem using the Hilbert Schmidt Independence Criterion has been treated by \citet{albert2019adaptive}, which is related to the Maximum Mean Discrepancy \citep[Section 7.4]{gretton2012kernel}.
A further setting of interest is goodness-of-fit testing, where a sample is compared against a model.
Our theoretical results can directly be applied to  goodness-of-fit testing using the MMD, as long as the expectation of the kernel under the model can be computed in closed form.
A more challenging problem arises when this expectation cannot be easily computed. 
In this case, a test may be constructed based on the Kernelised Stein Discrepancy (KSD---\citealp{liu2016kernelized,chwialkowski2016kernel}). 
This corresponds to computing a Maximum Mean Discrepancy in a modified Reproducing Kernel Hilbert Space, consisting of functions which have zero expectation under the model.
Building on the present paper, \citet{schrab2022ksd} develop an adaptive aggregated KSDAgg test of goodness-of-fit for the KSD and provide conditions which guarantee high test power for KSDAgg.

Third, our MMDAgg test proposed in this work, the KSDAgg test of \citet{schrab2022ksd}, and the aggregated HSIC test of \citet{albert2019adaptive}, are all quadratic-time hypothesis tests. 
While quadratic-time tests usually achieve higher power than linear-time tests, this comes at the expense of an important computational cost. 
To tackle this problem, relying on incomplete $U$-statistics, \citet{schrab2022efficient} propose efficient variants (including linear-time ones) of those three aggregated tests, called \mbox{MMDAggInc}, \mbox{KSDAggInc} and \mbox{HSICAggInc}.
They theoretically quantify the cost incurred in the minimax rate over Sobolev balls for this improvement in computational efficiency.

\FloatBarrier


\section*{Acknowledgements}
\label{acknowledgements}
We would like to thank the action editor Ingo Steinwart and the anonymous referees for their thorough reviews and suggestions which have helped to significantly improve the paper.
Antonin Schrab acknowledges support from the U.K.\ Research and Innovation under grant number EP/S021566/1.
Ilmun Kim acknowledges support from the Yonsei University Research Fund of 2022-22-0289 as well as support from the Basic Science Research Program through the National Research Foundation of Korea (NRF) funded by the Ministry of Education (2022R1A4A1033384), and the Korea government (MSIT) RS-2023-00211073.
B{\'e}atrice Laurent acknowledges the funding by ANITI ANR-19-PI3A-0004.
Benjamin Guedj acknowledges partial support by the U.S.\ Army Research Laboratory and the U.S.\ Army Research Office, and by the U.K.\ Ministry of Defence and the U.K.\ Engineering and Physical Sciences Research Council (EPSRC) under grant number EP/R013616/1; Benjamin Guedj also acknowledges partial support from the French National Agency for Research, grants ANR-18-CE40-0016-01 and ANR-18-CE23-0015-02.
Arthur Gretton acknowledges support from the Gatsby Charitable Foundation.

\appendix

\section*{Overview of Appendices}

In \Cref{experiments appendix}, we present results of additional experiments.
In \Cref{relation section}, we explain the relation between using permutations and using the wild bootstrap.
We present an efficient implementation of MMDAgg in \Cref{efficientMMDAgg}.
We highlight the proof strategy of deriving the minimax rate over a Sobolev ball in \Cref{appendix lower}.
Finally, we present the proofs of all our results in \Cref{proofs}.


\section{Additional experiments}
\label{experiments appendix}

In this section, we verify the level achieved by all the tests considered (\Cref{exp level}), and run multiple power experiments: widening the collection of bandwidths (\Cref{widen}), comparing wild bootstrap and permutations (\Cref{comparison permutations wild bootstrap}), using unbalanced sample sizes (\Cref{different sample sizes}), and increasing the sample sizes for the \ost{} test (\Cref{ost increase}).

\subsection{Level experiments} 
    \label{exp level}

    \begin{table}[h]
        \captionsetup{format=hang}
        \caption{Level experiments with samples drawn either from $d$-dimensional uniform distributions or from the MNIST dataset using the Gaussian (G.) and Laplace (L.) kernels with either a wild bootstrap (w.b.) or permutations (p.).}
        \label{tab:levels}
        \begin{center}
            \begin{tabular}{| m{0.09cm} | m{0.09cm} | m{0.39cm} || c | c | c | c | c| c | c |} 
                \cline{4-10}
                 \multicolumn{1}{c}{} & \multicolumn{1}{c}{} & \multicolumn{1}{c|}{} & \multirow{2}{*}{\shortstack[t]{\texttt{MMDAgg} \\ \texttt{uniform}}} & \multirow{2}{*}{\shortstack[t]{\texttt{MMDAgg} \\ \texttt{centred}}} & \multirow{2}{*}{\shortstack[t]{\texttt{MMDAgg} \\ \texttt{increasing}}} & \multirow{2}{*}{\shortstack[t]{\texttt{MMDAgg} \\ \texttt{decreasing}}} & \multirow{2}{*}{\shortstack[t]{\med{}}} & \multirow{2}{*}{\shortstack[t]{\spl{}}} & \multirow{2}{*}{\shortstack[t]{\ost{}}}  \\ [0.4cm] 
                \hhline{---;:=======}
                
                \parbox[t]{2mm}{\multirow{4}{*}{\hspace{-1mm}\rotatebox[origin=c]{90}{$d=1$}}} & \multirow{2}{*}{\hspace{-1.4mm}G.} & \multirow{1}{*}{\!\!w.b.} &  0.0476 & 0.052 & 0.0456 & 0.0434 & 0.047 & 0.054 & 0.0594 \\ 
                 \hhline{|~|~|-||-------}
                 & & \multirow{1}{*}{\,p.} & 0.0496 & 0.0532 & 0.0478 & 0.0454 & 0.0468 & 0.0528 & 0.0594 \\ 
                 \hhline{|~|-|-||-------}
                  & \multirow{2}{*}{\hspace{-1mm}L.} & \multirow{1}{*}{\!\!w.b.} & 0.0474 & 0.0488 & 0.0516 & 0.0504 & 0.0534 & 0.05 & 0.0586 \\ 
                 \hhline{|~|~|-||-------}
                 & & \multirow{1}{*}{\,p.}  & 0.047 & 0.0482 & 0.0496 & 0.0494 & 0.0522 & 0.0494 & 0.0586 \\ 
                \hhline{===::=======}
                
                \parbox[t]{2mm}{\multirow{4}{*}{\hspace{-1mm}\rotatebox[origin=c]{90}{$d=2$}}} & \multirow{2}{*}{\hspace{-1.4mm}G.} & \multirow{1}{*}{\!\!w.b.} & 0.039 & 0.0432 & 0.044 & 0.0496 & 0.0464 & 0.0482 & 0.0478 \\ 
                 \hhline{|~|~|-||-------}
                 & & \multirow{1}{*}{\,p.} & 0.0424 & 0.0446 & 0.0414 & 0.0498 & 0.0466 & 0.0472 & 0.0478 \\ 
                 \hhline{|~|-|-||-------}
                  & \multirow{2}{*}{\hspace{-1mm}L.} & \multirow{1}{*}{\!\!w.b.} & 0.0382 & 0.0502 & 0.0506 & 0.0478 & 0.0438 & 0.0548 & 0.0502 \\ 
                 \hhline{|~|~|-||-------}
                 & & \multirow{1}{*}{\,p.}  & 0.0418 & 0.0474 & 0.0514 & 0.049 & 0.0458 & 0.0548 & 0.0502 \\ 
                \hhline{===::=======}

                \parbox[t]{2mm}{\multirow{4}{*}{\hspace{-1mm}\rotatebox[origin=c]{90}{MNIST}}} & \multirow{2}{*}{\hspace{-1.4mm}G.} & \multirow{1}{*}{\!\!w.b.} & 0.0478 & 0.0528 & 0.0474 & 0.0488 & 0.0526 & 0.0498 & 0.0496 \\ 
                 \hhline{|~|~|-||-------}
                 & & \multirow{1}{*}{\,p.} & 0.042 & 0.05 & 0.0476 & 0.048 & 0.055 & 0.0484 & 0.0496 \\ 
                 \hhline{|~|-|-||-------}
                  & \multirow{2}{*}{\hspace{-1mm}L.} & \multirow{1}{*}{\!\!w.b.} & 0.054 & 0.052 & 0.0424 & 0.0548 & 0.0518 & 0.0444 & 0.05  \\ 
                 \hhline{|~|~|-||-------}
                 & & \multirow{1}{*}{\,p.} & 0.0526 & 0.0532 & 0.0442 & 0.0554 & 0.051 & 0.0448 & 0.05 \\ 
                 \hhline{---||-------}
            \end{tabular}
        \end{center}
        \begin{center}
            \begin{tabular}{| m{1cm} | m{0.39cm} || c | c | c | c | c| c | c | c |}
                \cline{3-9}
                 \multicolumn{1}{c}{} & \multicolumn{1}{c|}{} & \multirow{2}{*}{\shortstack[t]{\texttt{MMDAgg}$^\star$ \\ \texttt{Laplace}}} & \multirow{2}{*}{\shortstack[t]{\texttt{MMDAgg}$^\star$ \\ \texttt{Gaussian}}} & \multirow{2}{*}{\shortstack[t]{\texttt{MMDAgg}$^\star$ \\ \texttt{Laplace \& Gaussian}}} & \multirow{2}{*}{\shortstack[t]{\texttt{MMDAgg}$^\star$ \\ \texttt{All}}}  & \multirow{2}{*}{\shortstack[t]{$\ $ \\ \automl{}}}\\ [0.4cm] 
                \hhline{--;:=====}
                
                \multirow{2}{*}{\vspace{-0.025cm}\hspace{0.1cm}$d=1$} & \multirow{1}{*}{\!\!w.b.} &  0.0506 & 0.05 & 0.0496 & 0.0492 & \multirow{2}{*}{\shortstack[t]{$\ $ \\ 0.0462}} \\
                 \hhline{|~|-||----}
                 & \multirow{1}{*}{\,p.} & 0.0528 & 0.0492 & 0.0496 & 0.0494 & \\ 
                \hhline{==::=====}
                
                \multirow{2}{*}{\vspace{-0.025cm}\hspace{0.1cm}$d=2$} & \multirow{1}{*}{\!\!w.b.} & 0.0458 & 0.0428 & 0.0434 & 0.043 & \multirow{2}{*}{\shortstack[t]{$\ $ \\ 0.051}}\\ 
                 \hhline{|~|-||----}
                 & \multirow{1}{*}{\,p.} & 0.0456 & 0.0438 & 0.0438 & 0.0434 & \\ 
                \hhline{==::=====}

                \multirow{2}{*}{\vspace{-0.05cm}\hspace{-0.13cm}MNIST} & \multirow{1}{*}{\!\!w.b.} & 0.0624 & 0.0624 & 0.0632 & 0.0594 & \multirow{2}{*}{\shortstack[t]{$\ $ \\ 0.518}}\\ 
                 \hhline{|~|-||----}
                 & \multirow{1}{*}{\,p.} & 0.0612 & 0.062 & 0.0628 & 0.0622 & \\ 
                 \hhline{--||-------}
            \end{tabular}
        \end{center}
    \end{table}

    In \Cref{tab:levels}, we empirically verify that all the tests we consider have the desired level $\alpha=0.05$ in the three different settings considered in \Cref{fig:perturbed_uniform_1d,fig:perturbed_uniform_2d,fig:mnist}.
    For the aggregated MMDAgg tests of \Cref{weighting strategies}, we use the collection of bandwidths $\Lambda(-4,0)$ for samples drawn from a uniform distribution in one and two dimensions.
    For samples drawn from the set $\mathcal{P}$ of images of all MNIST digits, we use $\Lambda(10,14)$ and $\Lambda(12,16)$ for the Gaussian and Laplace kernels, respectively.
    To obtain more precise results, we use 5000 repetitions to estimate the levels.

    We observe in \Cref{tab:levels} that all the tests have well-calibrated levels, indeed all the estimated levels are relatively close to the prescribed level $0.05$.
    We consider three different types of data and run the tests with the Gaussian and Laplace kernel using either a wild bootstrap or permutations. 
    We note that there is no noticeable trend in the differences in the estimated levels across all those different settings.

\subsection{Power experiments: widening the collection of bandwidths}
    \label{widen}

    \begin{figure}[!t]
        \centering
        \includegraphics[width=\textwidth]{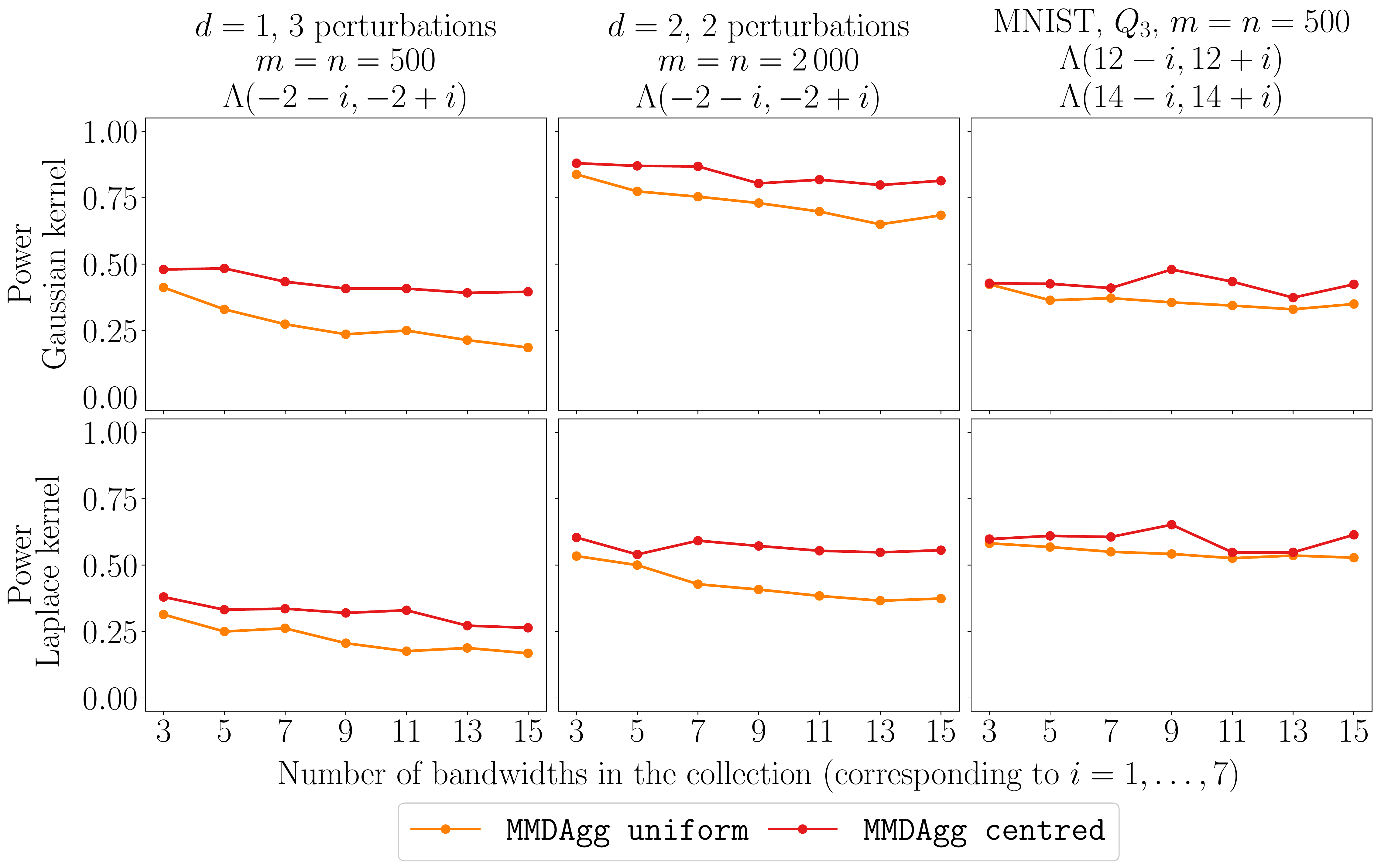}
        \captionsetup{format=hang}
        \caption{Power experiments varying the size of the collections of bandwidths using perturbed uniform $d$-dimensional distributions and the MNIST dataset with a wild bootstrap.}
        \label{fig:widen}
    \end{figure}

    In practice, we might not have strong prior knowledge to guide us in the choice of a collection consisting of only a few bandwidths. 
    For this reason, in practice, we recommend using the adaptive parameter-free collection introduced in \Cref{collection bandwidths} with both Laplace and Gaussian kernels.
    Nonetheless, it is interesting to study the properties of the family of collections of \Cref{weighting strategies} consisting of the median bandwidth scaled by powers of 2. 
    In \Cref{fig:widen}, we design an experiment where we start with a collection of 3 bandwidths chosen to be centred around the optimal bandwidth.
    We then widen the collection of bandwidths and observe how much the power deteriorates as more bandwidths are included in the collection.

    We consider collections ranging from 3 to 15 bandwidths for our two tests \mmdu{} and \mmdc{}, for the three types of data used in \Cref{fig:perturbed_uniform_1d,fig:perturbed_uniform_2d,fig:mnist}.
    For the perturbed uniform distributions in one and two dimensions, we use the collection of bandwidths $\Lambda(-2-i,-2+i)$ for $i=1,\dots,7$ for both kernels.
    For the MNIST dataset with the Gaussian and Laplace kernels, we use the collections of bandwidths
    $\Lambda(12-i,12+i)$ and $\Lambda(14-i,14+i)$ for $i=1,\dots,7$, respectively.
    Those collections are centred as those corresponding to the middle columns of \Cref{fig:perturbed_uniform_1d,fig:perturbed_uniform_2d,fig:mnist} (for the case $i=2$), so we expect the bandwidth in the centre of each collection to be a well-calibrated one.
    For this reason, it makes sense to consider only \mmdu{} and \mmdc{} in those experiments. 

    In all the settings considered in \Cref{fig:widen}, we observe only a very small decrease in power when considering a wider collection of bandwidths for \mmdc{}.
    This is due to the fact that even though we consider more bandwidths, we still put the highest weight on the well-calibrated one in the centre of the collection. 
    Nonetheless, the fact that almost no power is lost when considering more bandwidths for \mmdc{} is a great feature of our test, which is only possible due to the way we perform the level correction for MMDAgg. 
    For the MNIST dataset, we observe a slight increase in power for \mmdc{} with the collections of nine bandwidths for both kernels. 
    This could indicate that, as suggested in \Cref{exp power mnist}, the bandwidths in the centre of the collections are well-calibrated to distinguish $\mathcal{P}$ from $Q_4$ but are not necessarily the best choice to distinguish $\mathcal{P}$ from $Q_3$.
    
    Remarkably, the power for \mmdu{}, which puts equal weights on all the bandwidths, decays relatively slowly and this test does not use the information that
    the bandwidth in the centre of the collection is a well-calibrated one. 
    So, we expect similar results for any collections of the same sizes which include this bandwidth but not necessarily in the centre of the collection.
    This means that, in practice, without any prior knowledge, one can use uniform weights with a relatively wide collection of bandwidths without incurring a considerable loss in power.

    \begin{figure}[!t]
        \centering
        \includegraphics[width=\textwidth]{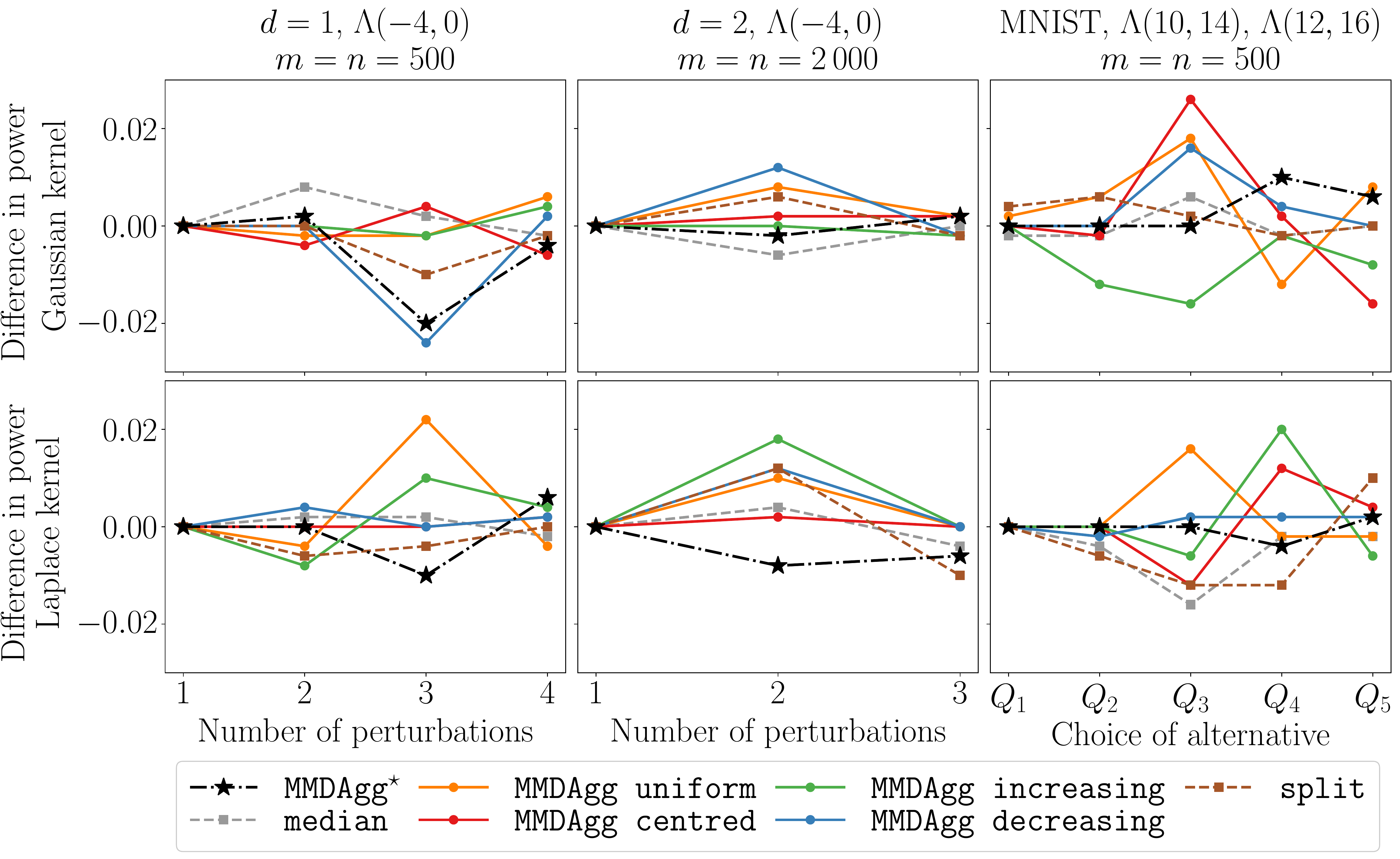}
        \captionsetup{format=hang}
        \caption{Power experiments considering the difference between using a wild bootstrap or permutations on perturbed uniform $d$-dimensional distributions and on the MNIST dataset.}
        \label{fig:wildperm}
    \end{figure}

\subsection{Power experiments: comparing wild bootstrap and permutations}
    \label{comparison permutations wild bootstrap}

    We consider the settings of the experiments presented in \Cref{fig:perturbed_uniform_1d,fig:perturbed_uniform_2d,fig:mnist} on synthetic and real-world data using the Gaussian and Laplace kernels. 
    We run the same experiments using permutations instead of a wild bootstrap for one collection of bandwidths for each of the different settings.
    We then consider the power obtained using a wild bootstrap minus the one obtained using permutations, and plot this difference in \Cref{fig:wildperm}.

    The absolute difference in power between using a wild bootstrap or permutations is minimal, it is at most roughly 0.02 and is even considerably smaller in most cases.
    Furthermore, the difference overall does not seem to be biased towards using either of the two procedures. 
    Since there is no significant difference in power, we suggest using a wild bootstrap when the sample sizes are the same since our implementation of it runs slightly faster in practice.
    Of course, when the sample sizes are different, one must use permutations.

\subsection{Power experiments: using unbalanced sample sizes}
    \label{different sample sizes}

\begin{figure}[!t]
    	\centering
    	\includegraphics[width=\textwidth]{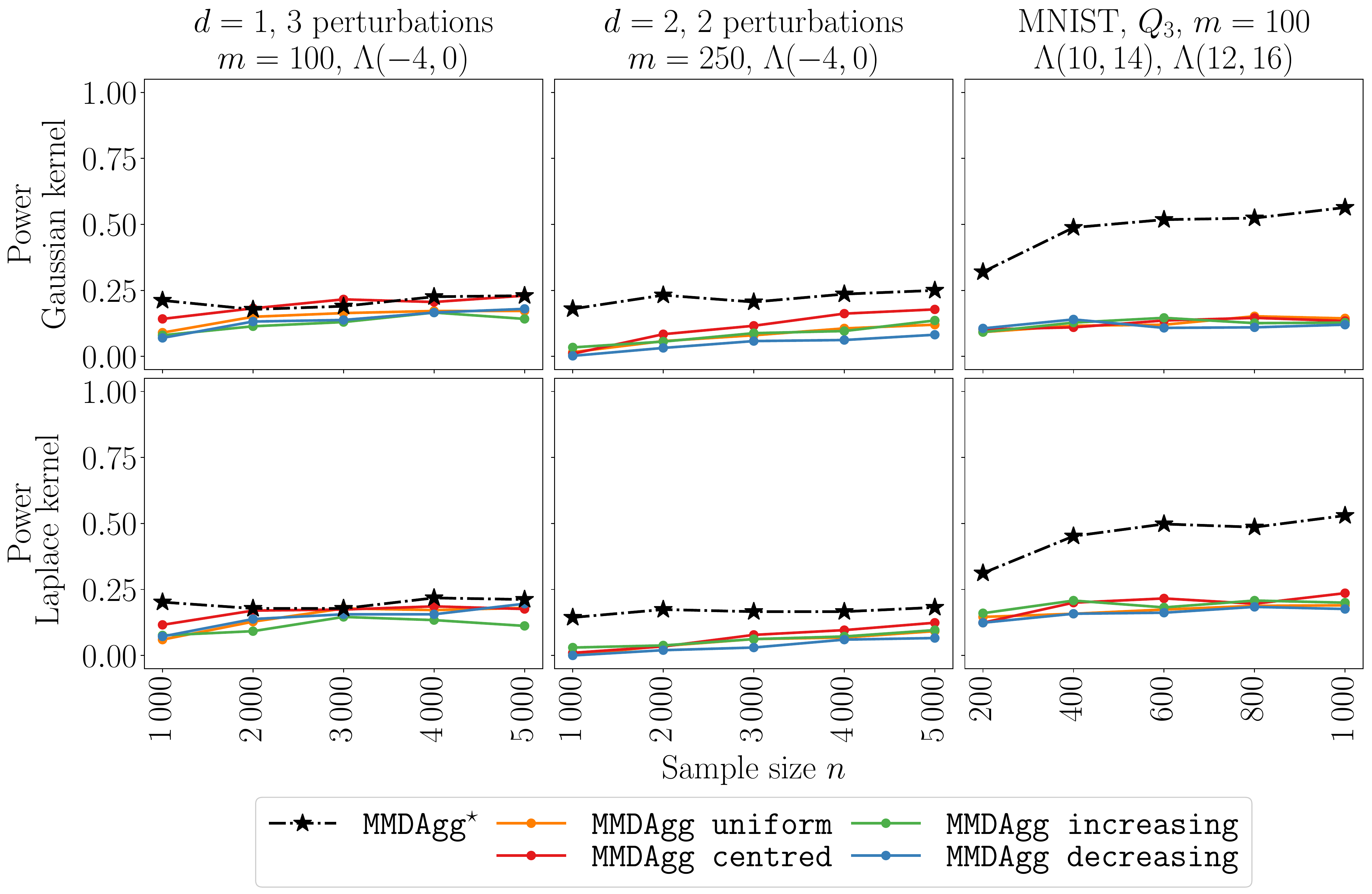}
        \captionsetup{format=hang}
    	\caption{Power experiments with different sample sizes $m\neq n$ on perturbed uniform $d$-dimensional distributions and on the MNIST dataset using permutations.}
        \label{fig:different}
    \end{figure}

    In \Cref{fig:different}, we consider fixing the sample size $m$ and increasing the size $n$ of the other sample, we use permutations since we work with different sample sizes.
    We consider the settings of \Cref{fig:perturbed_uniform_1d,fig:perturbed_uniform_2d,fig:mnist} with three and two perturbations for the uniform distributions in one and two dimensions, respectively.
    For the MNIST dataset, we use the set of images of all digits $\mathcal{P}$ against the set of images $Q_3$ which does not include the digits 4, 6 and 8.

    We observe the same patterns across the six experiments presented in \Cref{fig:different}. 
    When fixing one of the sample sizes to be small (100 or 250), we cannot achieve power higher than 0.25 (except for \mmdagg{} on the MNIST data, which as in \Cref{fig:mnist} performs much better than all other tests) by increasing the size of the other sample to be very large (up to 5000). 
    Indeed, we observe that a plateau is reached where considering an even larger sample size does not result in higher power. 
    In some sense, all the information provided by the small sample has already been extracted and using more points for the other sample has almost no effect.
    As shown in \Cref{fig:perturbed_uniform_1d,fig:perturbed_uniform_2d,fig:mnist}, we can obtain significantly higher power in all of those settings using samples of sizes $m=n=500$ or $m=n=2000$. 
    Having access to even more samples overall (5100 instead of 1000) but in such an unbalanced way results in very low power.
    This shows the importance of having, if possible, balanced datasets with sample sizes of the same order.   

\subsection{Power experiments: increasing sample sizes for the \ost{} test}
    \label{ost increase}

    \begin{figure}[!t]
        \centering
        \includegraphics[width=\textwidth]{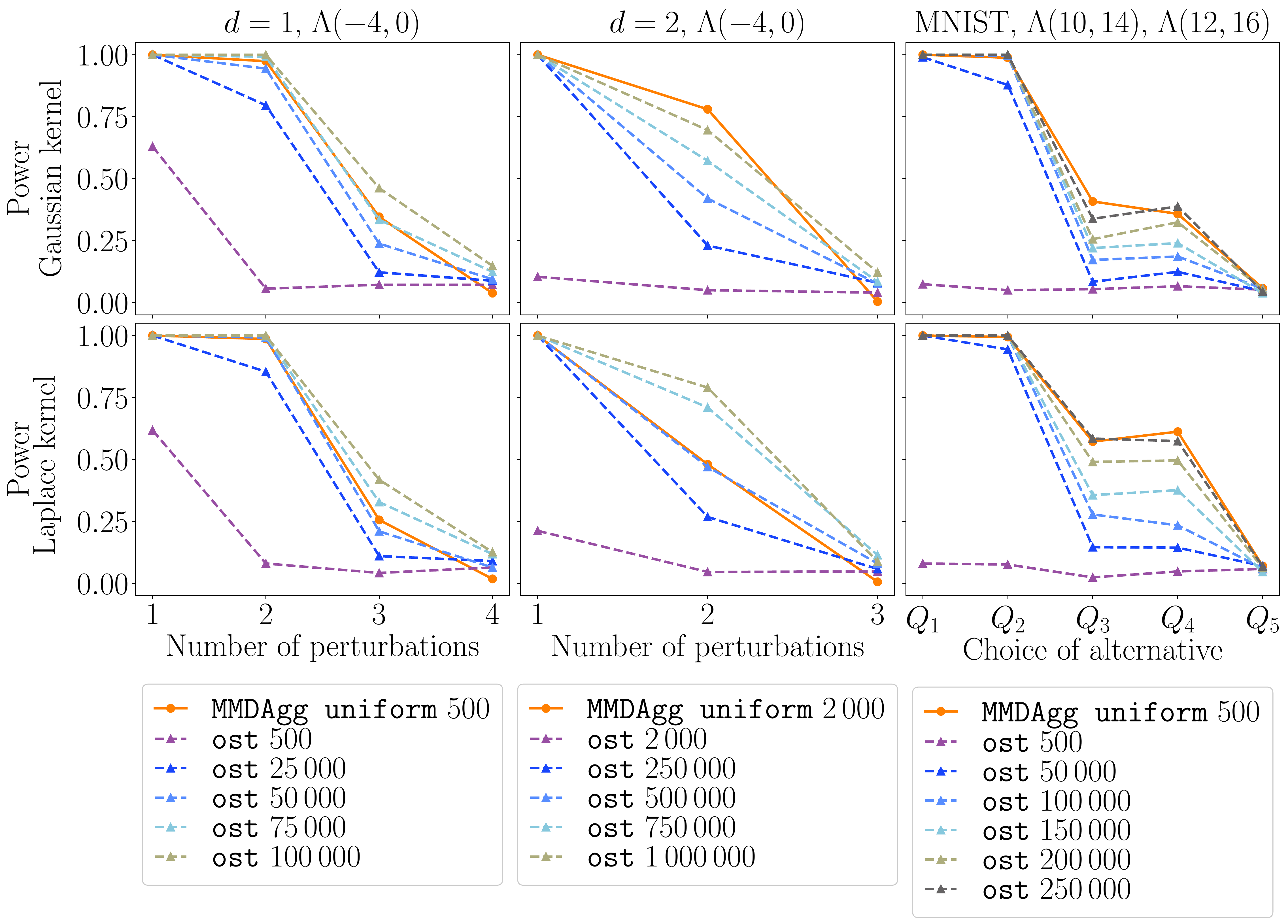}
        \captionsetup{format=hang}
        \caption{Power experiments increasing the sample sizes for the \ost{} test on perturbed uniform $d$-dimensional distributions and on the MNIST dataset. The legend lists the name of the test followed by the sample sizes used.}
        \label{fig:ost}
    \end{figure}

    In \Cref{fig:ost}, we report the results of \Cref{fig:perturbed_uniform_1d,fig:perturbed_uniform_2d,fig:mnist} for \mmdu{} and \ost{} which are run using the same collections of bandwidths. 
    As previously mentioned, the \ost{} test is restricted to the use of the linear-time MMD estimate, and hence obtains low power compared to the other tests which use the quadratic-time MMD estimate.
    We increase the sample sizes for the \ost{} test until it matches the power of \mmdu{} with fixed sample sizes.

    For the case of $1$-dimensional perturbed uniform distributions, we observe in \Cref{fig:ost} that \ost{} requires sample sizes of 75\,000 and 50\,000 in order to match the power obtained by \mmdu{} with $m=n=500$ for the Gaussian and Laplace kernels, respectively.
    For the case of $2$-dimensional perturbed uniform distributions, one million, and half a million samples are required for the Gaussian and Laplace kernels, respectively, to obtain the same power as \mmdu{} with 2000 samples.
    When working with the MNIST dataset, it takes 250\,000 samples for \ost{} to achieve similar power to the one obtained by \mmdu{} with 500 samples.

    Recall that the MNIST dataset consists of 70\,000 images, so it is interesting to see that the power of \ost{} keeps increasing for sample sizes which are more than ten times bigger than the size of the dataset. 
    This is due to the use of the linear-time MMD estimate which for even sample sizes $n=m$ is equal to 
    $$
        \frac{2}{n} 
        \sum_{i=1}^{n/2}
        h_k\p{X_{2i-1}, X_{2i}, Y_{2i-1}, Y_{2i}}
    $$
    for $h_k$ defined as in \Cref{h}.
    The two pairs of samples $(X_{2i-1},X_{2i})$ and $(Y_{2i-1},Y_{2i})$ only appear together as a 4-tuple in this estimate, for $i=1,\dots,n/2$.
    So, as long as we do not sample exactly the two same pairs of images together, this creates a new 4-tuple which is considered as new data for this estimate. 
    This explains why considering sample sizes much larger than 70\,000 still results in an increase in power.

\FloatBarrier


\section{Relation between permutations and wild bootstrap}
\label{relation section}

In this section, we assume that we have equal sample sizes $m=n$ and show the relation between using permutations and using a wild bootstrap for the estimator $\mmdhmm$ defined in \Cref{mmdmm}.

First, we introduce some notation. 
For a matrix $A = \p{a_{i,j}}_{1\leq i,j\leq 2n}$, we denote the sum of all its entries by $A_+$ and denote by $A^{\circ}$ the matrix $A$ with all the entries 
$$
	\{a_{i,i}, a_{n+i,n+i}, a_{n+i,i}, a_{i,n+i}: i=1,\dots,n\}
$$
set equal to 0. Note that $A$ is composed of four $(n\times n)$-submatrices, and that $A^\circ$ is the matrix $A$ with the diagonal entries of those four submatrices set to $0$.
We let $\mathbbm{1}_n\in\R^{n\times 1}$ denote the vector of length $n$ with all entries equal to $1$.
We also let $v\coloneqq (\mathbbm{1}_n,-\mathbbm{1}_n)\in\R^{2n\times 1}$ and note that
$$
	vv^\top=\mat{\mathbbm{1}_n\mathbbm{1}_n^\top & -\mathbbm{1}_n\mathbbm{1}_n^\top \\ -\mathbbm{1}_n\mathbbm{1}_n^\top & \mathbbm{1}_n\mathbbm{1}_n^\top}\in\R^{2n\times 2n}.
$$
We let $K_\lambda$ denote the 
kernel matrix $\p{k_\lambda(U_i,U_j)}_{1\leq i,j \leq 2n}$ where $U_i \coloneqq X_i$ and $U_{n+i} \coloneqq Y_i$ for $i=1,\dots,n$.
We let $\mathrm{Tr}$ denote the trace operator and $\circ$ denote the Hadamard product.

By definition of \Cref{mmdmm}, we have 

\begin{align*}
	\mmdhmmv &\coloneqq \frac{1}{n(n-1)} 
	\sum_{1\leq i\neq j \leq n}
	\h(X_i, X_j, Y_i, Y_j)\\
	&= \frac{1}{n(n-1)}\sum_{1\leq i\neq j \leq n} k_\lambda(X_i,X_j) + k_\lambda(Y_i,Y_j) -  k_\lambda(X_{i},Y_{j}) - k_\lambda(Y_{i}, X_{j})\\
	&= \frac{1}{n(n-1)}\,\p{K_\lambda^\circ\circ v v^\top}_+ \\
	&= \frac{1}{n(n-1)}\,\mathrm{Tr}\p{K_\lambda^\circ \,v v^\top} \\
	&= \frac{1}{n(n-1)}\,\mathrm{Tr}\p{v^\top K_\lambda^\circ \,v} \\
	& = \frac{1}{n(n-1)}\,v^\top K_\lambda^\circ \,v.
\end{align*}

For the wild bootstrap, as presented in \Cref{bootstrap subsection}, we have $n$ i.i.d.\ Rademacher random variables 
$
	\epsilon\coloneqq(\epsilon_{1},\dots,\epsilon_{n})
$ 
with values in $\{-1,1\}^n$,
and let $v_\epsilon \coloneqq (\epsilon,-\epsilon)\in\{-1,1\}^{2n}$, so that
$$
	v_\epsilon v_\epsilon^\top=\mat{\epsilon\epsilon^\top & -\epsilon\epsilon^\top \\ -\epsilon\epsilon^\top & \epsilon\epsilon^\top}\in\R^{2n\times 2n}.
$$
As in \Cref{mmd bootstrap}, we then have
\begin{align*}
	\widehat M_\lambda^{\,\epsilon} &\coloneqq \frac{1}{n(n-1)} \sum_{1\leq i\neq j \leq n} \epsilon_{i}\epsilon_{j}\h(X_i,X_j,Y_i,Y_j)\\
	&= \frac{1}{n(n-1)}\sum_{1\leq i\neq j \leq n} \epsilon_i\epsilon_j k_\lambda(X_i,X_j) + \epsilon_i\epsilon_j k_\lambda(Y_i,Y_j) - \epsilon_i\epsilon_j k_\lambda(X_{i},Y_{j}) - \epsilon_i\epsilon_j k_\lambda(Y_{i}, X_{j})\\
	&= \frac{1}{n(n-1)}\,\p{K_\lambda^\circ\circ v_\epsilon v_\epsilon^\top}_+ \\
	&= \frac{1}{n(n-1)}\,\mathrm{Tr}\p{K_\lambda^\circ \,v_\epsilon v_\epsilon^\top} \\
	&= \frac{1}{n(n-1)}\,v_\epsilon^\top K_\lambda^\circ \,v_\epsilon.
\end{align*}

We introduce more notation. 
For a matrix $A = \p{a_{i,j}}_{1\leq i,j\leq 2n}$ and some permutation $\tau\colon \{1,\dots,2n\}\to\{1,\dots,2n\}$, we denote by $A^{\circ\tau}$ the matrix $A$ with all the entries $$\{a_{\tau(i),\tau(i)}, a_{\tau(n+i),\tau(n+i)}, a_{\tau(n+i),\tau(i)}, a_{\tau(i),\tau(n+i)}: i=1,\dots,n\}$$
set to be equal to 0.
We denote by $A_\tau$ the permuted matrix $\p{a_{\tau(i),\tau(j)}}_{1\leq i,j\leq 2n}$. Similarly, for a vector $w=(w_1,\dots,w_{2n})$, we write the permuted vector as $w_\tau=(w_{\tau(1)},\dots,w_{\tau(2n)})$.

Recall that $v = (v_1,\dots,v_{2n}) = (\mathbbm{1}_n,-\mathbbm{1}_n)\in\R^{2n\times 1}$.
Similarly to \Cref{mmd permuted} in \Cref{permutations}, but for the estimator $\mmdhmm$, given a permutation $\sigma\colon \{1,\dots,2n\}\to\{1,\dots,2n\}$, we can define $\widehat M_\lambda^{\,\sigma}$ as
\begin{align*} 
	&\frac{1}{n(n-1)} 
	\sum_{1\leq i\neq j \leq n}
	\h(U_{\sigma(i)}, U_{\sigma(j)}, U_{\sigma(n+i)}, U_{\sigma(n+j)}) \\
	=\, &\frac{1}{n(n-1)}\!\!\sum_{1\leq i\neq j \leq n} 
	\!\!\!k_\lambda(U_{\sigma(i)},\!U_{\sigma(j)}\!) \!+\! k_\lambda(U_{\sigma(n+i)},\!U_{\sigma(n+j)}\!) \!-\!  k_\lambda(U_{\sigma(i)},\!U_{\sigma(n+j)}\!) \!-\! k_\lambda(U_{\sigma(n+i)},\!U_{\sigma(j)}\!) \\
	=\, &\frac{1}{n(n-1)}\,\p{\p{\p{K_\lambda}_\sigma}^{\circ}\circ v v^\top}_+\\ 
	=\, &\frac{1}{n(n-1)}\,\p{\p{K_\lambda^{\circ\sigma}}_\sigma\circ v v^\top}_+ \\ 
	=\, &\frac{1}{n(n-1)}\,\p{K_\lambda^{\circ\sigma}\circ \p{v v^\top}_{\sigma^{-1}}}_+ \\
	=\, &\frac{1}{n(n-1)}\,\mathrm{Tr}\p{K_\lambda^{\circ\sigma} v_{\sigma^{-1}} v_{\sigma^{-1}}^\top} \\
	=\, &\frac{1}{n(n-1)}\,v_{\sigma^{-1}}^\top K_\lambda^{\circ\sigma} v_{\sigma^{-1}}.
\end{align*}

Using those formulas, we are able to prove the following proposition, but first we introduce two notions. 
Fix $\ell\in\{1,\dots,n\}$.
We say that a permutation $\tau\colon \{1,\dots,2n\}\to\{1,\dots,2n\}$ {\em fixes} $X_\ell=U_\ell$ and $Y_\ell=U_{n+\ell}$ if $\tau(\ell) = \ell$ and $\tau(n+\ell) = n+\ell$. 
Moreover, we say that it {\em swaps} $X_\ell=U_\ell$ and $Y_\ell=U_{n+\ell}$ if $\tau(\ell) = n+\ell$ and $\tau(n+\ell) = \ell$.
We denote by $\mathscr{P}$ the set of all permutations which either fix or swap $X_i$ and $Y_i$ for all $i=1,\dots,n$.
Since the identity belongs to $\mathscr{P}$, every element in $\mathscr{P}$ is self-inverse, and the composition of two permutations in $\mathscr{P}$ gives an element in $\mathscr{P}$, the set $\mathscr{P}$ is a subgroup of the permutation group.

\prop{
	\label{relation proposition}
	Assume we have equal sample sizes $m=n$ and that we work with the estimator $\mmdhmm$ defined in \Cref{mmdmm}.
	Then, using a wild bootstrap is equivalent to using permutations which belong to the subgroup $\mathscr{P}$. 
	There is a one-to-one correspondence between these two procedures.
}

\begin{proof}
	First, note that for a permutation $\sigma\in\mathscr{P}$, we either have
	$\sigma(i) = i$, $\sigma(n+i) = n+i$ or $\sigma(i) = n+i$, $\sigma(n+i) = i$
	for $i=1,\dots,n$. Hence, the two sets
	$$
		\left\{\kk(U_i,U_i), \kk(U_{n+i},U_{n+i}), \kk(U_i,U_{n+i}),  \kk(U_{n+i}, U_i): i=1,\dots,n\right\}
	$$
	and
	$$
		\left\{\kk(U_{\sigma(i)},U_{\sigma(i)}), \kk(U_{\sigma(n+i)},U_{\sigma(n+i)}), \kk(U_{\sigma(i)},U_{\sigma(n+i)}),  \kk(U_{\sigma(n+i)}, U_{\sigma(i)}): i=1,\dots,n\right\}
	$$
	are equal, we deduce that for $\sigma\in\mathscr{P}$ we have $K_\lambda^{\circ\sigma} = K_\lambda^{\circ}$. 
	Moreover, since a permutation $\sigma\in\mathscr{P}$ either fixes or swaps $X_i$ and $Y_i$ for $i=1,\dots,n$, it must be self-inverse, that is $\sigma^{-1}= \sigma$.
	For the permutation $\sigma\in\mathscr{P}$, we then have
	$$
		\widehat M_\lambda^{\,\sigma} =
		\frac{1}{n(n-1)}\,v_\sigma^\top K_\lambda^{\circ} v_\sigma.
	$$
	for $v_{\sigma} = (v_{\sigma(1)},\dots,v_{\sigma(2n)})$ 
	where $v_i = 1$ and $v_{n+i} = -1$ for $i=1,\dots,n$.
	We recall that for the wild bootstrap we have $n$ i.i.d.\ Rademacher random variables 
	$
		\epsilon\coloneqq(\epsilon_{1},\dots,\epsilon_{n})
	$ 
	with values in $\{-1,1\}^n$
	and 
	$$
		\widehat M_\lambda^{\,\epsilon} = \frac{1}{n(n-1)}\,v_\epsilon^\top K_\lambda^\circ \,v_\epsilon.
	$$
	where $v_\epsilon \coloneqq (\epsilon,-\epsilon)\in\{-1,1\}^{2n}$.

	We need to show that for a given permutation $\sigma\in\mathscr{P}$ there exists some $\epsilon\in\{-1,1\}^n$ such that $v_\epsilon = v_\sigma$, and that for a given $\epsilon\in\{-1,1\}^n$ there exists a permutation $\sigma\in\mathscr{P}$ such that $v_\sigma = v_\epsilon$, and that this correspondence is one-to-one.

	Suppose we have a permutation $\sigma\in\mathscr{P}$ and let $\epsilon \coloneqq (v_{\sigma(1)},\dots,v_{\sigma(n)}) \in \{-1,1\}^n$.
	We claim that $v_\sigma = v_\epsilon$, that is, that $(v_{\sigma(1)},\dots,v_{\sigma(2n)}) = (v_{\sigma(1)},\dots,v_{\sigma(n)}, -v_{\sigma(1)},\dots,-v_{\sigma(n)})$, so we need to prove that $v_{\sigma(n+i)} = - v_{\sigma(i)}$ for $i=1,\dots,n$.
	As $\sigma\in\mathscr{P}$, for $i=1,\dots,n$, we either have 
	$\sigma(i) = i$ and $\sigma(n+i) = n+i$ in which case
	$$
		v_{\sigma(n+i)} = v_{n+i} = -1 = -v_i = -v_{\sigma(i)},
	$$
	or $\sigma(i) = n+i$ and $\sigma(n+i) = i$ in which case we have 
	$$
		v_{\sigma(n+i)} = v_{i} = 1 = -v_{n+i} = -v_{\sigma(i)}.
	$$
	This proves the first direction.

	Now, suppose we are given $\epsilon \coloneqq (\epsilon_1,\dots,\epsilon_n)$ i.i.d.\ Rademacher random variables.
	We have $v_\epsilon = (\epsilon, -\epsilon) \in \{-1,1\}^{2n}$ and we need to construct $\sigma\in\mathscr{P}$ such that $v_\sigma = v_\epsilon$, that is, $v_{\sigma(i)} = \epsilon_i$ and $v_{\sigma(n+i)} = -\epsilon_i$ for $i=1,\dots,n$.
	We can construct such a permutation $\sigma\in\mathscr{P}$ as follows:
	\begin{samepage}
		\begin{spacing}{1}
			\begin{algorithmic}
				\For $i = 1,\dots,n$:
				\If $\epsilon_i = 1$ \textbf{then let} $\sigma(i) \coloneqq i$ \textbf{and} $\sigma(n+i)\coloneqq n+i$ (i.e.\ $\sigma$ fixes $X_i$ and $Y_i$)
				\State we then have  $v_{\sigma(i)} = v_i = 1 = \epsilon_i$ and $v_{\sigma(n+i)} = v_{n+i} = -1 = -\epsilon_i$
				\EndIf
				\If $\epsilon_i = -1$ \textbf{then let} $\sigma(i) \coloneqq n+i$  \textbf{and}  $\sigma(n+i) \coloneqq i$ (i.e.\ $\sigma$ swaps $X_i$ and $Y_i$)
				\State we then have  $v_{\sigma(i)} = v_{n+i} = -1 = \epsilon_i$ and $v_{\sigma(n+i)} = v_i = 1 = -\epsilon_i$
				\EndIf
				\EndFor
			\end{algorithmic}
		\end{spacing}
	\end{samepage}
	\noindent
	This proves the second direction.

	Our two constructions show that the correspondence is one-to-one.
\end{proof}

This highlights the relation between those two procedures: using a wild bootstrap is equivalent to using a restricted set of permutations for the estimator $\mmdhmm$.


\section{Efficient implementation of MMDAgg (\Cref{MMDAgg})}
\label{efficientMMDAgg}

We discuss how to compute Step 1 of \Cref{MMDAgg} efficiently. 
To compute the $\abs{\Lambda}$ kernel matrices for the Gaussian and Laplace kernels, we compute the matrix of pairwise distances only once.
For each $\lambda\in \Lambda$, we compute all the values $\big(\widehat{M}_{\lambda,1}^{\,b}\big)_{1\leq b \leq B_1+1}$ and $\big(\widehat{M}_{\lambda,2}^{\,b}\big)_{1\leq b \leq B_2}$ together.
We compute the sums over the permuted kernel matrices efficiently, in particular we do not want to explicitly permute rows and columns of the kernel matrices as this is computationally expensive. 

We start by considering the wild bootstrap case, so we have equal sample sizes $m=n$. 
We let $K_\lambda$ denote the 
kernel matrix $\p{k_\lambda(U_i,U_j)}_{1\leq i,j \leq 2n}$ where $U_i \coloneqq X_i$ and $U_{n+i} \coloneqq Y_i$ for $i=1,\dots,n$.
Note that $K_\lambda$ is composed of four $(n\times n)$-submatrices, we denote by $K_\lambda^\circ$ the matrix $K_\lambda$ with the diagonal entries of those four submatrices set to $0$.
As explained in \Cref{relation section}, for
$n$ i.i.d.\ Rademacher random variables 
$
\epsilon\coloneqq(\epsilon_{1},\dots,\epsilon_{n})
$ with values in $\{-1,1\}^n$, we have 
$$
	\widehat M_\lambda^{\,\epsilon} = \frac{1}{n(n-1)}\,v_\epsilon^\top K_\lambda^\circ \,v_\epsilon
$$
where $v_\epsilon \coloneqq (\epsilon,-\epsilon)\in\{-1,1\}^{2n}$.
We want to extend this to be able to compute $\big(\widehat M_\lambda^{\,\epsilon^{(b)}}\big)_{1\leq b \leq B}$ for any $B\in\Nz$.
We can do this by letting $R$ be the $2n\times B$ matrix consisting of stacked vectors $\big(v_{\epsilon^{(b)}}\big)_{1\leq b\leq B}$ and computing
$$
	\frac{1}{n(n-1)}\textrm{diag}\p{R^\top K_\lambda^\circ R}.
$$
Note that, in general, given $2n\times B$ matrices $A=(a_{i,j})_{\substack{1\leq i \leq 2n \\ 1 \leq j \leq B}}$ and $C=(c_{i,j})_{\substack{1\leq i \leq 2n \\ 1 \leq j \leq B}}$, we have
$$
	\textrm{diag}\p{A^\top C} = \textrm{diag}\p{\p{\sum_{r=1}^{2n} a_{r,i} c_{r,j}}_{\substack{1\leq i \leq B \\ 1 \leq j \leq B}}} = \p{\sum_{r=1}^{2n} a_{r,i} c_{r,i}}_{1\leq i \leq B} \eqqcolon \sum_{\textrm{rows}} A\circ C
$$
where $\circ$ denotes the Hadamard product and where $\sum_{\textrm{rows}}$ takes a matrix as input and outputs a vector which is the sum of the row vectors of the matrix. 
We deduce that
$$
	\textrm{diag}\p{R^\top K_\lambda^\circ R} = \sum_{\textrm{rows}} R \circ (K_\lambda^\circ R).
$$

We found that this way of computing the values 
$\big(\widehat M_\lambda^{\,\epsilon^{(b)}}\big)_{1\leq b \leq B}$
is computationally faster than other alternatives.
Letting $\mathbbm{1}_n$ denote the vector of length $n$ with all entries equal to $1$, we can obtain an efficient version for Step 1 of \Cref{MMDAgg} using a wild bootstrap as follows.

\begin{figure}[H]
	\vspace{0.1cm}\noindent\hrulefill\vspace{-0.1cm}
	\begin{spacing}{1}
		\begin{algorithmic}
			\State \textit{\underline{Efficient Step 1 of \Cref{MMDAgg} using a wild bootstrap:}}
			\State generate $n\times(B_1+B_2+1)$ matrix $\widetilde R$ of Rademacher random variables
			\State concatenate $\tilde R$ and $-\widetilde R$ to form the $2n\times(B_1+B_2+1)$ matrix $R$
			\State replace the $(B_1+1)^\textrm{th}$ column of $R$ with the vector $(\mathbbm{1}_n,-\mathbbm{1}_n)$ 
			\For {$\lambda\in\Lambda$:}
				\State compute kernel matrix $K_\lambda^\circ$ with zero diagonals for its four submatrices
				\State compute $\frac{1}{n(n-1)}\sum_{\textrm{rows}} R \circ (K_\lambda^\circ R)$ to get $\big(\widehat{M}_{\lambda,1}^{\,1},\dots,\widehat{M}_{\lambda,1}^{\, B_1+1}, \widehat{M}_{\lambda,2}^{\,1},\dots,\widehat{M}_{\lambda,2}^{\, B_2}\big)$
				\State
				$\!\big(\widehat{M}_{\lambda,1}^{\,\bullet1},\dots,\widehat{M}_{\lambda,1}^{\,\bullet B_1+1}\big) =$ \texttt{sort\_by\_ascending\_order}$\big(\widehat{M}_{\lambda,1}^{\,1},\dots,\widehat{M}_{\lambda,1}^{\, B_1+1}\big)$
			\EndFor
		\end{algorithmic}
	\vspace{-0.4cm}\noindent\hrulefill
	\end{spacing}
\end{figure}

Before tackling the case of permutations, we recall the main steps of the strategy used for the wild bootstrap case. 
As explained in \Cref{relation section}, we first noted that 
\begin{align*}
	\mmdhmmv 
	&= \p{K_\lambda^\circ\circ \mat{\mathbbm{1}_n\mathbbm{1}_n^\top/(n(n-1)) & -\mathbbm{1}_n\mathbbm{1}_n^\top/(n(n-1)) \\ -\mathbbm{1}_n\mathbbm{1}_n^\top/(n(n-1)) & \mathbbm{1}_n\mathbbm{1}_n^\top/(n(n-1))}}_+ \\
	&= \frac{1}{n(n-1)}\,\p{K_\lambda^\circ\circ v v^\top}_+ \\
	&= \frac{1}{n(n-1)}\,v^\top K_\lambda^\circ \,v.
\end{align*}
for $v\coloneqq (\mathbbm{1}_n,-\mathbbm{1}_n)\in\R^{2n\times 1}$, where $A_+$ denotes the sum all the entries of a matrix $A$.
We then observed that it was enough to replace the vector $v$ with $v_\epsilon \coloneqq (\epsilon,-\epsilon)\in\{-1,1\}^{2n}$ to obtain 
$$
	\widehat M_\lambda^{\,\epsilon} = \frac{1}{n(n-1)}\,v_\epsilon^\top K_\lambda^\circ \,v_\epsilon.
$$
The whole reasoning was based on the fact that we could rewrite the matrix 
$$
\mat{\mathbbm{1}_n\mathbbm{1}_n^\top/(n(n-1)) & -\mathbbm{1}_n\mathbbm{1}_n^\top/(n(n-1)) \\ -\mathbbm{1}_n\mathbbm{1}_n^\top/(n(n-1)) & \mathbbm{1}_n\mathbbm{1}_n^\top/(n(n-1))}
$$
as an outer product of vectors.

Now, we consider the permutation-based procedure.
For a square matrix $A$, we let $A^0$ denote the matrix $A$ with its diagonal entries set equal to 0.
We have 
\begin{align*}
	\mmdhmnv &=
	\frac{1}{m(m-1)} \sum_{1\leq i\neq i' \leq m} k(X_i,X_{i'})
	+ \frac{1}{n(n-1)}  \sum_{1\leq j\neq j' \leq n} k(Y_j,Y_{j'})\\
	&\hspace{5.43cm}- \frac{2}{mn} \sum_{i=1}^m \sum_{j=1}^n k(X_i,Y_j) \\
	&=\p{K_\lambda^0\circ \mat{\mathbbm{1}_m\mathbbm{1}_m^\top/(m(m-1)) & -\mathbbm{1}_m\mathbbm{1}_n^\top/(mn) \\ -\mathbbm{1}_n\mathbbm{1}_m^\top/(mn) & \mathbbm{1}_n\mathbbm{1}_n^\top/(n(n-1))}}_+ 
\end{align*}
where it is not possible to rewrite the matrix as an outer product of vectors.
Instead, we break it down into a sum of three outer products of vectors.
\begin{align*}
	\mat{\mathbbm{1}_m\mathbbm{1}_m^\top/(m(m-1)) & -\mathbbm{1}_m\mathbbm{1}_n^\top/(mn) \\ -\mathbbm{1}_n\mathbbm{1}_m^\top/(mn) & \mathbbm{1}_n\mathbbm{1}_n^\top/(n(n-1))}
	=\ &\p{\frac{1}{m(m-1)} - \frac{1}{mn}} \mat{\mathbbm{1}_m\mathbbm{1}_m^\top & \mymathbb{0}_m\mymathbb{0}_n^\top \\ \mymathbb{0}_n\mymathbb{0}_m^\top & \mymathbb{0}_n\mymathbb{0}_n^\top}	\\
	&+ \p{\frac{1}{n(n-1)} - \frac{1}{mn}} \mat{\mymathbb{0}_m\mymathbb{0}_m^\top & \mymathbb{0}_m\mymathbb{0}_n^\top \\ \mymathbb{0}_n\mymathbb{0}_m^\top & \mathbbm{1}_n\mathbbm{1}_n^\top}	\\
	&+ \frac{1}{mn} \mat{\mathbbm{1}_m\mathbbm{1}_m^\top & -\mathbbm{1}_m\mathbbm{1}_n^\top \\ -\mathbbm{1}_n\mathbbm{1}_m^\top & \mathbbm{1}_n\mathbbm{1}_n^\top} \\
	=\ &\frac{n-m+1}{mn(m-1)} u u^\top
	+ \frac{m-n+1}{mn(n-1)} w w^\top
	+ \frac{1}{mn} v v^\top
\end{align*}
where $u\coloneqq (\mathbbm{1}_m,\mymathbb{0}_n)$,
$w\coloneqq (\mymathbb{0}_m,-\mathbbm{1}_n)$ and
$v\coloneqq (\mathbbm{1}_m,-\mathbbm{1}_n)$
of shapes $(m+n)\times 1$,
with $\mymathbb{0}_n$ denoting the vector of length $n$ with all entries equal to 0.
Using this fact, we obtain that
$$
	\mmdhmnv 
	= \frac{n-m+1}{mn(m-1)} u^\top K_\lambda^0 u
	+ \frac{m-n+1}{mn(n-1)} w^\top K_\lambda^0 w
	+ \frac{1}{mn} v^\top K_\lambda^0 v.
$$
For a vector $a=(a_1,\dots,a_\ell)$ and a permutation $\tau\colon \{1,\dots,\ell\}\to\{1,\dots,\ell\}$, we denote the permuted vector as $a_{\tau}=(a_{\tau(1)},\dots,a_{\tau(\ell)})$.
Recall that $U_i \coloneqq X_i$, $i=1,\dots,m$ and $U_{m+j} \coloneqq Y_j$, $j=1,\dots,n$.
Consider a permutation $\sigma\colon \{1,\dots,m+n\}\to\{1,\dots,m+n\}$ and let $\Xm^\sigma\coloneqq \big(U_{\sigma(i)}\big)_{1\leq i \leq m}$ and $\Yn^\sigma \coloneqq \big(U_{\sigma(m+j)}\big)_{1\leq j \leq n}$.
Following a similar reasoning to the one presented in \Cref{relation section}, we find
\begin{align*}
	\widehat M_\lambda^{\,\sigma} &\coloneqq 
	\widehat{\mathrm{MMD}}^2_{\lambda,\mathtt{a}}(\Xm^\sigma,\Yn^\sigma) \\
	&= \frac{n-m+1}{mn(m-1)} u_{\sigma^{-1}}^\top K_\lambda^0 u_{\sigma^{-1}}
	+ \frac{m-n+1}{mn(n-1)} w_{\sigma^{-1}}^\top K_\lambda^0 w_{\sigma^{-1}}
	+ \frac{1}{mn} v_{\sigma^{-1}}^\top K_\lambda^0 v_{\sigma^{-1}}
\end{align*}
since $\big\{k_\lambda(X_i,Y_i):i=1,\dots,n\big\} = \big\{k_\lambda(X_{\sigma(i)},Y_{\sigma(i)}):i=1,\dots,n\big\}$.
The aim is to compute $\big(\widehat M_\lambda^{\,\sigma^{(b)}}\big)_{1\leq b\leq B}$ efficiently for any $B\in\Nz$.
We let $U$, $V$ and $W$  denote the $(m+n)\times B$ matrices of stacked vectors 
$\big(u_{{\sigma^{(b)}}^{-1}}\big)_{1\leq b\leq B}$,
$\big(v_{{\sigma^{(b)}}^{-1}}\big)_{1\leq b\leq B}$ and
$\big(w_{{\sigma^{(b)}}^{-1}}\big)_{1\leq b\leq B}$, respectively.
We are then able to compute $\big(\widehat M_\lambda^{\,\sigma^{(b)}}\big)_{1\leq b\leq B}$ as
\begin{align*}
	&\frac{n-m+1}{mn(m-1)} \textrm{diag}\p{U^\top K_\lambda^0 U}
	+ \frac{m-n+1}{mn(n-1)} \textrm{diag}\p{W^\top K_\lambda^0 W}
	+ \frac{1}{mn} \textrm{diag}\p{V^\top K_\lambda^0 V}\\
	=\ &\frac{n-m+1}{mn(m-1)} \sum_{\textrm{rows}} U \circ \p{K_\lambda^0 U}
	+ \frac{m-n+1}{mn(n-1)} \sum_{\textrm{rows}} W \circ \p{K_\lambda^0 W}
	+ \frac{1}{mn} \sum_{\textrm{rows}} V \circ \p{K_\lambda^0 V}.
\end{align*}

Since the inverse map for permutations is a bijection between the space of all permutations and itself, it follows that uniformly generating $B$ permutations and taking their inverses is equivalent to directly uniformly generating $B$ permutations. 
So, in practice, we can simply uniformly generate permutations $\tau^{(1)},\dots,\tau^{(B)}$ and assume that these correspond to
${\sigma^{(1)}}^{-1},\dots,{\sigma^{(B)}}^{-1}$ for uniformly generated permutations $\sigma^{(1)},\dots,\sigma^{(B)}$.
We can now present an efficient version for Step 1 of \Cref{MMDAgg} using permutations.

\begin{figure}[H]
	\vspace{0.1cm}\noindent\hrulefill\vspace{-0.1cm}
	\begin{spacing}{1}
		\begin{algorithmic}
			\State \textit{\underline{Efficient Step 1 using permutations:}}
			\State construct $\mpn \times(B_1+B_2+1)$ matrix $U$ of stacked vectors of $(\mathbbm{1}_m,\mymathbb{0}_n)$
			\State construct $\mpn \times(B_1+B_2+1)$ matrix $V$ of stacked vectors of $(\mathbbm{1}_m,-\mathbbm{1}_n)$
			\State construct $\mpn \times(B_1+B_2+1)$ matrix $W$ of stacked vectors of $(\mymathbb{0}_m,-\mathbbm{1}_n)$
			\State use $B_1+B_2$ permutations to permute the elements of the columns of $U$, $V$ and $W$ without permuting the elements of the $(B_1+1)^\textrm{th}$ columns of $U$, $V$ and $W$
			\For {$\lambda\in\Lambda$:}
				\State compute kernel matrix $K_\lambda^0$ with zero diagonals
				\State compute $\big(\widehat{M}_{\lambda,1}^{\,1},\dots,\widehat{M}_{\lambda,1}^{\, B_1+1}, \widehat{M}_{\lambda,2}^{\,1},\dots,\widehat{M}_{\lambda,2}^{\, B_2}\big)$ as
				$$
				\frac{n-m+1}{mn(m-1)} \sum_{\textrm{rows}} U \circ \p{K_\lambda^0 U}
				+ \frac{m-n+1}{mn(n-1)} \sum_{\textrm{rows}} W \circ \p{K_\lambda^0 W}
				+ \frac{1}{mn} \sum_{\textrm{rows}} V \circ \p{K_\lambda^0 V}
				$$
				\State $\!\big(\widehat{M}_{\lambda,1}^{\,\bullet1},\dots,\widehat{M}_{\lambda,1}^{\,\bullet B_1+1}\big) =$ \texttt{sort\_by\_ascending\_order}$\big(\widehat{M}_{\lambda,1}^{\,1},\dots,\widehat{M}_{\lambda,1}^{\, B_1+1}\big)$
			\EndFor
		\end{algorithmic}
	\vspace{-0.4cm}\noindent\hrulefill
	\end{spacing}
\end{figure}


\section{Lower bound on the minimax rate over a Sobolev ball}
\label{appendix lower}

We claim that the two-sample minimax rate of testing over the Sobolev ball $\Sb$ is $n^{-2s/(4s+d)}$.
Formally, let $\alpha,\beta\in(0,1)$, $d\in\Nz$ and $M,s,R\in(0,\infty)$, we claim that there exists some positive constant $C_0'(M,d,s,R,\alpha,\beta)$ such that
\begin{equation}
	\label{li yuan result}
	\underline\rho\!\left(\Sb, \alpha,\beta,M\right) \coloneqq 
	\aainf{\Delta_\alpha}{\rho\!\left(\Delta_{\alpha}, \Sb, \beta,M\right)} 
	\geq C_0'(M,d,s,R,\alpha,\beta)\, n^{-2s/(4s+d)}
\end{equation}
where the infimum is taken over all tests $\Delta_{\alpha}$ of non-asymptotic level $\alpha$ and where $c'\leq \frac{m}{n}\leq C'$ for some positive constants $c'$ and $C'$. 

The proof mirrors the reasoning of \citet[Section 4, Theorem 4]{albert2019adaptive} who derive the independence minimax rate of testing $n^{-2s/(4s+d_1+d_2)}$ over the Sobolev ball $\mathcal{S}^s_{d_1+d_2}(R)$ considering paired samples on $\R^{d_1}\times\R^{d_2}$. 
While the case $d=1$ is not covered by their framework, the proof extends naturally to it.
We illustrate the main reasoning behind the proof as this is of interest for our experiments in \Cref{exp power synthetic}.

First, note that the minimax rate $\underline\rho\!\left(\Sb, \alpha,\beta,M\right)$ is
\begin{align*}
	\aainf{\Delta_\alpha}{\rho\!\left(\Delta_{\alpha}, \Sb, \beta,M\right)} &= \aainf{\Delta_\alpha} \inf\left\{\tilde\rho>0: \sup_{(p,q)\in\mathcal{F}^M_{\tilde\rho}(\Sb)}\pq{\Delta_{\alpha}(\Xm,\Yn)=0} \leq \bb\right\} \\
	&=  \inf\left\{\tilde\rho>0: \aainf{\Delta_\alpha}\sup_{(p,q)\in\mathcal{F}^M_{\tilde\rho}(\Sb)}\pq{\Delta_{\alpha}(\Xm,\Yn)=0} \leq \bb\right\}
\end{align*}
where $
\mathcal{F}^M_{\tilde\rho}(\Sb) \coloneqq \acc{(p,q): \max\p{\norm{p}_\infty,\norm{q}_\infty} \leq M, p-q \in \Sb,\, \norm{p-q}_2 > \tilde\rho}
$.
Hence, to prove \Cref{li yuan result} it suffices to construct two probability densities $p$ and $q$ on $\R^d$ which satisfy 
$\max\p{\norm{p}_\infty,\norm{q}_\infty} \leq M$,
$p-q \in \Sb$, 
$\norm{p-q}_2 < C_1' \mpn^{-2s/(4s+d)}$ for some $C_1'>0$,
and for which 
$\pq{\Delta_\alpha'(\Xm,\Yn)=0} > \bb$
holds for all tests $\Delta_\alpha$ with non-asymptotic level $\alpha$.
Intuitively, one constructs densities which are close enough in $L^2$-norm so that any test with non-asymptotic level $\alpha$ fails to distinguish them from one another.

\citet[Section 4]{albert2019adaptive} show that a suitable choice of $p$ and $q$ is to take the uniform probability density on $[0,1]^d$ and a perturbed version of it. 
To construct the latter, first define for all $u\in\R$ the function 
$$
	G(u) \coloneqq \exp\left(-\frac{1}{1-(4u+3)^2} \right) \mathbbm{1}_{\left(-1,-\frac{1}{2}\right)}(u)
	- \exp\left(-\frac{1}{1-(4u+1)^2} \right) \mathbbm{1}_{\left(-\frac{1}{2},0\right)}(u)
$$
which is plotted in \Cref{fig:uniform} in \Cref{exp power synthetic}. 
Consider $P\in\Nz$ and some vector $\theta~=~(\theta_\nu)_{\nu\in\{1,\dots,P\}^d}\in\{-1,1\}^{P^d}$ of length $P^d$ with entries either $-1$ or $1$ which is indexed by the $P^d$ $d$-dimensional elements of $\{1,\dots,P\}^d$.
Then, the perturbed uniform density is defined as
\begin{equation}
	\label{f_theta}
	f_\theta (u) \coloneqq \mathbbm{1}_{[0,1]^d}(u) + P^{-s} \sum_{\nu\in\{1,\dots,P\}^d} \theta_\nu \prod_{i=1}^dG\left(P u_i-\nu_i\right)
\end{equation}
for all $u\in\R^d$.
As illustrated in \Cref{fig:uniform}, this indeed corresponds to a uniform probability density with $P$ perturbations along each dimension.

This construction by \citet{albert2019adaptive} is a generalisation of the detailed construction of \citet[Section~5]{butucea2007goodness} for the 1-dimensional case for goodness-of-fit testing.
We also point out the work of \citet[Theorems 5, part (ii)]{li2019optimality} who use a similar construction to show that, for any alternative\footnote{To be more precise, the alternatives considered by \citet{li2019optimality} require both $p$ and $q$ to belong to the Sobolev ball (rather than just $p-q$ in our case) but do not require these densities to be bounded. Moreover, they use the non-homogeneous Sobolev space definition (with an extra `$+1$' in the weighting term) while we use the homogeneous definition (without that extra term).} in $\mathcal{F}^M_{\tilde\rho}(\Sb)$ with rate $\tilde\rho$ smaller or equal to $n^{-2s/(4s+d)}$, there exists some $\alpha\in(0,1)$ such that any test with asymptotic level $\alpha$ must have power asymptotically strictly smaller than one.
The original idea behind all those constructions is due to 
\citet{ingster1987minimax,ingster1993minimax} with his work on nonparametric minimax rates for goodness-of-fit testing.


\section{Proofs}
\label{proofs}

In this section, we prove the statements presented in \Cref{theory}.
We first introduce some standard results.

First, recall that we assume 
$m\leq n$
and
$n \leq C m$
for some constant $C\geq1$ as in \Cref{mn}.
It follows that 
\begin{equation}
	\label{mn equivalence}
	\frac{1}{m}+\frac{1}{n} \leq \frac{C+1}{n} = \frac{2(C+1)}{2n} \leq \frac{2(C+1)}{m+n}
	\quad \textrm{and} \quad
	\frac{1}{m+n}\leq \frac{2}{m+n} \leq \frac{1}{m}+\frac{1}{n}.
\end{equation}

For the kernels $K_1,\dots,K_d$ satisfying the properties presented in \Cref{kernel}, we define the constants
\begin{equation}
	\label{kappa}
	\kappa_1(d) \coloneqq \prod_{i=1}^d \int_{\R} \abs{K_i\p{x_i}} \dd x_i < \infty
	\quad \textrm{and} \quad 
	\kappa_2(d) \coloneqq \prod_{i=1}^d \int_{\R} K_i\p{x_i}^2 \dd x_i < \infty
\end{equation}
which are well-defined as $K_i\in L^1(\R)\cap L^2(\R)$ for $i=1,\dots,d$ by assumption.
We do not make explicit the dependence on $K_1,\dots,K_d$ in the constants as we consider those to be chosen \emph{a priori}.
Moreover, we often use the kernel properties of $\kk$ presented in \Cref{hypintkcarre}, that are
$$
	\int_{\R^d} \kk(x,y) \dd x 
	= \prod_{i=1}^d \frac{1}{\lambda_i} \int_{\R} K_i\p{\frac{x_i-y_i}{\lambda_i}} \!\dd x_i
	= \prod_{i=1}^d \int_{\R} K_i\p{x_i'} \!\dd x_i'
	= 1
$$
and
$$
	\int_{\R^d} \kk(x,y)^2 \dd x 
	= \prod_{i=1}^d \frac{1}{\lambda_i^2} \int_{\R} K_i\p{\frac{x_i-y_i}{\lambda_i}}^2 \dd x_i
	= \frac{1}{\LL}\prod_{i=1}^d \int_{\R} K_i\p{x_i'}^2 \dd x_i'
	= \frac{\kappa_2}{\LL}.
$$

We often use in our proofs the standard result that, for $a_1,\dots,a_\ell\in\R$, we have
$$
	\p{\sum_{i=1}^\ell a_i}^2 \leq \p{\sum_{i=1}^\ell1^2} \p{\sum_{i=1}^\ell a_i^2} = \ell \sum_{i=1}^\ell a_i^2
$$
which holds by Cauchy--Schwarz inequality.

In our proofs, we show that there exist some constants which are large enough so that our results hold.
We keep track of those constants and show how they depend on each other.
The aim is to show that such constants exist, we do not focus on obtaining the tightest constants possible.

\subsection{Proof of \Cref{level}}
	\label{prooflevel}

	Recall that in \Cref{permutations,bootstrap subsection} we have constructed elements $\big(\widehat{M}_\lambda^{\,b}\big)_{1\leq b \leq B+1}$ for the two MMD estimators defined in \Cref{mmdmn,mmdmm}, respectively. 
	The first one uses permutations while the second uses a wild bootstrap.
	For those two cases, we first show that the elements $\big(\widehat{M}_\lambda^{\,b}\big)_{1\leq b \leq B+1}$ are exchangeable under the null hypothesis $\HH_0\colon p=q$. 
	We are then able to prove that the test $\db$ has the prescribed level using the exchangeability of $\big(\widehat{M}_\lambda^{\,b}\big)_{1\leq b \leq B+1}$.

	\paragraph{Exchangeability using permutations as in \Cref{permutations}.\\}
	Recall that in this case we have permutations $\sigma^{(1)},\dots,\sigma^{(B)}$ of $\{1,\dots,m+n\}$. We also have
	$
		\widehat M_\lambda^{\,b} \coloneqq 
		\widehat{\mathrm{MMD}}^2_{\lambda,\mathtt{a}}(\Xm^{\sigma^{(b)}},\Yn^{\sigma^{(b)}})
	$
	for $b=1,\dots,B$ 
	and
	$
		\widehat M_\lambda^{\,B+1} \coloneqq 
		\widehat{\mathrm{MMD}}^2_{\lambda,\mathtt{a}}(\Xm,\Yn).
	$
	Following the same reasoning as in the proof of \citet[Proposition 1, Equation C.1]{albert2019adaptive}, we can deduce that $\big(\widehat{M}_\lambda^{\,b}\big)_{1\leq b \leq B+1}$ are exchangeable under the null hypothesis.
	The only difference is that they work with the Hilbert Schmidt Independence Criterion rather than with the Maximum Mean Discrepancy, but this does not affect the reasoning of the proof.

	\paragraph{Exchangeability using a wild bootstrap as in \Cref{bootstrap subsection}. \\}

	The exchangeability under the null using the wild bootstrap follows from the one using permutations since by \Cref{relation proposition} using the wild bootstrap corresponds to using a subgroup of the permutation group. 
	We show below how the original statistic and the permuted one can be seen to have the same distribution under the null.

	For $b=1,\dots,B$, we have $n$ i.i.d.\ Rademacher random variables $\epsilon^{(b)}\coloneqq \big(\epsilon^{(b)}_1,\dots,\epsilon^{(b)}_n\big)$ with values in $\{-1,1\}^n$ and
	$$
		\widehat M_\lambda^{\,b} \coloneqq \frac{1}{n(n-1)} \sum_{1\leq i\neq j \leq n} \epsilon_{i}^{(b)}\epsilon_{j}^{(b)}\h(X_i,X_j,Y_i,Y_j)
	$$
	where $\h$ is defined in \Cref{h}.
	We also have
	$$
		\widehat{M}_\lambda^{\, B+1}\coloneqq \mmdhmmv =
		\frac{1}{n(n-1)} \sum_{1\leq i\neq j \leq n} \h(X_i,X_j,Y_i,Y_j).
	$$
	By the reproducing property of the kernel $k_\lambda$, we have
	\begin{align*}
		\bigg( \sup_{f \in \mathcal{F}_\lambda}  \frac{1}{n} \sum_{i=1}^n \big(f(X_i) - f(Y_i)\big)   \bigg)^2
		&= \bigg( \sup_{f \in \mathcal{F}_\lambda} \bigg\langle f, \frac{1}{n}\sum_{i=1}^n \kk(X_i,\cdot) - \frac{1}{n}\sum_{i=1}^n \kk(Y_i,\cdot)\bigg\rangle_{\mathcal{H}_{k_\lambda}} \bigg)^2 \\
		&= \norm{\frac{1}{n}\sum_{i=1}^n \kk(X_i,\cdot) - \frac{1}{n}\sum_{i=1}^n \kk(Y_i,\cdot)}_{\mathcal{H}_{k_\lambda}}^2 \\
		&= \frac{1}{n^2} \sum_{i=1}^n\sum_{j=1}^n \h(X_i,X_j,Y_i,Y_j)  
	\end{align*}
	where $\mathcal{F}_\lambda \coloneqq \{f\in\mathcal{H}_{k_\lambda} \,:\, \norm{f}_{\mathcal{H}_{k_\lambda}} \leq 1\}$.
	Under the null hypothesis $\HH_0\colon p=q$, all the samples $(X_1,\ldots,X_n,Y_1,\ldots,Y_n)$ are independent and identically distributed. So, the distribution of $\big( \sup_{f \in \mathcal{F}_\lambda} \frac{1}{n} \sum_{i=1}^n \big(f(X_i) - f(Y_i)\big) \big)^2$ does not change if we randomly exchange $X_i$ and $Y_i$ for each $i=1,\ldots,n$. 
	This can be formalized using $n$ i.i.d.\ Rademacher random variables $\epsilon_1,\dots,\epsilon_{n}$, we have
	\begin{align*}
		\bigg( \sup_{f \in \mathcal{F}_\lambda} \frac{1}{n} \sum_{i=1}^n \big(f(X_i) - f(Y_i)\big) \bigg)^2\ \underset{\HH_0}{\overset{d}{=}}\ \bigg( \sup_{f \in \mathcal{F}_\lambda} \frac{1}{n} \sum_{i=1}^n \epsilon_i\big(f(X_i) - f(Y_i)\big) \bigg)^2,
	\end{align*} 
	where the notation $\underset{\HH_0}{\overset{d}{=}}$ means that the two random variables have the same distribution under the null hypothesis $\HH_0\colon p=q$. 
	Since we also have
	\begin{align*}
		\bigg( \sup_{f \in \mathcal{F}_\lambda}  \frac{1}{n} \sum_{i=1}^n \epsilon_i\big(f(X_i) - f(Y_i)\big)   \bigg)^2
		&= \bigg( \sup_{f \in \mathcal{F}_\lambda} \bigg\langle f, \frac{1}{n}\sum_{i=1}^n \epsilon_i\kk(X_i,\cdot) - \frac{1}{n}\sum_{i=1}^n \epsilon_i\kk(Y_i,\cdot)\bigg\rangle_{\mathcal{H}_{k_\lambda}} \bigg)^2 \\
		&= \norm{\frac{1}{n}\sum_{i=1}^n \epsilon_i\kk(X_i,\cdot) - \frac{1}{n}\sum_{i=1}^n \epsilon_i\kk(Y_i,\cdot)}_{\mathcal{H}_{k_\lambda}}^2 \\
		&= \frac{1}{n^2} \sum_{i=1}^n\sum_{j=1}^n \epsilon_i \epsilon_j \h(X_i,X_j,Y_i,Y_j) ,
	\end{align*}
	we obtain
	\begin{align*}
		\sum_{i=1}^n\sum_{j=1}^n \h(X_i,X_j,Y_i,Y_j) \ \underset{\HH_0}{\overset{d}{=}}\ \sum_{i=1}^n\sum_{j=1}^n \epsilon_i \epsilon_j \h(X_i,X_j,Y_i,Y_j) . 
	\end{align*}
	Since $\epsilon_i^2 = 1$ for $i=1,\dots,n$, subtracting $\sum_{i=1}^n \h(X_i,X_i,Y_i,Y_i) $ from both sides, we get
	\begin{align*}
		\sum_{1 \leq i \neq j \leq n} \h(X_i,X_j,Y_i,Y_j) \ \underset{\HH_0}{\overset{d}{=}}\ \sum_{1 \leq i \neq j \leq n} \epsilon_i \epsilon_j \h(X_i,X_j,Y_i,Y_j) . 
	\end{align*}

	\paragraph{Level of the test.\\}

	Following a similar reasoning to the one presented by \citet[Proposition~1]{albert2019adaptive}, we have
	\begin{align*}
		\dbv=1 \quad &\Longleftrightarrow\quad 
		\mmdhv \,>\, \qbv \\
		&\Longleftrightarrow\quad\widehat{M}_\lambda^{\, B+1}  > \widehat{M}_\lambda^{\,\bullet\ceil{(B+1)(1-\alpha)}} \\
		&\Longleftrightarrow\quad\sum_{b=1}^{B+1} \one{\widehat{M}_\lambda^{\, b}<\widehat{M}_\lambda^{\, B+1}}\geq \ceil{(B+1)(1-\alpha)}\\
		&\Longleftrightarrow\quad B+1 -\sum_{b=1}^{B+1} \one{\widehat{M}_\lambda^{\, b}<\widehat{M}_\lambda^{\, B+1}}\leq B+1 - \ceil{(B+1)(1-\alpha)} \\
		&\Longleftrightarrow\quad\sum_{b=1}^{B+1} \one{\widehat{M}_\lambda^{\, b}\geq\widehat{M}_\lambda^{\, B+1}}\leq \floor{\alpha(B+1)} \\
		&\Longleftrightarrow\quad\sum_{b=1}^{B+1} \one{\widehat{M}_\lambda^{\, b}\geq\widehat{M}_\lambda^{\, B+1}}\leq \alpha(B+1) \\
		&\Longleftrightarrow\quad\frac{1}{B+1} \left(1 + \sum_{b=1}^{B} \one{\widehat{M}_\lambda^{\, b}\geq\widehat{M}_\lambda^{\, B+1}}\right)\leq \alpha
	\end{align*}
	where we have used the fact that 
	$
		B+1 - \ceil{(B+1)(1-\alpha)} = \floor{\alpha(B+1)}
	$.
	Using the exchangeability of $\big(\widehat{M}_\lambda^{\,b}\big)_{1\leq b \leq B+1}$, the result of \citet[Lemma~1]{romano2005exact} guarantees that
	$$
		\ppr{\frac{1}{B+1} \left(1 + \sum_{b=1}^{B} \one{\widehat{M}_\lambda^{\, b}\geq\widehat{M}_\lambda^{\, B+1}}\right)\leq \alpha} 
		\leq\ \alpha.
	$$
	We deduce that
	$$
		\ppr{\dbv=1}  \leq \alpha.
	$$

\subsection{Proof of \Cref{controlpower}}
	\label{proofcontrolpower}
	Let 
	$$
		\mathcal{A} \coloneqq \acc{\mmdhv\leq\qbv} 
	$$
	and 
	$$
		\mathcal{B} \coloneqq \acc{\mmdv  \geq \sqrt{\frac{2}{\bb}\vpq{\mmdhv}} + \qbv}.
	$$
	By assumption, we have $\pqr{\mathcal{B}}\geq 1-\frac{\beta}{2}$, and we want to show $\pqr{\mathcal{A}}\leq \beta$.
	Note that 
	\begin{align*}
		\pqr{\mathcal{A}\,|\,\mathcal{B}} 
		&= \pqr{\mmdhv\leq\qbv\,\big|\,\mathcal{B}} \\
		&\leq \pq{\mmdhv\leq\mmdv- \sqrt{\frac{2}{\bb}\vpq{\mmdhv}}} \\
		&= \pq{\mmdv-\mmdhv\geq \sqrt{\frac{2}{\bb}\vpq{\mmdhv}}}\\
		&\leq \pq{\abs{\mmdv-\mmdhv}\geq \sqrt{\frac{2}{\bb}\vpq{\mmdhv}}} \\
		&\leq \frac{\beta}{2}		
	\end{align*}
	by Chebyshev's inequality \citep{chebyshev1899oeuvres} as $\epq{\mmdhv} = \mmdv$. 
	We then have 
	\begin{align*}
		\pqr{\mathcal{A}} &= \pqr{\mathcal{A}\,|\,\mathcal{B}}\pqr{\mathcal{B}} + \pqr{\mathcal{A}\,|\,\mathcal{B}^c}\pqr{\mathcal{B}^c} \\
		&\leq \frac{\beta}{2} \cdot 1 + 1 \cdot \frac{\beta}{2} \\
		&= \beta.
	\end{align*}

\subsection{Proof of \Cref{boundvar}}
	\label{proofboundvar}

	We prove this result separately for our two MMD estimators $\mmdhmn$ and $\mmdhmm$ defined in \Cref{mmdmn,mmdmm}, respectively. 

	\paragraph{Variance bound for MMD estimator $\mmdhmn$ defined in \Cref{mmdmn}.\\}

	In this case, we use the fact that $\mmdhmn$ can be written as a two-sample $U$-statistic as in \Cref{mmdmnU}.
	As noted by \citet[Appendix E, Part 1]{kim2020minimax} one can derive from the explicit variance formula of the two-sample $U$-statistic \citep[Equation 2 p.38]{lee1990ustatistic} that there exists some positive constant $c_0$ such that
	$$
		\vpq{\mmdhmnv} \leq c_0 \left( \frac{\sigma_{\lambda,1,0}^2}{m} + \frac{\sigma_{\lambda,0,1}^2}{n}+ \mn^2 \sigma_{\lambda,2,2}^2 \right)
	$$
	for
	\begin{align*}
		\sigma_{\lambda,1,0}^2 &\coloneqq \VV{X}{\EE{X',Y,Y'}{\h(X,X',Y,Y')}},\\
		\sigma_{\lambda,0,1}^2 &\coloneqq \VV{Y}{\EE{X,X',Y'}{\h(X,X',Y,Y')}},\\
		\sigma_{\lambda,2,2}^2 &\coloneqq \VV{X,X',Y,Y'}{\h(X,X',Y,Y')},
	\end{align*}
	where $X,X'\overset{\textrm{iid}}{\sim} p$ and $Y,Y' \overset{\textrm{iid}}{\sim} q$ are all independent of each other.
	Making use of \Cref{mn equivalence}, we deduce that there exists a positive constant $c_0^\dagger$ such that
	$$
		\vpq{\mmdhmnv} \leq c_0^\dagger \left( \frac{\sigma_{\lambda,1,0}^2+\sigma_{\lambda,0,1}^2}{m+n} + \frac{\sigma_{\lambda,2,2}^2}{\mpn^2} \right).
	$$
	Recall that $\vl(u) \coloneqq \prod_{i=1}^d \frac{1}{\lambda_i}K_i\p{\frac{u_i}{\lambda_i}}$ for $u\in\R^d$ and that $\psi\coloneqq p-q$. 
	Letting $G_\lambda = \psi*\vl$, we then have for all $u\in\R^d$
	\begin{align*}
		G_\lambda(u) &= \left(\psi*\vl\right)(u) \\
		&= \int_{\R^d} \psi(u')\vl(u-u')\dd u'\\
		&= \int_{\R^d} \psi(u')\kk(u,u')\dd u'\\
		&= \int_{\R^d} \kk(u,u')\left(p(u')-q(u')\right)\dd u'\\
		&= \EE{X'}{\kk(u,X')}-\EE{Y'}{\kk(u,Y')}.
	\end{align*}
	Note that
	\begin{align*}
		\EE{X',Y'}{\h(X,X',Y,Y')} =~ &\e_{X',Y'}\!\left[\kk(X,X') + \kk(Y,Y') - \kk(X,Y') - \kk(X',Y)\right] \\
		=~ &\e_{X'}\!\left[\kk(X,X')\right] - \e_{Y'}\!\left[\kk(X,Y')\right] \\
		&-\p{\e_{X'}\left[\kk(Y,X')\right] - \e_{Y'}\left[\kk(Y,Y')\right]}\\
		=~ &G_\lambda(X) - G_\lambda(Y).
	\end{align*}
	Hence, we get
	\begin{align*}
		\sigma_{\lambda,1,0}^2 &\coloneqq \VV{X}{\EE{X',Y,Y'}{\h(X,X',Y,Y')}}\\
		&= \VV{X}{\EE{Y}{G_\lambda(X) - G_\lambda(Y)}}\\
		&= \VV{X}{G_\lambda(X) - \EE{Y}{G_\lambda(Y)}}\\
		&= \VV{X}{G_\lambda(X)}\\
		&\leq \EE{X}{G_\lambda(X)^2}\\
		&= \int_{\R^d} G_\lambda(x)^2 p(x) \dd x \\
		&\leq \norm{p}_\infty \int_{\R^d} G_\lambda(x)^2 \dd x \\
		&\leq M \norm{G_\lambda}_2^2\\
		&= M \norm{\psi*\vl}_2^2
	\end{align*}
	and, similarly, we get
	$$
		\sigma_{\lambda,0,1}^2 \coloneqq \VV{Y}{\EE{X,X',Y'}{\h(X,X',Y,Y')}}\leq M \norm{\psi*\vl}_2^2.
	$$
	For the third term, we have
	\begin{align*}
		\sigma_{\lambda,2,2}^2 &\coloneqq \VV{X,X',Y,Y'}{\h(X,X',Y,Y')}\\
		&= \VV{X,X',Y,Y'}{\kk(X,X') + \kk(Y,Y') - \kk(X,Y') - \kk(X',Y)}\\
		&\leq \EE{X,X',Y,Y'}{\big(\kk(X,X') + \kk(Y,Y') - \kk(X,Y') - \kk(X',Y)\big)^2}\\
		&\leq 4 \p{\EE{X,X'}{ \kk(X,X')^2} + \EE{Y,Y'}{ \kk(Y,Y')^2} + 2\EE{X,Y}{ \kk(X,Y)^2}}.
	\end{align*}
	Note that
	\begin{align*}
		\EE{X,Y}{\kk(X,Y)^2}
		&= \int_{\R^d} \int_{\R^d} \kk(x,y)^2 p(x)q(y)\,\dd x \dd y \\
		&\leq \norm{p}_\infty  \int_{\R^d} \p{\int_{\R^d} \kk(x,y)^2 \dd x} q(y)\, \dd y \\
		&= \norm{p}_\infty \frac{\kappa_2}{\LL} \int_{\R^d} q(y)\, \dd y \\
		&\leq \frac{M \kappa_2}{\LL}
	\end{align*}
	where $\kappa_2$ depends on $d$ and is defined in \Cref{kappa}.
	Similarly, we have
	\begin{equation}
		\label{variance k upper bound}
		\EE{X,X'}{ \kk(X,X')^2} \leq \frac{M \kappa_2}{\LL} 
		\qquad \textrm{and} \qquad
		\EE{Y,Y'}{ \kk(Y,Y')^2}  \leq \frac{M \kappa_2}{\LL}.
	\end{equation}
	We deduce that 
	\begin{equation*}
		\sigma_{\lambda,2,2}^2 
		\coloneqq \VV{X,X',Y,Y'}{\h(X,X',Y,Y')}
		\leq \frac{16 M \kappa_2}{\LL}.
	\end{equation*}
	Letting $C_1(M,d) \coloneqq \max\acc{2c_0^\dagger M, 16 c_0^\dagger   M \kappa_2}$ and
	combining the results, we obtain 
	\begin{align*}
		\vpq{\mmdhmnv} &\leq c_0^\dagger  \left( \frac{\sigma_{\lambda,1,0}^2+\sigma_{\lambda,0,1}^2}{m+n} + \frac{\sigma_{\lambda,2,2}^2}{\mpn^2} \right)\\
		&\leq C_1(M,d) \left( \frac{\norm{\psi*\vl}_2^2}{m+n} + \frac{1}{\mpn^2\LL} \right). \\
	\end{align*}

	\paragraph{Variance bound for MMD estimator $\mmdhmm$ defined in \Cref{mmdmm}. \\}
	The MMD estimator
	$$
		\mmdhmmv \coloneqq \frac{1}{n(n-1)}\sum_{1\leq i\neq j\leq n} \h(X_i,X_j,Y_i,Y_j) 
	$$
	is a one-sample $U$-statistic of order 2. 
	Hence, we can apply the result of \citet[Lemma~10]{albert2019adaptive} to get
	$$
		\vpq{\mmdhmmv} \leq \widetilde c_0 \left( \frac{\sigma_{\lambda,1,1}^2}{n}+ \frac{\sigma_{\lambda,2,2}^2}{n^2} \right)
	$$
	for some positive constant $\widetilde c_0$,
	where 
	$$
		\sigma_{\lambda,1,1}^2 \coloneqq \VV{X,Y}{\EE{X',Y}{\h(X,X',Y,Y')}}
	$$
	and
	$$
		\sigma_{\lambda,2,2}^2 \coloneqq \VV{X,X',Y,Y'}{\h(X,X',Y,Y')} \leq \frac{16 M \kappa_2}{\LL}
	$$
	as shown earlier. Using the above results, we get
	\begin{align*}
		\sigma_{\lambda,1,1}^2 &\coloneqq \VV{X,Y}{\EE{X',Y}{\h(X,X',Y,Y')}}\\
		&= \VV{X,Y}{G_\lambda(X) - G_\lambda(Y)}\\
		&\leq \EE{X,Y}{\p{G_\lambda(X) - G_\lambda(Y)}^2} \\
		&\leq 2\p{\EE{X}{G_\lambda(X)^2} + \EE{Y}{G_\lambda(Y)^2}} \\
		&\leq 4M \norm{\psi*\vl}_2^2.
	\end{align*}
	Letting $\widetilde C_1(M,d) \coloneqq 4\max\acc{4\widetilde c_0M, 16\widetilde c_0  M \kappa_2}$, we deduce that
	\begin{align*}
		\vpq{\mmdhmmv} &\leq \widetilde c_0 \left( \frac{\sigma_{\lambda,1,1}^2}{n}+ \frac{\sigma_{\lambda,2,2}^2}{n^2} \right) \\
		&\leq \frac{1}{4}\widetilde C_1(M, d) \left( \frac{\norm{\psi*\vl}_2^2}{n}+ \frac{1}{n^2\LL} \right) \\
		&\leq \widetilde C_1(M, d) \left( \frac{\norm{\psi*\vl}_2^2}{2n}+ \frac{1}{(2n)^2\LL} \right).
	\end{align*}

\subsection{Proof of \Cref{boundquantile}}
	\label{proofboundquantile}

	Recall that $(\widehat M_\lambda^{\,b})_{1\leq b\leq B}$ is defined in \Cref{permutations,bootstrap subsection} for the estimators $\mmdhmn$ and $\mmdhmm$, respectively, and that 
	$
		\widehat{M}_\lambda^{\, B+1} \coloneqq \mmdh(\Xm,\Yn)
	$ 
	for both estimators.
	Let us recall that the $(1\!-\!\alpha)$-quantile function of a random variable $X$ with cumulative distribution function $F_X$ is given by
	\begin{align*}
		q_{1-\alpha} = \inf\{x \in \mathbb{R}: 1 - \alpha \leq F_X(x) \}.
	\end{align*}
	We denote by $F_B$ and $F_{B+1}$ the empirical cumulative distribution functions of $(\widehat M_\lambda^{\,b})_{1\leq b\leq B}$ and $(\widehat M_\lambda^{\,b})_{1\leq b\leq B+1}$, respectively.

	For the case of the estimator $\mmdhmn$, we denote by $F_\infty$ the cumulative distribution function of the conditional distribution of 
	$\widehat M_\lambda^{\,\sigma}$ (defined in \Cref{mmd permuted}) given $\Xm$ and $\Yn$, where the randomness comes from the uniform choice of permutation $\sigma$ among all possible permutations of $\{1,\dots,m+n\}$. 
	For the case of the estimator $\mmdhmm$, we similarly denote by $F_\infty$ the cumulative distribution function of the conditional distribution of 
	$\widehat M_\lambda^{\,\epsilon}$ (defined in \Cref{mmd bootstrap}) given $\Xm$ and $\Yn$, where the randomness comes from the $n$ i.i.d.\ Rademacher variables 
	$
		\epsilon\coloneqq(\epsilon_{1},\dots,\epsilon_{n})
	$ with values in $\{-1,1\}^n$.

	Based on the above definitions, we can write 
	$$
		\widehat{q}_{1-\alpha}^{\,\lambda,\infty}(\Xm,\Yn) =  \inf\{u \in \mathbb{R}: 1 - \alpha \leq F_{\infty}(u) \}
	$$
	and
	\begin{align*}
		\qbv &=  \inf\{u \in \mathbb{R}: 1 - \alpha \leq F_{B+1}(u) \} \\	
		&=  \inf\bigg\{u \in \mathbb{R}: 1 - \alpha \leq \frac{1}{B+1}\sum_{b=1}^{B+1} \one{\widehat M_\lambda^{\, b}\leq u}\bigg\}\\	
		&=  \inf\bigg\{u \in \mathbb{R}: (B+1)(1 - \alpha) \leq \sum_{b=1}^{B+1} \one{\widehat M_\lambda^{\, b}\leq u}\bigg\}\\	
		&= \widehat{M}_\lambda^{\,\bullet\ceil{(B+1)(1-\alpha)}}
	\end{align*}
	where 
	$
		\widehat{M}_\lambda^{\,\bullet 1}  \leq \dots \leq \widehat{M}_\lambda^{\,\bullet B+1}
	$
	denote the ordered simulated test statistics $(\widehat M_\lambda^{\,b})_{1\leq b\leq B+1}$.

	Now, for any given $\delta > 0$, define the event
	\begin{align*}
		\mathcal{A} \coloneqq \Bigg\{ \sup_{u \in \mathbb{R}} | F_B(u) - F_\infty(u)| \leq \sqrt{\frac{1}{2B} \log \bigg( \frac{4}{\delta} \bigg)} \Bigg\}. 
	\end{align*}
	As noted by \citet[Remark 2.1]{kim2020minimax}, Dvoretzky--Kiefer--Wolfowitz inequality \citep{dvoretzky1956asymptotic}, more precisely the version with the tight constant which is due to \citet{massart1990tight}, then guarantees that $\mathbb{P}_{r}(\mathcal{A}|\Xm,\Yn) \geq 1 - \frac{\delta}{2}$ for any $\Xm$ and $\Yn$, so we deduce that $\pqr{\mathcal{A}} \geq 1 - \frac{\delta}{2}$. 
	We now assume that the event $\mathcal{A}$ holds, so the bound we derive holds with probability $1-\frac{\delta}{2}$.
	Notice that we cannot directly apply the Dvoretzky--Kiefer--Wolfowitz inequality to $F_{B+1}$ since it is not based on i.i.d.\ samples. 
	Nevertheless, under the event $\mathcal{A}$, we have 
	\begin{align*}
		\qbv &= \inf\{u \in \mathbb{R}: 1 - \alpha \leq F_{B+1}(u) \} \\
		&= \inf \bigg\{ u \in \mathbb{R} : 1 - \alpha \leq \frac{1}{B+1} \sum_{b=1}^{B+1} \one{\widehat M_\lambda^{\, b} \leq u} \bigg\} \\
		&\leq \inf \bigg\{ u \in \mathbb{R} : 1 - \alpha \leq \frac{1}{B+1} \sum_{b=1}^B \one{\widehat M_\lambda^{\, b} \leq u} \bigg\} \\
		&= \inf \bigg\{ u \in \mathbb{R} : (1 - \alpha) \frac{B+1}{B} \leq F_B(u) \bigg\} \\ 
		&\leq \inf \bigg\{ u \in \mathbb{R} : {(1 - \alpha) \frac{B+1}{B} \leq F_\infty(u) - \sqrt{\frac{1}{2B} \log \bigg(\frac{4}{\delta}\bigg)}} \bigg\} \\
		&= \inf \bigg\{ u \in \mathbb{R} : \underbrace{(1 - \alpha) \frac{B+1}{B} + \sqrt{\frac{1}{2B} \log \bigg(\frac{4}{\delta}\bigg)}}_{\coloneqq 1 - \alpha^\ast} \leq F_\infty(u) \bigg\} \\
		&= \widehat{q}_{1-\alpha^\ast}^{\,\lambda,\infty}(\Xm,\Yn).
	\end{align*}
	Now, we take $B$ large enough (only depending on $\alpha$ and $\delta$) such that 
	\begin{align*}
		(1 - \alpha) \frac{B+1}{B} + \sqrt{\frac{1}{2B} \log \bigg(\frac{4}{\delta}\bigg)} \leq 1 - \frac{\alpha}{2}
	\end{align*}
	so that $\widehat{q}_{1-\alpha^\ast}^{\,\lambda,\infty}(\Xm,\Yn) \leq \widehat{q}_{1-\alpha/2}^{\,\lambda,\infty}(\Xm,\Yn)$ under the event $\mathcal{A}$.
	By reducing this problem to a quadratic equation with respect to $B$, we find 
	$$
		B\geq \frac{2}{\alpha^2}\p{\frac{1}{2}\ln\p{\frac{4}{\delta}}+\alpha-\alpha^2 +\sqrt{\p{\frac{1}{2}\ln\p{\frac{4}{\delta}}+\alpha-\alpha^2}^2-\alpha^2(1-\alpha)^2}}.
	$$
	In particular, by upper bounding $-\alpha^2(1-\alpha)^2$ by $0$, we find that the above inequality holds as soon as 
	$$
		B\geq  \frac{4}{\alpha^2}\p{\frac{1}{2}\ln\p{\frac{4}{\delta}}+\alpha(1-\alpha)}.
	$$
	Note that this condition is in particular trivially satisfied if 
	$$
		B\geq \frac{3}{\alpha^2}\p{\ln\p{\frac{4}{\delta}}+\alpha(1-\alpha)}.
	$$
	With this choice of $B$, we have 
	$$
		\pqr{\qbv \leq \widehat{q}_{1-\alpha/2}^{\,\lambda,\infty}(\Xm,\Yn)}\geq 1-\frac{\delta}{2}.
	$$ 
	We now upper bound $\widehat{q}^{\,\lambda, \infty}_{1-\alpha}(\Xm,\Yn)$ for the two estimators $\mmdhmn$ and $\mmdhmm$ separately. We then use this to prove the required upper bound on $\qbv$.

	\paragraph{Quantile bound for MMD estimator $\mmdhmn$ defined in \Cref{mmdmn}. \\}
	In this case, we base our reasoning on the work of \citet[proof of Lemma C.1]{kim2020minimax}. 
	Recall from \Cref{mn} that we assume that $m\leq n$ and $n\leq C m$ for some positive constant $C$.
	We use the notation presented in \Cref{permutations} that
	$U_i \coloneqq X_i$ and $U_{m+j} \coloneqq Y_j$ 
	for $i=1,\dots,m$
	and $j=1,\dots,n$.
	By the result of \citet[Equation 59]{kim2020minimax}, there exists some $c_1>0$ such that
	$$
		\widehat{q}_{1-\alpha}^{\,\lambda,\infty}(\Xm,\Yn) \leq c_1 \sqrt{\frac{1}{m^2(m-1)^2}\sum_{1\leq i\neq j \leq m+n} \kk(U_i,U_j)^2} \lna
	$$ 
	almost surely.
	As shown in \Cref{variance k upper bound}, for the constant $\kappa_2(d)$ defined in \Cref{kappa}, we have 
	\begin{equation}
		\label{expectation k square upper bound}
		\max\!\acc{\EE{X,X'}{\kk(X,X')^2}, \EE{X,Y}{\kk(X,Y)^2}, \EE{Y,Y'}{\kk(Y,Y')^2}}
		\leq \frac{M\kappa_2}{\LL}
	\end{equation}
	where $X,X'\overset{\textrm{iid}}{\sim} p$ and $Y,Y'\overset{\textrm{iid}}{\sim} q$ are all independent of each other.
	We deduce that 
	\begin{align*}
		\epq{\frac{1}{m^2(m-1)^2}\sum_{1\leq i\neq j \leq m+n} \kk(U_i,U_j)^2}
		&\leq \frac{(m+n)(m+n-1)}{m^2(m-1)^2}\frac{M\kappa_2}{\LL}\\
		&\leq \frac{4M\kappa_2}{\LL}\frac{(m+n)^2}{m^4}\\
		&=    \frac{64M\kappa_2}{\LL}\frac{(m+n)^2}{(2m)^4}\\
		&\leq \frac{64M\kappa_2}{\LL}\frac{(m+n)^2}{(m+C^{-1}n)^4}\\
		&\leq \frac{64M\kappa_2C^4}{\LL}\frac{1}{(m+n)^2}
	\end{align*}
	where we use the fact that $n\leq Cm$.
	Using Markov's inequality, we get that, for any $\delta\in(0,1)$, we have
	\begin{align*}
		&1 - \frac{\delta}{2} \\
		\leq ~&\pq{\!\frac{1}{m^2(m-1)^2}\!\!\!\sum_{1\leq i\neq j \leq m+n}\!\!\! \kk(U_i,U_j)^2 \leq \frac{2}{\delta} \mathbb{E}_{p \times q}\!\cc{\!\frac{1}{m^2(m-1)^2}\!\!\!\sum_{1\leq i\neq j \leq m+n} \!\!\!\kk(U_i,U_j)^2\!\!}\!} \\
		\leq ~&\pq{\frac{1}{m^2(m-1)^2}\sum_{1\leq i\neq j \leq m+n} \kk(U_i,U_j)^2 \leq \frac{2}{\delta}\frac{64M\kappa_2C^4}{\mpn^2\LL}} \\
		\leq ~&\pq{\widehat{q}^{\,\lambda, \infty}_{1-\alpha/2}(\Xm,\Yn) \leq \frac{1}{\sqrt\delta}8c_1C^2\sqrt{2M\kappa_2}\frac{\ln\!\p{\frac{2}{\alpha}}}{\mpn\sqrt{\LL}}} \\
		\leq ~&\pq{\widehat{q}^{\,\lambda, \infty}_{1-\alpha/2}(\Xm,\Yn) \leq \frac{1}{\sqrt\delta}16c_1C^2\sqrt{2M\kappa_2}\frac{\ln\!\p{\frac{1}{\alpha}}}{\mpn\sqrt{\LL}}}
	\end{align*}
	as $\ln\p{\frac{2}{\alpha}}\leq 2 \ln\p{\frac{1}{\alpha}}$ since $\alpha\in(0,0.5)$.
	We now let $C_2(M,d) \coloneqq 16c_1C^2\sqrt{2M\kappa_2}$. 
	Then, for all $B\in\N$ such that
	$B\geq \frac{3}{\alpha^2}\p{\ln\p{\frac{4}{\delta}}+\alpha(1-\alpha)}$, we have
	\begin{align*}
		&\pqr{\qbv\leq \frac{1}{\sqrt{\delta}} C_2(M,d) \frac{\ln\p{\frac{1}{\alpha}}}{\mpn\sqrt{\LL}}} \\
		\geq\ &\PPP_{p\times q \times r}\Bigg(\acc{\qbv \leq \widehat{q}_{1-\alpha/2}^{\,\lambda,\infty}(\Xm,\Yn)}\\
		&\hspace{1.5cm}\cap\acc{\widehat{q}^{\,\lambda, \infty}_{1-\alpha/2}(\Xm,\Yn) \leq \frac{1}{\sqrt{\delta}} C_2(M,d)\frac{\ln\p{\frac{1}{\alpha}}}{\mpn\sqrt{\LL}}}\Bigg)\\
		\geq\ &1-\frac{\delta}{2}-\frac{\delta}{2}\\
		=\ &1- \delta
	\end{align*}
	where we use the standard fact that for events $\mathcal{B}$ and $\mathcal{C}$ satisfying $\PPP(\mathcal{B})\geq 1-\delta_1$ and $\PPP(\mathcal{C})\geq 1-\delta_2$, we have
	$\PPP(\mathcal{B}\cap \mathcal{C}) = 1-\PPP(\mathcal{B}^c\cup \mathcal{C}^c) \geq 1-\PPP(\mathcal{B}^c)-\PPP(\mathcal{C}^c) \geq 1- \delta_1 - \delta_2.$

	\paragraph{Quantile bound for MMD estimator $\mmdhmm$ defined in \Cref{mmdmm}. \\}

	For this case, in order to upper bound $\widehat{q}^{\,\lambda, \infty}_{1-\alpha}(\Xn,\Yn)$, we can use the result of \citet[Corollary 3.2.6]{victor1999decoupling} and Markov's inequality as done by \citet[Appendix D]{fromont2012kernels}. We obtain that there exists a positive constant $\widetilde c_1$ such that
	$$
		\PPP_{r}\p{\abs{\sum_{1\leq i\neq j \leq n} \epsilon_{i}\epsilon_{j}\h(X_i, X_j, Y_i, Y_j)}\geq \widetilde c_1\sqrt{\sum_{1\leq i\neq j \leq n} \h(X_i, X_j, Y_i, Y_j)^2}\ln\p{\frac{2}{\alpha}} \,\bigg|\, \Xn,\Yn}\leq \alpha
	$$
	where $\epsilon_1,\dots,\epsilon_n$ are i.i.d.\ Rademacher variables, and so $\sum_{1\leq i\neq j \leq n} \epsilon_{i}\epsilon_{j}\h(X_i, X_j, Y_i, Y_j)$ is a Rademacher chaos.
	We deduce that
	$$
		\widehat{q}^{\,\lambda, \infty}_{1-\alpha}(\Xn,\Yn)\leq 
		\widetilde c_1 \sqrt{\frac{1}{n^2(n-1)^2}\sum_{1\leq i\neq j \leq n} \h(X_i, X_j, Y_i, Y_j)^2}
		\ln\p{\frac{2}{\alpha}} 
	$$
	almost surely.
	Using \Cref{expectation k square upper bound}, we have
	\begin{align*}
		&\epq{\frac{1}{n^2(n-1)^2}\sum_{1\leq i\neq j \leq n} \h(X_i, X_j, Y_i, Y_j)^2} \\
		=~&\frac{1}{n^2(n-1)^2}\epq{\sum_{1\leq i\neq j \leq n} \big(k(X_i,X_j) + k(Y_i,Y_j) - k(X_i,Y_j) - k(X_j,Y_i)\big)^2}\\
		\leq~&\frac{4}{n(n-1)} \bigg(\EE{X,X'}{\kk(X,X')^2} + \EE{Y,Y'}{\kk(Y,Y')^2} + 2\EE{X,Y}{\kk(X,Y)^2}\bigg)\\
		\leq~&\frac{32 M \kappa_2}{n^2\LL}
	\end{align*}
	where $X,X'\overset{\textrm{iid}}{\sim} p$ and $Y,Y'\overset{\textrm{iid}}{\sim} q$ are all independent of each other, and where $\kappa_2$ is the constant defined in \Cref{kappa} depending on $d$. 
	Similarly to the previous case, we can then use Markov's inequality to get that, for any $\delta\in(0,1)$, we have
	\begin{align*}
		&1 - \frac{\delta}{2} \\
		\leq\ &\pq{\!\frac{1}{n^2(n-1)^2}\!\!\!\sum_{1\leq i\neq j \leq n}\!\!\!\! \h(X_i, X_j, Y_i, Y_j)^2 \leq \frac{2}{\delta} \mathbb{E}_{p \times q}\!\cc{\!\!\frac{1}{n^2(n-1)^2}\!\!\!\sum_{1\leq i\neq j \leq n}\!\!\!\! \h(X_i, X_j, Y_i, Y_j)^2\!\!}\!} \\
		\leq\ &\pq{\frac{1}{n^2(n-1)^2}\sum_{1\leq i\neq j \leq n} \h(X_i, X_j, Y_i, Y_j)^2 \leq \frac{2}{\delta}\frac{32 M \kappa_2}{n^2\LL}} \\
		\leq\ &\pq{\widehat{q}^{\,\lambda, \infty}_{1-\alpha/2}(\Xn,\Yn) \leq \frac{1}{\sqrt\delta}8\widetilde c_1\sqrt{M \kappa_2}\frac{\ln\!\p{\frac{4}{\alpha}}}{n\sqrt{\LL}}} \\
		\leq\ &\pq{\widehat{q}^{\,\lambda, \infty}_{1-\alpha/2}(\Xn,\Yn) \leq \frac{1}{\sqrt\delta}48\widetilde c_1\sqrt{M \kappa_2}\frac{\ln\!\p{\frac{1}{\alpha}}}{2n\sqrt{\LL}}}
	\end{align*}
	as $\ln\p{\frac{4}{\alpha}}\leq 3 \ln\p{\frac{1}{\alpha}}$ since $\alpha\in(0,0.5)$. 
	Letting $\widetilde C_2(M,d) \coloneqq 48\widetilde c_1\sqrt{M \kappa_2}$ and applying the same reasoning as earlier, we get
	$$
		\pqr{\qbv\leq \frac{1}{\sqrt{\delta}} \widetilde C_2(M,d) \frac{\ln\p{\frac{1}{\alpha}}}{2n\sqrt{\LL}}} 
		\leq 1- \delta
	$$
	for all $B\in\N$ satisfying
	$B\geq \frac{3}{\alpha^2}\p{\ln\p{\frac{4}{\delta}}+\alpha(1-\alpha)}$.

\subsection{Proof of \Cref{phipower}}
	\label{proofphipower}

	First, as shown by \citet[Lemma~6]{gretton2012kernel}, the Maximum Mean Discrepancy can be written as 
	\begin{align*}
		\mmdv =\ &\EE{X,X'}{\kk(X,X')} -2\, \e_{X, Y}\left[\kk(X,Y)\right]+ \EE{Y,Y'}{\kk(Y,Y')}\\
		=\ &\int_{\R^d}\int_{\R^d} \kk(x,x') p(x)p(x') \,\dd x \dd x' \\
		&-2 \int_{\R^d}\int_{\R^d} \kk(x,y) p(x)q(y) \,\dd x \dd y \\
		&+ \int_{\R^d}\int_{\R^d} \kk(y,y') q(y)q(y') \,\dd y \dd y' \\
		=\ &\int_{\R^d}\int_{\R^d} \kk(u,u') \big(p(u)-q(u)\big) \big(p(u')-q(u')\big)  \,\dd u \dd u'
	\end{align*}
	for $X,X'\overset{\textrm{iid}}{\sim} p$ and $Y,Y'\overset{\textrm{iid}}{\sim} q$ all independent of each other.
	Using the function $\varphi_\lambda$ defined in \Cref{phi definition} and $\psi\coloneqq p-q$, we obtain
	\begin{align*}
		\mmdv
		=\ &\int_{\R^d}\int_{\R^d} \varphi_\lambda(u-u') \psi(u)\psi(u')  \,\dd u \dd u'\\
		=\ &\int_{\R^d}\psi(u)\int_{\R^d} \psi(u') \varphi_\lambda(u-u')   \,\dd u' \dd u\\
		=\ &\int_{\R^d}\psi(u)\big(\psi*\varphi_\lambda\big)(u) \,\dd u\\
		=\ &\langle\psi, \psi * \varphi_\lambda\rangle_2 \\
		=\ &\frac{1}{2} \Big( \norm{\psi}^2_2 + \norm{\psi * \varphi_\lambda}^2_2 - \norm{\psi-\psi * \varphi_\lambda}^2_2  \Big)
	\end{align*} 
	where the last equality is obtained by expanding $\norm{\psi-\psi * \varphi_\lambda}^2_2$.
	By \Cref{controlpower}, a sufficient condition to ensure that 
	$\pii\leq \bb$
	is
	$$
		\pqr{\mmdv  \geq \sqrt{\frac{2}{\bb}\vpq{\mmdhv}} + \qbv} \geq 1-\frac{\beta}{2}
	$$
	and an equivalent sufficient condition is
	$$
		\pqr{\norm{\psi}^2_2 \geq \norm{\psi-\psi * \varphi_\lambda}^2_2 - \norm{\psi * \varphi_\lambda}^2_2 + 2\sqrt{\frac{2}{\bb}\vpq{\mmdh}} + 2\qb} \geq 1-\frac{\beta}{2}.
	$$
	By \Cref{boundvar}, we have
	\begin{align*}
		\vpq{\mmdhv} &\leq C_1(M,d) \left(\frac{\norm{\psi*\vl}_2^2}{m+n}+\frac{1}{\mpn^2\LL}\right)\\
		2\sqrt{\frac{2}{\bb}\vpq{\mmdhv}} &\leq  2\sqrt{\frac{2C_1}{\beta}\frac{\norm{\psi*\vl}_2^2}{m+n}+\frac{2C_1}{\beta \mpn^2 \LL}}\\
		&\leq  2\sqrt{\norm{\psi*\vl}_2^2\frac{2C_1}{\beta \mpn}} + \frac{2\sqrt{2C_1}}{\sqrt{\beta}  \mpn \sqrt{\LL}} \\
		&\leq \norm{\psi*\vl}_2^2 + \frac{2C_1}{\beta \mpn} + \frac{2\sqrt{2C_1}}{\sqrt{\beta}  \mpn \sqrt{\LL}} \\
		&\leq \norm{\psi*\vl}_2^2 + \frac{2C_1+2\sqrt{2C_1}}{\beta \mpn \sqrt{\LL}} \lna \\
		&\leq \norm{\psi*\vl}_2^2 + \frac{6C_1}{\beta \mpn \sqrt{\LL}}  \lna \\
		\norm{\psi*\vl}_2^2-2\sqrt{\frac{2}{\bb}\vpq{\mmdh}}&\geq -6C_1\frac{\lna}{\beta \mpn \sqrt{\LL}}  
	\end{align*}
	where for the third inequality we used the fact that $\sqrt{x+y}\leq\sqrt{x}+\sqrt{y}$ for all $x,y>0$, for the fourth inequality we used the fact that $2\sqrt{xy}\leq x + y$ for all $x,y>0$, and for the fifth inequality we use the fact that $\LL\leq 1$, $\beta\in(0,1)$ and $ \ln\!\left(\frac{1}{\alpha}\right) > 1$. 
	A similar reasoning has been used by \citet[Theorem 1]{fromont2013two} and \citet[Theorem 1]{albert2019adaptive}.

	Let $C_3(M,d)\coloneqq 6C_1(M,d)+2\sqrt{2}C_2(M,d)$ where $C_1$ and $C_2$ are the constants from \Cref{boundquantile,boundvar}, respectively.
	Assume that our condition holds, that is 
	$$ 
		\norm{\psi}^2_2 - \norm{\psi-\psi * \varphi_\lambda}^2_2 \geq (2\sqrt{2}C_2+6C_1)\frac{\ln\!\left(\frac{1}{\alpha}\right)}{\beta \mpn \sqrt{\LL}} .
	$$
	Omitting the variables for $\qbv$ and for $\mmdhv$, we then get
	\begin{align*}
		&\pqr{2\qb\leq \norm{\psi}^2_2 - \norm{\psi-\psi * \varphi_\lambda}^2_2 + \norm{\psi * \varphi_\lambda}^2_2 - 2\sqrt{\frac{2}{\bb}\vpq{\mmdh}}} \\
		\geq\ &\pqr{2\qb\leq (6C_1+2\sqrt{2}C_2)\frac{\ln\!\left(\frac{1}{\alpha}\right)}{\beta \mpn \sqrt{\LL}}-6C_1\frac{\ln\!\left(\frac{1}{\alpha}\right)}{\beta \mpn \sqrt{\LL}}} \\
		=\ &\pqr{\qb\leq \sqrt{2}C_2\frac{\ln\!\left(\frac{1}{\alpha}\right)}{\beta \mpn \sqrt{\LL}} } \\
		\geq\ &\pqr{\qb\leq C_2\sqrt{\frac{2}{\beta}}\frac{\ln\!\left(\frac{1}{\alpha}\right)}{ \mpn \sqrt{\LL}} } \\
		\geq\ &1-\frac{\beta}{2}
	\end{align*}
	where the third inequality holds because $\beta\in(0,1)$ and the last one holds
	by \Cref{boundquantile} since $B\geq \frac{3}{\alpha^2}\big(\ln\big(\frac{8}{\beta}\big)+\alpha(1-\alpha)\big)$. \Cref{controlpower} then implies that
	$$
		\pii\leq \bb.
	$$

\subsection{Proof of \Cref{sobolevusr}}
	\label{proofsobolevusr}
	\Cref{phipower} gives us a condition on $\norm{\psi}^2_2-\norm{\psi-\psi * \vl}^2_2$ to control the power of the test $\db$. We now want to upper bound $\norm{\psi-\psi * \vl}^2_2$ in terms of the bandwidths when assuming that the difference of the densities lie in a Sobolev ball.
	We first prove that if $\psi \coloneqq p-q \in \Sb$ for some $s>0$ and $R>0$, then 
	there exists some $S\in(0,1)$ such that 
	\begin{equation}
		\label{intermediate result}
		\norm{\psi-\psi * \vl}^2_2 - S^2 \norm{\psi}^2_2 \leq C_4'(d,s,R) 
		\sum_{i=1}^d \lambda_i^{2s}
	\end{equation}
	for some positive constant $C_4'(d,s,R)$.

	For $j=1,\dots,d$, since $K_j\in L^1(\R)\cap L^2(\R)$, it follows by the Riemann-Lebesgue Lemma that its Fourier transform $\widehat K_j$ is continuous. 
	For $j=1,\dots,d$, note that 
	$$
		\widehat K_j(0)= \int_{\R} K_j(x) e^{-ix0} \dd x = \int_{\R} K_j(x) \dd x = 1
	$$
	and, since $K_j\in L^1(\R)\cap L^2(\R)$, also that
	$$
		\prod_{j=1}^d\abs{\widehat K_j(\xi_j)} \leq \prod_{j=1}^d\int_{\R} \abs{K_j(x) e^{-ix\xi_j}} \dd x = \prod_{j=1}^d\int_{\R} \abs{K_j(x)} \dd x \eqqcolon \kappa_1 < \infty
	$$
	as defined in \Cref{kappa}.
	We deduce that $\abs{1-\prod_{i=1}^d \widehat K_i(\xi_i)}\leq 1+\kappa_1$ for all $\xi\in\R^d$.
	Let us define $g\colon\R^d\to\R$ by $g(\xi)=1-\prod_{i=1}^d \widehat K_i(\xi_i)$ for $\xi\in\R^d$. We have $g(0,\dots,0)=0$, so by continuity of $g$, there exists some $t>0$ such that
	$$
		S \coloneqq \aasup{\norm{\xi}_2\leq t} \abs{g(\xi)} <1.
	$$
	For any $s>0$, we also define
	$$
		T_s \coloneqq \aasup{\norm{\xi}_2> t} \frac{\abs{1-\prod_{i=1}^d \widehat K_i(\xi_i)}}{\norm{\xi}_2^s} \leq \frac{1+\kappa_1}{t^s}<\infty.
	$$

	Let $\Psi \coloneqq \psi-\psi*\vl$. 
	As it is a scaled product of $K_1,\dots,K_d\in L^1(\R)\cap L^2(\R)$, we have $\varphi_\lambda\in L^1(\R^d)\cap L^2(\R^d)$.
	Since we assume that $\psi\in\Sb$, we have $\psi\in L^1(\R^d)\cap L^2(\R^d)$. 
	For $p\in\{1,2\}$, since $\psi\in L^1(\R^d)$, we have $\norm{\psi*\vl}_p \leq \norm{\psi}_1 \norm{\vl}_p < \infty$.
	Hence, we deduce that $\Psi\in L^1(\R^d)\cap L^2(\R^d)$.
	By Plancherel's Theorem, we then have 
	\begin{align*}
		(2\pi)^d \norm{\Psi}_2^2 &= |\hspace{-0.3mm}|\widehat \Psi|\hspace{-0.3mm}|_2^2 \\
		(2\pi)^d \norm{\psi-\psi*\vl}_2^2 &= \norm{(1-\widehat\vl)\widehat\psi }_2^2.
	\end{align*}
	In general, for $a>0$ the Fourier transform of a function $x \mapsto \frac{1}{a} f\!\left(\frac{x}{a}\right)$ is $\xi \mapsto \widehat f(a\xi)$. 
	Since $\vl(u) \coloneqq \prod_{i=1}^d \frac{1}{\lambda_i}K_i\p{\frac{u_i}{\lambda_i}}$ for $u\in\R^d$,
	we deduce that 
	$
		\widehat\vl(\xi) = \prod_{i=1}^d \widehat K_i(\lambda_i \xi_i)
	$
	for $\xi\in\R^d$.
	Therefore, we have
	\begin{align*}
		&(2\pi)^{d}\norm{\psi-\psi*\vl}_2^2 \\
		=\ &\norm{(1-\widehat\vl)\widehat\psi }_2^2 \\
		=\ &\int_{\R^d} \abs{1-\widehat\vl(\xi)}^2\abs{\widehat\psi(\xi)}^2 \dd \xi \\
		=\ &\int_{\R^d} \abs{1-\prod_{i=1}^d \widehat K(\lambda_i \xi_i)}^2\abs{\widehat\psi(\xi)}^2 \dd \xi \\
		=\ &\int_{\norm{\xi}_2\leq t} \abs{1-\prod_{i=1}^d \widehat K(\lambda_i \xi_i)}^2\abs{\widehat\psi(\xi)}^2 \dd \xi + \int_{\norm{\xi}_2> t} \abs{1-\prod_{i=1}^d \widehat K(\lambda_i \xi_i)}^2\abs{\widehat\psi(\xi)}^2 \dd \xi \\
		\leq\ &S^2\int_{\norm{\xi}_2\leq t} \abs{\widehat\psi(\xi)}^2 \dd \xi + T_s^2\int_{\norm{\xi}_2> t} \norm{(\lambda_1\xi_1,\dots,\lambda_d\xi_d)}_2^{2s}\abs{\widehat\psi(\xi)}^2 \dd \xi \\
		\leq\ &S^2\norm{\widehat \psi}_2^2 + T_s^2\int_{\R^d} \p{\sum_{i=1}^d \lambda_i^2\xi_i^2}^s\abs{\widehat\psi(\xi)}^2 \dd \xi \\
		\leq\ &S^2(2\pi)^d\norm{\psi}_2^2 + T_s^2\int_{\R^d} \left(\sum_{i=1}^d\lambda_i^2\right)^s\left(\sum_{i=1}^d\xi_i^2\right)^s\abs{\widehat\psi(\xi)}^2 \dd \xi \\
		=\ &S^2(2\pi)^d\norm{\psi}_2^2 + T_s^2 \norm{\lambda}_2^{2s} \int_{\R^d} \norm{\xi}_2^{2s}\abs{\widehat\psi(\xi)}^2 \dd \xi \\
		\leq\ &S^2(2\pi)^d\norm{\psi}_2^2 + T_s^2 \norm{\lambda}_2^{2s} (2\pi)^d R^2 
	\end{align*}
	since $\psi\in\Sb$, and where we have used Plancherel's Theorem for $\psi\in L^1(\R^d)\cap L^2(\R^d)$. 
	We have proved that there exists some $S\in(0,1)$ such that
	$$
		\norm{\psi-\psi*\vl}_2^2\leq S^2 \norm{\psi}_2^2 + T_s^2 R^2 \norm{\lambda}_2^{2s}.
	$$
	If $s \geq 1$, then $x\mapsto x^s$ is convex and, by Jensen's inequality (finite form), we have
	$$
		\norm{\lambda}_2^{2s} 
		= \left(\sum_{i=1}^d\lambda_i^2\right)^s
		= d^s \left(\sum_{i=1}^d\frac{1}{d}\lambda_i^2\right)^s 
		\leq d^s \sum_{i=1}^d\frac{1}{d}\left(\lambda_i^2\right)^s
		= d^{s-1} \sum_{i=1}^d \lambda_i^{2s}
		\leq d^{1+s} \sum_{i=1}^d \lambda_i^{2s}.
	$$
	If $s <1$, then $\gamma \coloneqq \frac{1}{s}>1$ and so, it is a standard result that $\norm{\cdot}_\gamma \leq \norm{\cdot}_1$. We then have
	$$
		\norm{\lambda}_2^{2s} 
		= \left(\sum_{i=1}^d\lambda_i^2\right)^s
		= \left(\sum_{i=1}^d \left(\lambda_i^{2s}\right)^\gamma\right)^{1/\gamma}
		= \norm{\lambda^{2s}}_\gamma
		\leq \norm{\lambda^{2s}}_1
		= \sum_{i=1}^d \lambda_i^{2s}
		\leq d^{1+s} \sum_{i=1}^d \lambda_i^{2s}.
	$$
	Hence, for all $s>0$, we have $\norm{\lambda}_2^{2s} \leq d^{1+s} \sum_{i=1}^d \lambda_i^{2s}$.
	We conclude that
	$$
		\norm{\psi-\psi*\vl}_2^2\leq S^2\norm{\psi}_2^2 + T_s^2 R^2 d^{1+s} \sum_{i=1}^d \lambda_i^{2s}
	$$
	which proves the statement presented in \Cref{intermediate result} with $C_4'(d,s,R)\coloneqq T_s^2 R^2 d^{1+s}$.

	We now consider the constant $C_3(M,d)$ from \Cref{phipower}.
	Suppose we have
	\begin{align*}
		\norm{\psi}^2_2 &\geq \frac{T_s^2 R^2 d^{1+s}}{1-S^2} \sum_{i=1}^d \lambda_i^{2s} + \frac{C_3}{(1-S^2)}\frac{\ln\!\left(\frac{1}{\alpha}\right)}{\beta  \mpn \sqrt{\LL}} \\
		(1-S^2)\norm{\psi}^2_2 &\geq T_s^2 R^2 d^{1+s} \sum_{i=1}^d \lambda_i^{2s} + C_3\frac{\ln\!\left(\frac{1}{\alpha}\right)}{\beta  \mpn \sqrt{\LL}} \\
		\norm{\psi}^2_2 &\geq S^2\norm{\psi}_2^2 + T_s^2 R^2 d^{1+s} \sum_{i=1}^d \lambda_i^{2s} + C_3\frac{\ln\!\left(\frac{1}{\alpha}\right)}{\beta  \mpn \sqrt{\LL}} \\
		\norm{\psi}^2_2 &\geq \norm{\psi-\psi * \varphi_\lambda}^2_2 + C_3\frac{\ln\!\left(\frac{1}{\alpha}\right)}{\beta  \mpn \sqrt{\LL}} 		
	\end{align*}
	then, by \Cref{phipower}, we can ensure that
	$$
		\pii\leq \bb.
	$$
	By definition of uniform separation rates, we deduce that
	\begin{align*}
		\rho\!\left(\db, \Sb, \beta,M\right)^2 &\leq \frac{T_s^2 R^2 d^{1+s}}{1-S^2} \sum_{i=1}^d \lambda_i^{2s} + \frac{C_3}{(1-S^2)}\frac{\ln\!\left(\frac{1}{\alpha}\right)}{\beta  \mpn \sqrt{\LL}}  \\
		&\leq C_4(M, d,s,R,\beta) \p{\sum_{i=1}^d \lambda_i^{2s} + \frac{\ln\!\left(\frac{1}{\alpha}\right)}{ \mpn \sqrt{\LL}} }		
	\end{align*}
	for $C_4(M, d,s,R,\beta)\coloneqq \max\acc{\frac{T_s^2 R^2 d^{1+s}}{1-S^2}, \frac{C_3(M,d)}{\beta(1-S^2)}}$.

\subsection{Proof of \Cref{sobolevusroptimal}}
	\label{proofsobolevusroptimal}
	By \Cref{sobolevusr}, if $\lambda_1\cdots\lambda_d\leq1$, we have
	$$
		\rho\!\left(\db, \Sb, \beta,M\right)^2 \leq C_4(M, d,s,R,\beta) \p{\sum_{i=1}^d \lambda_i^{2s} + \frac{\ln\!\left(\frac{1}{\alpha}\right)}{ \mpn \sqrt{\LL}} }.
	$$
	We want to express the bandwidths in terms of the sum of sample sizes $m+n$ raised to some negative power such that the terms $\sum_{i=1}^d \lambda_i^{2s}$ 
	and $\frac{1}{ \mpn \sqrt{\LL}}$ have the same behaviour in $m+n$. 
	With the choice of bandwidths $\lambda^*_i \coloneqq \mpn^{-2/(4s+d)}$ for $i=1,\dots,d$, the term $\sum_{i=1}^d (\lambda^*_i)^{2s}$ has order $\mpn^{-4s/(4s+d)}$
	and the term $\frac{1}{ \mpn \sqrt{\lambda_1^*\cdots\lambda_d^*}}$ has order $\mpn^{d/(4s+d)-1} = \mpn^{-4s/(4s+d)}$. So, indeed, this choice of bandwidths leads to the same behaviour in $m+n$ for the two terms, which gives the smallest order of $m+n$ possible. 
	It is clear that $\lambda_1^*\cdots \lambda_d^*<1$,
	we find that 
	\begin{align*}
		\rho\!\left(\Delta^{\lambda^*,B}_\alpha, \Sb, \beta,M\right)^2 &\leq C_4(M, d,s,R,\beta) \p{\sum_{i=1}^d (\lambda_i^*)^{2s} + \frac{\ln\!\left(\frac{1}{\alpha}\right)}{ \mpn \sqrt{\lambda_1^*\cdots \lambda_d^*}} }\\
		&\leq C_4(M, d,s,R,\beta) \p{\mpn^{-4s / (4s+d)} + \lna \mpn^{-4s / (4s+d)}}\\
		&\leq C_4(M, d,s,R,\beta) \lna \mpn^{-4s / (4s+d)}\\
		&= C_5(M, d,s,R,\alpha,\beta)^2\, \mpn^{-4s / (4s+d)}
	\end{align*}
	for $C_5(M, d,s,R,\alpha,\beta)\coloneqq \sqrt{C_4(M, d,s,R,\beta) \lna}$.
	We deduce that
	$$
		\rho\!\left(\Delta^{\lambda^*,B}_\alpha, \Sb, \beta,M\right)
		\leq C_5(M, d,s,R,\alpha,\beta)\, \mpn^{-2s / (4s+d)}.
	$$

\subsection{Proof of \Cref{levelagg}}
	\label{prooflevelagg}
	By definition of $\uu$, we have
	$$
		\frac{1}{B_2}\sum_{b=1}^{B_2}\one{\!\aamax{\lambda\in\Lambda}\!\p{\!\widehat M_{\lambda,2}^{\, b}\p{\mu^{(b,2)} \big|\Xm,\Yn}\!-\widehat q_{1-\uuv
		\ww }^{\,\lambda, B_1}\!\p{\Zbo\big|\Xm,\Yn}\!}\!>0}\!\leq \alpha.
	$$
	Taking the expectation on both sides, we get 
	$$
		\pprr{\aamax{\lambda\in\Lambda}\p{\mmdhv-\widehat q_{1-\uuv
		 \ww }^{\,\lambda, B_1}\!\p{\Zbo\big|\Xm,\Yn}}>0} \leq \alpha
	$$
	as under the null hypothesis $\HH_0\colon p=q$, we have $\big(\widehat M_{\lambda,2}^{\, b}\p{\mu^{(b,2)} \big|\Xm,\Yn}\big)_{1\leq b \leq B_2}$  distributed like $\mmdhv$.	
	Using the bisection search approximation which satisfies 
	$$
		\wuuv \leq \uuv,
	$$ 
	we get
	\begin{align*}
		&\pprr{\aamax{\lambda\in\Lambda}\p{\mmdhv-\widehat q_{1-\wuuv \ww }^{\,\lambda, B_1}\!\p{\Zbo\big|\Xm,\Yn}}>0} \\
		\leq\ &\pprr{\aamax{\lambda\in\Lambda}\p{\mmdhv-\widehat q_{1-\uuv \ww }^{\,\lambda, B_1}\!\p{\Zbo\big|\Xm,\Yn}}>0} \\
		\leq\ &\alpha.
	\end{align*}
	We deduce that 
	$$
		\pprr{\dbbv=1}
		\leq \alpha.
	$$

\subsection{Proof of \Cref{sobolevusragg}}
	\label{proofsobolevusragg}

	Consider some $ u^*\in(0,1)$ to be determined later.
	For $b=1,\dots,B_2$, let 
	\[
		W_b\p{\mu^{(b,2)} \big|\Xm,\Yn,\Zbo} \coloneqq \one{\aamax{\lambda\in\Lambda}\p{\widehat M_{\lambda,2}^{\, b}\p{\mu^{(b,2)} \big|\Xm,\Yn}-\widehat q_{1- u^* \ww }^{\,\lambda, B_1}\!\p{\Zbo\big|\Xm,\Yn}}>0},
	\]
	so that, following a similar argument to the one presented in \Cref{prooflevelagg}, we obtain that $\EE{p\times q\times r \times r}{W_b\p{\mu^{(b,2)} \big|\Xm,\Yn,\Zbo}}$ is equal to
	\[
		\ppr{\aamax{\lambda\in\Lambda}\p{\mmdhv-\widehat q_{1- u^* \ww }^{\,\lambda, B_1}\!\p{\Zbo\big|\Xm,\Yn}}>0}.
	\]
	Consider the events 
	\[
		\mathcal{A}' \coloneqq \acc{\frac{1}{B_2}\sum_{b=1}^{B_2}W_b\p{\mu^{(b,2)} \big|\Xm,\Yn,\Zbo} - \EE{r}{W_b\p{\mu^{(b,2)} \big|\Xm,\Yn,\Zbo}} \leq \sqrt{\frac{1}{2B_2}\ln{\p{\frac{2}{\beta}}}}}
	\]
	and
	\begin{align*}
		\mathcal{A} \coloneqq \Bigg\{&\frac{1}{B_2}\sum_{b=1}^{B_2}W_b\p{\mu^{(b,2)} \big|\Xm,\Yn,\Zbo} - \EE{p\times q\times r \times r}{W_b\p{\mu^{(b,2)} \big|\Xm,\Yn,\Zbo}} \\
		&\leq \sqrt{\frac{1}{2B_2}\ln{\p{\frac{2}{\beta}}}}\Bigg\}.
	\end{align*}
	Using Hoeffding's inequality, we obtain that $\mathbb{P}_r\p{\mathcal{A}'\big|\Xm,\Yn,\Zbo}\geq 1-\frac{\beta}{2}$ for any $\Xm$, $\Yn$ and $\Zbo$, we deduce that $\pqrr{\mathcal{A}}\geq 1-\frac{\beta}{2}$.

	First, assuming that the event $\mathcal{A}$ holds, we show that $\uuv \geq \alpha$. 
	Since we assume that the event $\mathcal{A}$ holds, the bounds we obtain hold with probability $1-\frac{\beta}{2}$.
	We have
	\begin{align*}
		&\frac{1}{B_2}\sum_{b=1}^{B_2}\one{\aamax{\lambda\in\Lambda}\p{\widehat M_{\lambda,2}^{\, b}\p{\mu^{(b,2)} \big|\Xm,\Yn}-\widehat q_{1- u^* \ww }^{\,\lambda, B_1}\!\p{\Zbo\big|\Xm,\Yn}}>0}\\
		=\ &\frac{1}{B_2}\sum_{b=1}^{B_2} W_b\p{\mu^{(b,2)} \big|\Xm,\Yn,\Zbo} \\
		\leq\ &\EE{p\times p \times r\times r}{W_b\p{\mu^{(b,2)} \big|\Xm,\Yn,\Zbo}} + \sqrt{\frac{1}{2B_2}\ln{\p{\frac{2}{\beta}}}} \\
		=\ &\ppr{\aamax{\lambda\in\Lambda}\p{\mmdhv-\widehat q_{1- u^* \ww }^{\,\lambda, B_1}\!\p{\Zbo\big|\Xm,\Yn}}>0}+ \sqrt{\frac{1}{2B_2}\ln{\p{\frac{2}{\beta}}}}\\
		=\ &\ppro{\bigcup_{\lambda\in\Lambda}\acc{\mmdhv>\widehat q_{1- u^* \ww }^{\,\lambda, B_1}\!\p{\Zbo\big|\Xm,\Yn}}}+ \sqrt{\frac{1}{2B_2}\ln{\p{\frac{2}{\beta}}}}.
	\end{align*}
	Using Boole's inequality, we obtain
	\begin{align*}
		&\frac{1}{B_2}\sum_{b=1}^{B_2}\one{\aamax{\lambda\in\Lambda}\p{\widehat M_{\lambda,2}^{\, b}\p{\mu^{(b,2)} \big|\Xm,\Yn}-\widehat q_{1- u^* \ww }^{\,\lambda, B_1}\!\p{\Zbo\big|\Xm,\Yn}}>0}\\
		\leq\ &\sum_{\lambda\in\Lambda}\ppro{\mmdhv>\widehat q_{1- u^* \ww }^{\,\lambda, B_1}\!\p{\Zbo\big|\Xm,\Yn}} + \sqrt{\frac{1}{2B_2}\ln{\p{\frac{2}{\beta}}}}\\
		\leq\ &\sum_{\lambda\in\Lambda} u^* \ww + \sqrt{\frac{1}{2B_2}\ln{\p{\frac{2}{\beta}}}}\\
		\leq\ &u^* + \sqrt{\frac{1}{2B_2}\ln{\p{\frac{2}{\beta}}}}\\
		=\ &\frac{3\alpha}{4} + \sqrt{\frac{1}{2B_2}\ln{\p{\frac{2}{\beta}}}}
	\end{align*}
	for $ u^* \coloneqq \frac{3\alpha}{4}$, where we have used \Cref{level} and the fact that $\sum_{\lambda\in\Lambda} w_\lambda \leq 1$.
	Now, for $B_2 \geq \frac{8}{\alpha^2} \ln\p{\frac{2}{\beta}}$, we get
	$$
		\frac{3\alpha}{4} + \sqrt{\frac{1}{2B_2}\ln{\p{\frac{2}{\beta}}}}
		\leq \alpha
	$$
	and so, we obtain
	$$
		\frac{1}{B_2}\sum_{b=1}^{B_2}\one{\aamax{\lambda\in\Lambda}\p{\widehat M_{\lambda,2}^{\, b}\p{\mu^{(b,2)} \big|\Xm,\Yn}-\widehat q_{1- u^* \ww }^{\,\lambda, B_1}\!\p{\Zbo\big|\Xm,\Yn}}>0} 
		\leq \alpha.
	$$
	Recall that $\uuv$ is defined as 
	$$
		\sup\!\bigg\{\!u\!\in\!\!\Big(\!0,\aamin{\lambda\in\Lambda}\ww^{-1}\!\Big)\!\!:\! \frac{1}{B_2}\!\sum_{b=1}^{B_2}\!\one{\!\aamax{\lambda\in\Lambda}\p{\widehat M_{\lambda,2}^{\, b}\p{\!\mu^{(b,2)} \big|\Xm,\Yn\!}\!-\widehat q_{1-u \ww }^{\,\lambda, B_1}\!\p{\Zbo\big|\Xm,\Yn}\!}\!\!>\!0\!}\!\leq\! \alpha\!\bigg\},
	$$
	we deduce that 
	$$
		\uuv\geq  u^* = \frac{3\alpha}{4}
	$$
	for $B_2 \geq \frac{8}{\alpha^2} \ln\p{\frac{2}{\beta}}$ when the event $\mathcal{A}$ holds. 

	Under the event $\mathcal{A}$, after performing $B_3$ steps of the bisection method, we have
	\begin{align*}
		\wuuv &\geq \uuv - \frac{\aamin{\lambda\in\Lambda}\ww^{-1}}{2^{B_3}} \\
		&\geq \frac{3\alpha}{4} - \frac{\aamin{\lambda\in\Lambda}\ww^{-1}}{2^{B_3}}\\
		&\geq \frac{\alpha}{2}
	\end{align*}
	for $B_3 \geq \log_2\!\Big(\frac{4}{\alpha}\,\aamin{\lambda\in\Lambda}\ww^{-1}\Big)$.

	We are interested in upper bounding the probability of type II error $\pqrr{\mathcal{B}}$ for the event $\mathcal{B}\coloneqq \acc{\dbbv=0}$. We have
	\begin{align*}
		\pqrr{\mathcal{B}} &= \pqrr{\mathcal{B}\big|\mathcal{A}}\, \pqrr{\mathcal{A}} + \pqrr{\mathcal{B}\big|\mathcal{A}^c}\, \pqrr{\mathcal{A}^c} \\
		&\leq \pqrr{\mathcal{B}\big|\mathcal{A}} + \frac{\beta}{2}
	\end{align*}
	where
	\begin{align*}
		&\pqrr{\mathcal{B}\big|\mathcal{A}}\\
		=\ &\pqrr{\dbbv = 0\,\Big|\,\mathcal{A}} \\
		=\ &\pqrr{\bigcap_{\lambda\in\Lambda}\acc{\mmdhv\leq \widehat q_{1-\wuuv \ww }^{\,\lambda, B_1}\!\p{\Zbo\big|\Xm,\Yn}}\,\Big|\,\mathcal{A}}\\
		\leq\ &\aamin{\lambda\in\Lambda}\ {\pqrr{{\mmdhv\leq \widehat q_{1-\wuuv \ww }^{\,\lambda, B_1}\!\p{\Zbo\big|\Xm,\Yn}\,\Big|\,\mathcal{A}}}}\\
		\leq\ &\aamin{\lambda\in\Lambda}\ \pqro{{\mmdhv\leq \widehat q_{1- \alpha \ww /2}^{\, \lambda, B_1}\p{\Zbo\big|\Xm,\Yn}}}\\
		=\ &\aamin{\lambda\in\Lambda} \ {\pqro{{\Delta}^{\lambda, B_1}_{\alpha \ww /2}\!\left(\Zbo\big|\Xm,\Yn\right)=0}},
	\end{align*}
	we deduce that
	\begin{equation}
		\label{bound aggregated power}
		\hspace{-0.02cm}\pqrr{\dbbv \!=\! 0} \!\leq\! \frac{\beta}{2}\!+\aamin{\lambda\in\Lambda} \, {\pqro{{\Delta}^{\lambda, B_1}_{\alpha \ww /2}\!\left(\Zbo\big|\Xm,\Yn\right)\!=\!0\!}}\hspace{-0.03cm}.\hspace{-2cm}		
	\end{equation}
	In order to upper bound $\pqrr{\dbbv = 0}$ by $\beta$ it is sufficient to upper bound $\aamin{\lambda\in\Lambda} \ {\pqro{{\Delta}^{\lambda, B_1}_{\alpha \ww /2}\!\left(\Xm,\Yn,\Zbo\right)=0}}$ by $\frac{\beta}{2}$.
	By definition of uniform separation rates, it follows that
	\begin{equation*}
		\rho\!\left(\dbb, \Sb, \beta,M\right)^2 \leq 4\,\aamin{\lambda\in\Lambda} \ \rho\!\left({\Delta}^{\lambda, B_1}_{\alpha \ww /2}, \Sb, \beta,M\right)^2.
	\end{equation*}
	For each $\lambda\in\lambda$, since $\big(1-\alpha \ww/2\big) \alpha \ww /2 \leq (1-\alpha)\alpha$ as $\alpha\in(0,e^{-1})$ and $\ww\leq 1$, we have
	\begin{align*}
		B_1 &\geq \Big(\mathrm{max}_{\lambda\in\Lambda}\, \ww^{-2}\Big) \frac{12}{\alpha^2}\left(\log\left(\frac{8}{\beta}\right)+\alpha(1-\alpha)\right) \\
		&\geq \frac{3}{\left(\alpha \ww / 2\right)^2} \left(\log\left(\frac{8}{\beta}\right)+\frac{\alpha \ww}{2}\left(1-\frac{\alpha\ww}{2}\right)\right),
	\end{align*}
	so we can apply \Cref{sobolevusr} to the tests
	$\left({\Delta}^{\lambda, B_1}_{\alpha \ww /2}\right)_{\lambda\in\Lambda}$ to obtain
	\begin{align*}
		\rho\!\left(\dbb, \Sb, \beta,M\right)^2 &\leq 4\,\aamin{\lambda\in\Lambda} \ \rho\!\left({\Delta}^{\lambda, B_1}_{\alpha \ww /2}, \Sb, \beta,M\right)^2 \\
		&\leq 4 C_4(M, d,s,R,\beta)\, \aamin{\lambda\in\Lambda}\p{\sum_{i=1}^d \lambda_i^{2s} + \frac{\ln\!\left(\frac{2}{\alpha w_\lambda}\right)}{ \mpn \sqrt{\LL}}}\\
		&\leq C_6(M, d,s,R,\beta)\, \aamin{\lambda\in\Lambda}\p{\sum_{i=1}^d \lambda_i^{2s} + \frac{\ln\!\left(\frac{1}{\alpha w_\lambda}\right)}{ \mpn \sqrt{\LL}}}
	\end{align*}
	for $C_6(M, d,s,R,\beta)\coloneqq 8 C_4(M, d,s,R,\beta)$ where $C_4(M, d,s,R,\beta)$ is the constant from \Cref{sobolevusr}, and where we used the fact that
	$$
		\ln\p{\frac{2}{\alpha w_\lambda}} 
		= \ln(2) + \ln\p{\frac{1}{\alpha w_\lambda}}
		\leq \big(\ln(2)+1\big)\ln\p{\frac{1}{\alpha w_\lambda}}
		\leq 2 \ln\p{\frac{1}{\alpha w_\lambda}}
	$$
	as $\ln\p{\frac{1}{\alpha w_\lambda}}\geq \ln\p{\frac{1}{\alpha}} >1$ since $\alpha\in(0,e^{-1})$.

\subsection{Proof of \Cref{sobolevusraggoptimal}}
	\label{proofsobolevusraggoptimal}
	First, note that we indeed have
	$$
		\sum_{\lambda \in \Lambda} \ww 
		< \frac{6}{\pi^2}\sum_{\ell=1}^\infty \frac{1}{\ell^2}
		= 1
	$$
	and also that for all $\lambda=(2^{-\ell},\dots,2^{-\ell})\in\Lambda$ we have
	$\LL = 2^{-d\ell} < 1$
	as $\ell,d\in\Nz$.
	Let $\lambda^* = (2^{-\ell^*},\dots,2^{-\ell^*})\in\Lambda$ where
	$$
		\ell^* \coloneqq \ceil*{\frac{2}{4s+d}\log_2\!\left(\frac{m+n}{\ln(\ln(m+n))}\right)} \leq \ceil*{\frac{2}{d}\log_2\!\left(\frac{m+n}{\ln(\ln(m+n))}\right)}.
	$$
	Since $\aamin{\lambda\in\Lambda}\ww^{-1} = \frac{6}{\pi^2}$, we have $B_3 \geq \log_2\!\Big(\frac{4}{\alpha}\aamin{\lambda\in\Lambda}\ww^{-1}\Big)$, so we can apply \Cref{sobolevusragg} to get
	\begin{align*}
		\rho\!\left(\dbb, \Sb, \beta,M\right)^2 &\leq C_6(M, d,s,R,\beta)\, \aamin{\lambda\in\Lambda}\p{\sum_{i=1}^d \lambda_i^{2s} + \frac{\ln\!\left(\frac{1}{\alpha}\right)+\ln\!\left(\frac{1}{w_\lambda}\right)}{ \mpn \sqrt{\LL}}} \\
		&\leq C_6(M, d,s,R,\beta)\p{\sum_{i=1}^d (\lambda^*_i)^{2s} + \frac{\ln\!\left(\frac{1}{\alpha }\right)+\ln\!\left(\frac{1}{w_{\lambda^*}}\right)}{ \mpn \sqrt{\lambda^*_1\cdots \lambda^*_d}}}.
	\end{align*}
	Note that
	$
		\ell^* \leq \frac{2}{4s+d}\log_2\!\left(\frac{m+n}{\ln(\ln(m+n))}\right)+1
	$
	which gives 
	$
		\lambda_i^* = 2^{-\ell^*} \geq 2^{-1} \left(\frac{m+n}{\ln(\ln(m+n))}\right)^{-2 / (4s+d)}
	$
	for $i=1,\dots,d$.
	We get
	$
		\sqrt{\lambda^*_1\cdots\lambda^*_d}
		\geq 2^{-\frac{d}{2}} \left(\frac{m+n}{\ln(\ln(m+n))}\right)^{-d / (4s+d)}
	$
	and so
	$$
		\frac{1}{\sqrt{\lambda^*_1\cdots\lambda^*_d}}\leq 2^{\frac{d}{2}} \left(\frac{m+n}{\ln(\ln(m+n))}\right)^{d / (4s+d)}.
	$$
	Note also that
	\begin{align*}
		\ell^* &\leq \frac{2}{4s+d} \log_2\!\left(\frac{m+n}{\ln(\ln(m+n))}\right)+1 \\
		&\leq \frac{2}{4s+d} \log_2(m+n)+1\\
		&\leq \p{\frac{2}{d\ln(2)}+1} \ln(m+n)\\
		&< 4 \ln(m+n)
	\end{align*}
	as $\ln(\ln(m+n))>1$ and $\ln(m+n)>1$. 
	We get
	\begin{align*}
		\ln\!\left(\frac{1}{w_{\lambda^*}}\right)
		&= 2\ln\!\left(\ell^*\right)+\ln\!\left(\frac{\pi^2}{6}\right)\\
		&\leq  2 \ln(4 \ln(m+n)) +\ln\!\left(\frac{\pi^2}{6}\right)\\
		&\leq  \p{2 \ln(4) + 1 + \ln\!\left(\frac{\pi^2}{6}\right)} \ln(\ln(m+n)) \\
		&< 5 \ln(\ln(m+n)) 
	\end{align*}
	as $\ln(\ln(m+n))>1$.
	Combining those upper bounds, we get
	\begin{align*}
		\frac{\ln\!\left(\frac{1}{\alpha}\right)+\ln\!\left(\frac{1}{w_{\lambda^*}}\right)}{ \mpn \sqrt{\lambda^*_1\dots\lambda^*_d}}
		&\leq \frac{1}{ \mpn \sqrt{\lambda^*_1\dots\lambda^*_d}}\left(\ln\!\left(\frac{1}{\alpha}\right) + 5 \ln(\ln(m+n))\right) \\
		&\leq \p{\lna+5}\frac{\ln(\ln(m+n))}{m+n} \frac{1}{\sqrt{\lambda^*_1\dots\lambda^*_d}} \\
		&\leq 2^{\frac{d}{2}}\p{\lna+5} \frac{\ln(\ln(m+n))}{m+n} \left(\frac{m+n}{\ln(\ln(m+n))}\right)^{d / (4s+d)} \\
		&= 2^{\frac{d}{2}}\p{\lna+5}\left(\frac{m+n}{\ln(\ln(m+n))}\right)^{-4s / (4s+d)}
	\end{align*}
	as $\ln(\ln(m+n))>1$.
	Note also that
	$$
		\ell^* \geq \frac{2}{4s+d} \log_2\!\left(\frac{m+n}{\ln(\ln(m+n))}\right)
	$$
	giving
	$$
		(\lambda_i^*)^{2s} = (2^{- \ell^*})^{2s} \leq  \left(\frac{m+n}{\ln(\ln(m+n))}\right)^{-4s / (4s+d)}
	$$
	for $i=1,\dots,d$. Hence, we get
	$$
		\sum_{i=1}^d (\lambda_i^*)^{2s} \leq d \left(\frac{m+n}{\ln(\ln(m+n))}\right)^{-4s / (4s+d)}.
	$$
	We obtain
	\begin{align*}
		\rho\!\left(\dbb, \Sb, \beta,M\right)^2 
		&\leq C_6(M, d,s,R,\beta)\p{\sum_{i=1}^d (\lambda^*_i)^{2s} + \frac{\ln\!\left(\frac{1}{\alpha}\right)+\ln\!\left(\frac{1}{w_{\lambda^*}}\right)}{ \mpn \sqrt{\lambda^*_1\cdots \lambda^*_d}}} \\
		&\leq C_7(M, d,s,R,\alpha,\beta)^2 \left(\frac{m+n}{\ln(\ln(m+n))}\right)^{-4s / (4s+d)}
	\end{align*}
	where
	$
		C_7(M, d,s,R,\alpha,\beta)\coloneqq \sqrt{
		C_6(M, d,s,R,\beta) \max \acc{
		d, 2^{\frac{d}{2}}\p{\lna+5}
		}
		}
	$.
	We conclude that
	$$
		\rho\!\left(\dbb, \Sb, \beta,M\right) \leq C_7(M, d,s,R,\alpha,\beta)
		\left(\frac{m+n}{\ln(\ln(m+n))}\right)^{-2s / (4s+d)}.
	$$
	Hence, the test $\dbb$ is optimal in the minimax sense up to an iterated logarithmic term.
	Since it does not depend on the unknown parameters $s$ and $R$, our aggregated MMDAgg test $\dbb$ is minimax adaptive over the Sobolev balls $\Sbb$.

\clearpage
\bibliographystyle{abbrv}
\bibliography{biblio}

\end{document}